\documentclass[10pt,twocolumn,letterpaper]{article}

\usepackage[pagenumbers]{cvpr}

\usepackage[T1]{fontenc}
\usepackage{colortbl,tabularx}
\usepackage{cuted}
\usepackage{currfile}
\usepackage{algorithm}
\usepackage{algpseudocode}

\newcommand{\method}{RALLY}

\renewcommand{\dbltopfraction}{0.9}
\renewcommand{\dblfloatpagefraction}{0.8}
\renewcommand{\textfraction}{0.08}

\usepackage{titletoc}
\newcommand{\minisection}[1]{\noindent{\textbf{#1}.}}

\definecolor{cvprblue}{rgb}{0.21,0.49,0.74}
\usepackage[breaklinks,colorlinks,allcolors=cvprblue]{hyperref}

\def\confName{CVPR}
\def\confYear{2027}

\makeatletter
\def\@maketitle{%
   \newpage
   \null
   \vskip -0.15in
   \begin{center}
      {\Large \bf \@title \par}
      \vspace*{10pt}
      {\large
        \lineskip .5em
        \begin{tabular}[t]{c}\@author\end{tabular}
        \par}
      \vspace*{2pt}
   \end{center}
}
\def\abstract{%
   \centerline{\large\bf Abstract}%
   \vspace*{6pt}\noindent%
   \it\ignorespaces%
}
\makeatother

\title{Structuring Robotic Harnesses Through Adversarial and Collaborative Play}
\author{
Hanjun Yoo$^{1*}$ \quad Kelvin Li$^{1*}$ \quad Kian Hosseinkhani$^{2*}$ \quad Kaito Suzuki$^{1}$\\
Youngwoon Lee$^{3}$ \quad Ke Li$^{2}$ \quad Trevor Darrell$^{1}$ \quad Roei Herzig$^{1}$\\[0.4em]
{\normalsize $^{1}$University of California, Berkeley \quad $^{2}$Simon Fraser University \quad $^{3}$Seoul National University}\\[1pt]
{\normalsize Project page: \url{https://rally-robot.github.io/}}
}

\begin{document}
\maketitle
\begin{strip}
\vspace{-9.9mm}
\centering
\includegraphics[width=\textwidth]{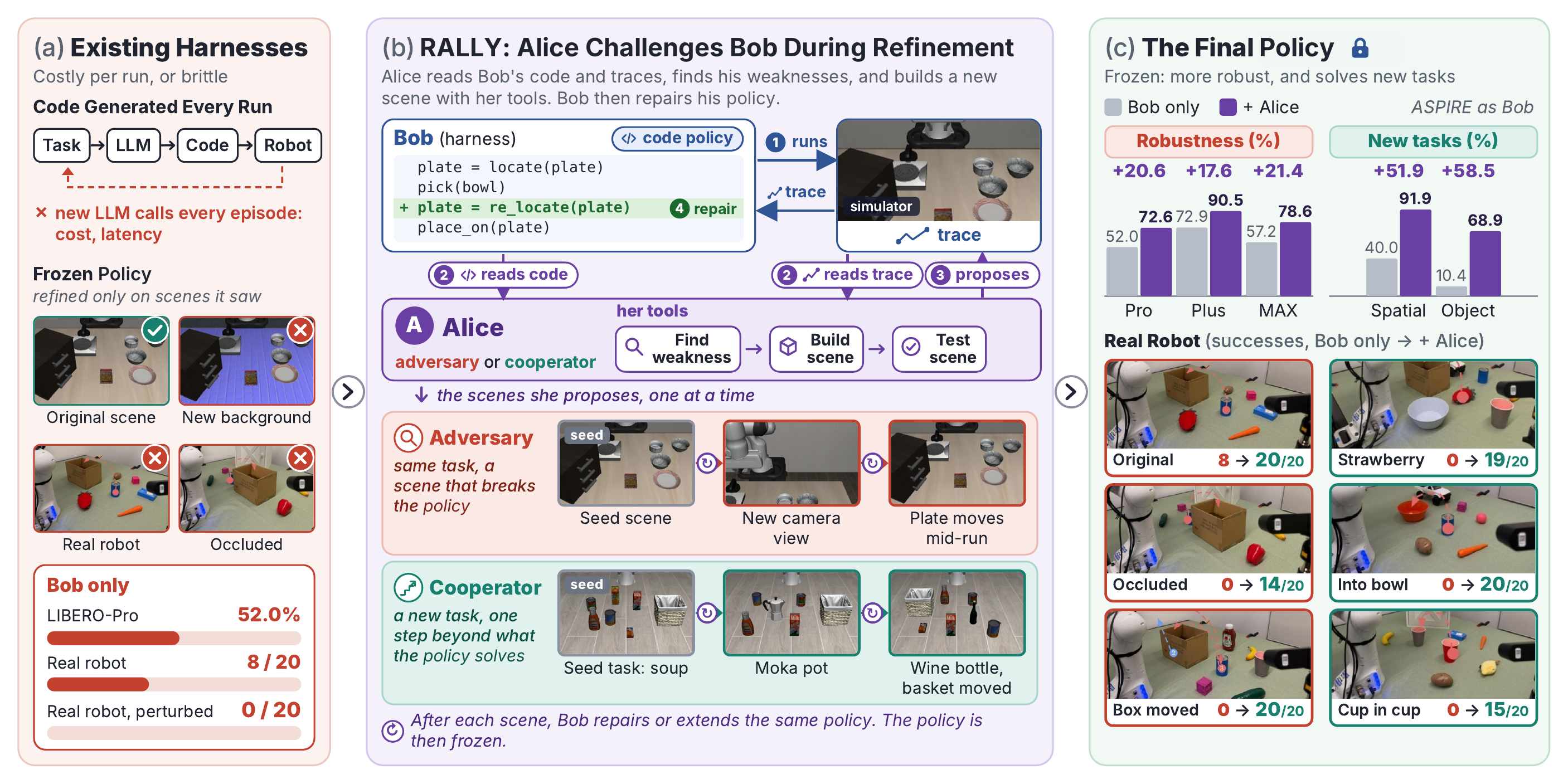}
\captionsetup{hypcap=false}
\captionof{figure}{\textbf{Alice and Bob refine one frozen code policy together.}
(a)~Existing harnesses call an LLM in every episode or refine a frozen policy only on its given scenes, failing under real-world perturbations.
(b)~In \method{}, Alice reads Bob's code and traces to find weaknesses, then proposes task-preserving scene variations as an adversary or related tasks as a cooperator. Bob refines the same policy, which is then frozen for reuse.
(c)~The final policy is more robust on LIBERO-Pro, LIBERO-Plus, and LIBERO-MAX, generalizes to new LIBERO tasks, and transfers zero-shot to a real Franka arm along with completing new tasks and showing robustness to perturbation.}
\captionsetup{hypcap=true}
\label{fig:\currfilebase}%
\vspace{-2mm}
\end{strip}

{\renewcommand{\thefootnote}{\fnsymbol{footnote}}\footnotetext[1]{Equal contribution.}}

\begin{abstract}

Robotic harnesses let foundation models solve physical tasks through generated code policies. 
However, these policies still struggle to remain robust to variations in objects and their locations, camera views, or backgrounds, and to generalize across tasks.
In this paper, we argue that refining a code policy only on the given task and scenes leaves failures in new scenes or tasks undetected until deployment.
Inspired by how coaches rally with players to expose weaknesses and teach new skills, we introduce \method{}, a harness-agnostic framework that structures policy refinement as a two-player game. 
Alice is an additional agent that reads the policy of an existing harness (Bob) and proposes environment variations or related tasks on which Bob refines it.
We consider two settings. To improve policy robustness, Alice acts as an adversary and proposes task-preserving variations (e.g., camera views or object positions) meant to break the current policy. To help the policy generalize across tasks, Alice acts as a collaborator and proposes related tasks (e.g., manipulating a different object), and Bob extends the same policy to solve each new task.
We refine all policies on LIBERO. With Alice as an adversary, \method{} raises zero-shot success for ASPIRE (+19.9\%), RHO (+15.0\%), and GaP (+13.7\%) on LIBERO-Pro, LIBERO-Plus, and LIBERO-MAX. With Alice as a collaborator, policies refined on one task of a suite reach 62.8\% average success on the suite's other tasks, compared with 11.7\% without Alice. Notably, \method{} enables Sonnet 5.5 to outperform Fable 5.2 at a third of its cost, and our simulation-refined policies succeed in 92.5\% of real-robot trials on new tasks.

\end{abstract}

\section{Introduction}
\label{sec:introduction}

Robotic harnesses use the programming capabilities of frozen foundation models to solve physical tasks. Given a task and access to the robot's perception and control tools, the model writes code, runs it, and uses the resulting feedback to revise it~\cite{fu2026capx,lu2026aspire}. The model might need to repeat this loop several times before the code succeeds, with each pass adding model calls and execution trials~\cite{rho2026}. Many harnesses call the model each time the robot executes a task, generating new code~\cite{liang2023code} and, in closed-loop settings, revising or extending it from execution feedback~\cite{fu2026capx,zhang2026rats,agent2026policy}. Each execution therefore waits for new model calls, which slows the robot down and adds cost, even for a task that was already solved. Instead, several harnesses use a \emph{frozen code policy}. The model first generates and refines a policy, and the harness then runs it unchanged, with no further model calls for generation or refinement~\cite{lu2026aspire,chen2026gap,rho2026,encore2026}.

Variation is inherent to the physical world, so a frozen code policy must keep working when the same task is performed in new scenes. For example, the appearance and location of objects, the camera view, or the background vary in the noisy physical world, and the scene can change without any action of the robot~\cite{liberopro2025,liberoplus,liberomax}. A harness can make a policy robust to a variation by refining it on scenes that contain the variation. However, there are too many possible variations to refine on individually, and many are unknown in advance. Even after repeated refinement on the original task scenes, the policy can still fail under unseen variations. If the frozen policy fails, the harness must call the model again, reducing the benefit of reuse.

A separate challenge arises when the task itself changes, as existing harnesses refine each policy for one task. 
Yet many manipulation tasks are related: they share many sub-goals, such as finding, grasping, and placing an object. They differ in which object is manipulated, what the target is, or which steps are required. As these tasks share sub-goals, several harnesses save the policies of solved tasks in a database and retrieve the relevant ones when a new task arrives~\cite{fu2026capx,zhang2026rats,lu2026aspire,chen2026gap}. With such a database, the model must still generate a new policy for that task, and this policy may need its own refinement. These harnesses still generate a separate policy for every task, repeating model calls and execution trials that one shared policy could avoid.

Our objective is to make generated policies robust to varying conditions and reusable across related tasks.
When refinement covers only the setups and tasks initially provided, failures in new scenes or tasks may remain undetected until deployment, when the policy is already frozen. Fixing these failures requires further refinement, with additional model calls and execution trials, undermining the benefit of frozen code policies.
Inspired by how human coaches rally with their players to expose weaknesses and teach new skills progressively, we introduce \method{} (\textit{Refinement through Adversarial and coLLaborative plaY}), a harness-agnostic framework that adds such a coach to an existing harness during refinement. The harness, which we call Bob, writes a policy and revises it after failures. The coach, which we call Alice, is an additional agent that reads Bob's policy, sees where it fails, and proposes environment variations or related tasks for the same policy to handle next. The resulting policy is then frozen for reuse.

For robustness, Alice is an adversary that tries to make Bob's policy fail. Starting from a policy that succeeds in the original scenes of one task, Alice proposes task-preserving variations that target weaknesses in the current policy. Bob revises the policy after each failure. For reuse across tasks, Alice instead acts as a collaborator, introducing related tasks that share sub-goals. Bob starts from a task-specific policy and extends it to solve each new task. In both roles, Alice chooses the tasks and variations, while Bob retains its own policy generation and refinement process. \method{} therefore applies to different harnesses without modifying them. To help Alice propose scenes or tasks that expose weaknesses, we design \method{}-Bob, a harness that writes explicit if-else checks showing how the policy chooses the robot’s next action.

We evaluate \method{} on LIBERO with three existing harnesses, ASPIRE, RHO, and GaP, and with our \method{}-Bob. With adversarial Alice, success under held-out scene perturbations rises by 14 to 20 percent over Bob alone. With cooperative Alice, a policy refined on one task of a suite extends to the other nine: averaged over four harnesses and four suites, success on these tasks rises from 11.7\% to 62.8\%, and reaches 76.7\% with \method{}-Bob. With these gains, \method{} beats stronger models in both performance and cost: Sonnet 5.5 with \method{} is more robust than Fable 5.2 without Alice (82.6\% versus 71.1\%), at about a third of its cost.  Finally, \method{} is the first to make code policies refined entirely in simulation robust enough to run in the real world under unseen tasks: our policies succeed in 92.5\% of trials across four new tasks, while the harness alone fails in every trial.

To summarize, our main contributions are fourfold: (i) We introduce \method{}, a harness-agnostic framework for producing code policies robust to unseen variations and reusable across tasks. Our method frames refinement as a two-player game: Alice reads Bob’s policy to identify weaknesses and proposes scene variations or related tasks. (ii) We instantiate Alice in two roles. An adversary makes a task-specific policy robust to scene variations, and a collaborator enables a task-specific policy to generalize across related tasks. (iii) We show that \method{} improves robustness to unseen scene variations and generalization across related tasks for three state-of-the-art harnesses. (iv) \method{} enables code policies refined entirely in simulation to run on a real robot under unseen perturbations and new tasks.

\section{Related Work}
\label{sec:related_work}

\minisection{Robotic harnesses} \looseness=-1 Robotic harnesses use a foundation model to generate a code policy that executes a task~\cite{liang2023code,singh2023progprompt,vemprala2024chatgpt,huang2023voxposer,huang2023instruct2act,mu2024robocodex,chen2024roboscript,shi2025maestro}.
Many harnesses call the model at every execution to generate code, and some call it again to fix the code when it fails~\cite{fu2026capx,zhang2026rats,naouali2026vlcp,agent2026policy}, which adds latency and cost each time.
To avoid this recurring cost, other harnesses first refine a policy, by running it and fixing its failures, and then execute it frozen~\cite{lu2026aspire,chen2026gap,rho2026,encore2026}.
However, they refine each policy only on the original scenes of one task, so it often fails when objects, camera views, or backgrounds vary in the physical world~\cite{liberopro2025,liberoplus,liberomax,wang2026liberox,wang2025vlatest}.
Furthermore, as many tasks share stages, several harnesses reuse code across tasks, but still generate a new policy for every task~\cite{fu2026capx,zhang2026rats,lu2026aspire,chen2026gap}.
In contrast, our \method{} addresses both limitations with a harness-agnostic two-player framework, in which Alice reads the policy and proposes the scenes on which the harness refines it.
As an adversarial agent, Alice targets the policy's weaknesses with variations of the same task; as a cooperative agent, Alice introduces related tasks, so that one policy solves all of them.

\minisection{Robustness and generalization in manipulation} Manipulation policies are often trained for a single task from its demonstrations~\cite{chi2023diffusion,zhao2023act,florence2021implicit,mandlekar2021robomimic}.
Rather than one model per task, generalist policies, such as vision-language-action models, cover many tasks with one model~\cite{brohan2023rt1,brohan2023rt2,openx2024,octo2024,kim2024openvla,black2024pi0,black2025pi05,bjorck2025groot,hosseinkhani2026imlevla}.
Yet a new task usually needs fine-tuning on its own demonstrations~\cite{kim2024openvla,kim2025openvlaoft}.
Domain randomization and data augmentation improve robustness by exposing policies to varied scenes and observations during training~\cite{tobin2017domain,peng2018sim,openai2019rubik,laskin2020rad,yu2023rosie}.
For robotic harnesses, robustness likewise depends on the situations encountered during refinement.
Refining only on the original task scenes can leave failures under new conditions undiscovered.
Some harnesses, such as RHO, execute their policy many times without changing it, so it must also keep working when the scene changes.
However, RHO, refined on the original LIBERO tasks, succeeds in only 45.0\% of the trials under the perturbations of LIBERO-Pro~\cite{rho2026}, in line with our own runs (Table~\ref{tab:main-results}).
\method{} instead builds this robustness, and reuse across tasks, into the code of the policy itself, through an opponent and a teacher that read the policy, as in two-player games.

\minisection{Two-player games} \looseness=-1 Skill development in humans has long been tied either to an opponent, who exposes weaknesses~\cite{ericsson1993deliberate}, or to a teacher, who sets steps of growing difficulty~\cite{vygotsky1978mind}.
In machine learning, training against an opponent first appeared in board games, where an agent plays against itself and learns from every game~\cite{samuel1959checkers,tesauro1995tdgammon}.
Later, deep neural networks were trained against an opponent, both through self-play in games~\cite{silver2016alphago,silver2018alphazero,openai2019dota,vinyals2019starcraft,bansal2018emergent,baker2019emergent} and through generative adversarial networks, where one network minimizes the objective that the other maximizes~\cite{goodfellow2014gan}.
In robotics, adversarial training improves grasping and control policies~\cite{pinto2017rarl,pinto2017supervision,dennis2020paired}, while red-teaming methods audit robotic foundation models for failures that can guide fine-tuning~\cite{karnik2024embodied,srikanth2026qdig,zhang2026redvla,zhao2026qdeg,wang2024vulnerabilities,wang2025vlatest}.
Curriculum learning methods instead use a teacher to select tasks, often from easy to hard~\cite{bengio2009curriculum,narvekar2020curriculum,portelas2020acl,sukhbaatar2018asymmetric,openai2021asymmetric,florensa2018goalgan,matiisen2019teacher,wang2019poet,jiang2021plr,parkerholder2022accel,rutherford2024noregrets,liang2024eurekaverse,faldor2025omniepic,ryu2025curricullm,zala2024envgen,wang2023voyager,wang2024gensim,wang2024robogen}.
However, the policy in these works is usually a neural network, so the opponent and the teacher cannot read how its output is produced.
In contrast, \method{} brings the opponent and the teacher to robotic harnesses, where the policy is code, so Alice can read how its output is produced and propose variations or tasks accordingly.

\section{Method}
\label{sec:method}

\subsection{Problem Setup and Overview}
\label{subsec:method:overview}

\minisection{Motivation} We aim to produce frozen code policies that are robust to perturbations and generalize to related tasks without revision during deployment. \method{} introduces Alice, who reads Bob’s policy code and execution traces to propose additional scenes and tasks on which Bob refines its policy.

\minisection{Problem Setup} In a robotic harness, a foundation model receives the language instruction $l$ of a task and generates a code policy $\pi$ that calls perception and control tools. These tools are functions that the harness provides: a perception tool processes the camera images (e.g., to locate an object), and a control tool moves the robot (e.g., the gripper to a pose). Following~\cite{liberopro2025}, a task $\mathcal{T} = (l, e, G)$ consists of three parts: (1)~an instruction $l$ that states what the robot must do; (2)~a scene $e$ that holds the objects, their initial positions, and the visual context (e.g., the camera view, the lighting, and the background); and (3)~a success condition $G$ that returns 1 if the scene satisfies the instruction after execution (e.g., the bowl is on the plate), and 0 otherwise.
Within a task, only the scene $e$ varies, while the success condition $G$ stays the same~\cite{liu2023libero,liberopro2025,liberomax}. When the policy $\pi$ is run in a scene, each run produces an execution trace $z$ that stores calls to the perception and control tools and their outputs.

\minisection{The Alice--Bob Cycle} \method{} involves two players: Bob, an existing harness (e.g., ASPIRE~\cite{lu2026aspire}), and Alice, a foundation model that specifies a scene, an instruction, and a success condition. These three components form a proposal on which Bob refines the policy. Once Bob has generated the policy and refined it on the original task scenes, the two players interact in cycles (Algorithm~\ref{alg:method}). Each cycle \(t\) starts by running the current policy \(\pi\) on the original task scenes \(\mathcal{E}\) and the scenes previously proposed by Alice \(\mathcal{E}'\), which produces the execution traces \(\mathcal{Z}\). Alice then uses the policy \(\pi\) and these traces to build a set of candidate proposals \(\mathcal{C}_t\). Alice selects one of these proposals, \(p_t=(l_t,e_t,G_t)\), for Bob’s next refinement. Bob first runs the current policy on this proposal, then refines it and returns the candidate policy \(\pi'\).

After refinement, \(\pi'\) is evaluated on the same scenes as \(\pi\). Each scene is evaluated with its associated instruction and success condition. Finally, the candidate policy \(\pi'\) replaces \(\pi\) only if it satisfies two conditions: (i) \(\pi'\) must have a higher success rate than \(\pi\) on the proposal, and (ii) \(\pi'\) must not have a lower success rate than \(\pi\) on any original task scene or Alice-proposed scene. At the end of the cycle, the selected proposal \(p_t\), including its instruction and success condition, is added to \(\mathcal{E}'\). Success rate is the fraction of runs in which the success condition returns 1 (Appendix~\ref{sec:appendix:success}). After \(N\) cycles, the final policy is frozen and run on unseen scenes with no further refinement.

\begin{algorithm}[t!]
\caption{The cycle of \method{}}
\label{alg:method}
\small
\begin{algorithmic}[1]
\Require Bob; instruction $l$; original task scenes $\mathcal{E}$; $N$ cycles
\Ensure frozen policy $\pi$
\State $\pi \gets \textsc{Refine}(\textsc{Generate}(l), \mathcal{E})$
    \Comment{Bob's policy}
\State $\mathcal{E}' \gets \emptyset$
    \Comment{no stored proposal yet}
\For{$t = 1 \dots N$}
    \State $\mathcal{Z} \gets \textsc{Run}(\pi, \mathcal{E} \cup \mathcal{E}')$
        \Comment{execution traces}
    \State $\mathcal{C}_t \gets \textsc{Build}(\pi, \mathcal{Z})$
        \Comment{\textbf{Step 1:} candidate proposals}
    \State $p_t \gets \textsc{Select}(\pi, \mathcal{C}_t)$
        \Comment{\textbf{Step 2:} $p_t = (l_t, e_t, G_t)$}
    \State $\pi' \gets \textsc{Refine}(\pi, p_t)$
        \Comment{\textbf{Step 3:} candidate policy}
    \If{$\pi'$ passes the two checks}
        \State $\pi \gets \pi'$
            \Comment{\textbf{Step 4:} $\pi'$ is kept}
    \EndIf
    \State $\mathcal{E}' \gets \mathcal{E}' \cup \{p_t\}$
\EndFor
\State \Return $\pi$
    \Comment{frozen, with no further refinement}
\end{algorithmic}
\end{algorithm}

\minisection{Alice's Roles and \method{}-Bob} Adversarial Alice proposes scene variations that preserve the original task’s success condition, while cooperative Alice proposes related tasks that share many sub-goals with the original task. How a harness represents decisions in its policy affects how easily Alice can identify weaknesses and Bob can revise the relevant behavior. We therefore propose \method{}-Bob (Sec.~\ref{subsec:method:structured}), a harness that generates structured policies with physically grounded decision conditions and corresponding response strategies. Alice can then propose scenes or tasks that test individual decisions in the policy, and Bob can focus revisions on those decisions.

\subsection{Adversarial and Cooperative Alice}
\label{subsec:method:alice}

Alice proposes scene variations and related tasks to make Bob's policy more robust and reusable across tasks.
The related tasks form a set $\mathcal{P}$ and share most of their sub-goals with the original task $\mathcal{T}$.
A stage is one part of a task, such as finding, grasping, moving, or placing an object.
Alice works in three steps: reading the policy and execution traces to find a weakness, building and testing candidates, and using earlier results to guide later proposals.

\minisection{Finding a Weakness in the Current Policy}
To find a weakness, Alice reads the policy and its execution traces \(\mathcal{Z}\) (Sec.~\ref{subsec:method:overview}).
In many two-player games, a player sees only the opponent's moves, not the policy behind them.
Bob's policy, in contrast, is code, so Alice can inspect it directly for a weakness.
This access to Bob's policy is a key advantage over adversarial training and curriculum learning, where the policy is usually a neural network whose weaknesses are often inferred from its actions.
When a run fails, Alice examines the call where it went wrong and the policy lines that may explain the failure.
When every run succeeds, Alice looks for code that could fail in a new scene or task, such as a step that always picks the same object.
To investigate a possible perception error, Alice uses a tool that builds a scene, runs only the policy's perception calls, and returns their outputs.
Alice associates each suspected weakness with the relevant policy lines and the failure they may cause, and then decides what to change in the scene or task to trigger this failure.

\minisection{Scene and Task Proposals}
The physical world contains many variations, such as new camera views or object positions, and no set of scenes used during refinement can cover them all.
Policies refined on the original task scenes often fail when the scene changes significantly~\cite{liberopro2025,liberoplus,liberomax,rho2026}.
Adversarial Alice therefore proposes the variations most likely to break the current policy, so that Bob makes the policy robust to them.
Each variation $\mathcal{T}'=(l',e',G)$ keeps the success condition $G$ of the original task, while the scene $e'$ differs from $e$.
Alice builds each variation with scene functions, a set of tools that sets or changes one part of a scene.
Alice applies these functions to an original task scene and chooses their parameters within given ranges to trigger a suspected weakness (Table~\ref{tab:scene_changes}).
Each change is either static, applied before the run starts, or dynamic, triggered once during the run without any action of the robot.
For example, Alice can move an object during execution if the policy locates objects only at the start.

In manipulation, many tasks share most of their sub-goals (e.g., finding, grasping, moving, or placing an object).
Existing harnesses still generate and refine a separate policy for each of these tasks, which repeats model calls and execution trials.
To reduce this repeated work, cooperative Alice proposes related tasks that the current policy cannot yet solve.
Bob extends the same policy to cover them, reusing the sub-goals that it already handles.
Each new task \(\mathcal{T}^{+}=(l^{+},e^{+},G^{+})\) has its own instruction \(l^{+}\) and success condition \(G^{+}\).
To build a candidate task, Alice calls scene functions on an empty scene, where each call sets one part of the task, such as the manipulated object or the target (Table~\ref{tab:scene_changes}).
The policy receives \(l^{+}\) at the start of each run, so it knows which task to perform.

To select the proposal for Bob's next refinement, Alice first runs the current policy on each candidate in \(\mathcal{C}_t\).
Candidates that the policy already solves need no refinement, so Alice selects a failing candidate.
If the policy succeeds on every candidate, the cycle ends without refinement.
For a cooperative task, failure alone is not enough, since the task may need more changes than Bob can make with a limited number of attempts per cycle.
Alice therefore prefers the failing task with the smallest estimated code change, judged from the stage at which the policy failed and a summary of the final object positions.

\minisection{Using Results from Earlier Proposals}
Alice's objective is to choose proposals that improve Bob's policy through refinement.
We therefore score the improvements achieved on earlier proposals and provide these scores in Alice's prompt to guide later selections without updating Alice's model weights.
In the first cycle, no scores are available, so Alice selects using only the suspected weaknesses and the policy's runs on the candidates.

For adversarial proposals, Alice and Bob have different scores, but both gain when the policy's success rate improves, so the game is general-sum.
Bob's score is the success rate of the policy on the new variation.
Alice's score rewards variations on which the policy initially fails and subsequently improves:
\begin{equation}
U = (1 - s_0) \cdot \max(0,\; s_{\mathrm{end}} - s_0),
\label{eq:alice_score}
\end{equation}
where \(s_0\) is the success rate of the policy on the variation before Bob refines on it, and \(s_{\mathrm{end}}\) is its success rate on the same variation two cycles later.
Alice earns a positive score only if the success rate improves.
If two refined variations of the same change type earn almost no score, Alice can no longer propose that type (Appendix~\ref{app:refinement_cycle}).

For cooperative proposals, Alice and Bob share one score, so the game is common-payoff.
The score sums the success-rate gains across the proposed tasks:
\begin{equation}
U^{+} = \sum_{k} \max(0,\; s_k - s_{0,k}),
\label{eq:shared_score}
\end{equation}
where \(s_{0,k}\) is the success rate of the policy on the \(k\)-th proposed task before Bob refines on it, and \(s_k\) is the success rate of the current policy on the same task.
We re-measure \(s_k\) after every cycle, so the score reflects the gains the current policy retains across proposed tasks.

\subsection{\method{}-Bob for Alice}
\label{subsec:method:structured}

    A policy uses checks of the robot’s state and surroundings to decide what to do next. We design \method{}-Bob to write these checks as explicit if-else decisions, so that Alice can read the decisions in the code and target them. Each policy is a sequence of actions, such as acquiring and placing an object, and each action contains its if-else decisions. Each branch specifies what the robot does next based on observations of the physical environment. For example, after a lift, the policy checks whether the bottle rose; if not, it opens the gripper, retreats, and retries with a different grasp depth.

While Alice can inspect code policies produced by any harness, this structure aims to make inspection more efficient by exposing which physical situations select which actions. For example, if a recovery rule retries a failed grasp from the same direction, adversarial Alice could propose a placement where that approach is obstructed. The rule provides a direct link between a potential weakness and a scenario that tests it. Cooperative Alice could read these decisions to propose related tasks that reuse existing actions and need few code changes.

\section{Experimental Results}

\begin{table*}[t]
    \centering
    \caption{\textbf{Main results.} Success rates (\%) with Sonnet 5.5, before (shaded) and after training with Alice. \textit{Adversarial}: refined in LIBERO over the 40 tasks of the four suites and evaluated on LIBERO-Pro, Plus, and MAX, whose scenes are unseen during refinement. ASPIRE reports LIBERO-Pro results after refining on the perturbed tasks themselves, and no harness reports Plus or MAX, so we run all baselines ourselves. Results are macro-averaged over tasks; Overall averages Pro, Plus, and MAX, where MAX averages Base and Dynamic. \textit{Cooperative}: one policy per suite, refined on task 0 (shaded) and then extended with Alice, evaluated on the other nine tasks.}
    \label{tab:main-results}
    \label{tab:harness-performance}
    \label{tab:cooperative-main}
    \begingroup
    \footnotesize
    \setlength{\tabcolsep}{3pt}
    \renewcommand{\arraystretch}{1.05}
    \begin{tabularx}{\textwidth}{ll*{9}{>{\centering\arraybackslash}X}}
        \toprule
        & & \multicolumn{5}{c}{\textbf{Adversarial Alice} (robustness)}
          & \multicolumn{4}{c}{\textbf{Cooperative Alice} (other nine tasks)} \\
        \cmidrule(lr){3-7}\cmidrule(lr){8-11}
        \textbf{Harness} & \textbf{Policy}
        & PRO & PLUS & MAX Base & MAX Dyn. & Overall
        & Spatial & Object & Goal & LIBERO-10 \\
        \midrule
        \rowcolor{black!6}
        \textbf{ASPIRE} & Bob only
        & 52.0 & 72.9 & 66.2 & 48.2 & 60.7
        & 40.0 & 10.4 & 0.0 & 0.0 \\
        & + Alice
        & \textbf{72.6} & \textbf{90.5} & \textbf{84.6}
        & \textbf{72.6} & \textbf{80.6}
        & \textbf{91.9} & \textbf{68.9} & \textbf{60.7} & \textbf{57.0} \\
        \rowcolor{black!6}
        \textbf{RHO} & Bob only
        & 39.7 & 57.7 & 53.4 & 37.5 & 47.6
        & 23.0 & 0.0 & 0.0 & 0.0 \\
        & + Alice
        & \textbf{55.7} & \textbf{71.5} & \textbf{68.4}
        & \textbf{52.8} & \textbf{62.6}
        & \textbf{80.0} & \textbf{62.2} & \textbf{54.8} & \textbf{47.4} \\
        \rowcolor{black!6}
        \textbf{GaP} & Bob only
        & 40.1 & 56.1 & 50.0 & 39.6 & 47.0
        & 50.4 & 11.9 & 0.0 & 0.0 \\
        & + Alice
        & \textbf{52.6} & \textbf{69.8} & \textbf{64.7}
        & \textbf{54.3} & \textbf{60.7}
        & \textbf{80.0} & \textbf{67.4} & \textbf{10.4} & \textbf{17.8} \\
        \rowcolor{black!6}
        \textbf{\method{}-Bob} & Bob only
        & 55.3 & 77.2 & 71.1 & 57.1 & 65.5
        & 40.7 & 11.1 & 0.0 & 0.0 \\
        & + Alice
        & \textbf{76.9} & \textbf{90.5} & \textbf{86.9}
        & \textbf{77.8} & \textbf{83.2}
        & \textbf{91.1} & \textbf{84.4} & \textbf{71.1} & \textbf{60.0} \\
        \bottomrule
    \end{tabularx}
    \endgroup
\end{table*}
\label{sec:evaluation}

We apply \method{} to four harnesses, ASPIRE~\cite{lu2026aspire}, RHO~\cite{rho2026}, GaP~\cite{chen2026gap}, and our \method{}-Bob, and test whether adversarial Alice makes frozen code policies robust to held-out scene perturbations (Sec.~\ref{subsec:eval:adversarial}) and whether cooperative Alice enables one policy to generalize to related tasks excluded from refinement (Sec.~\ref{subsec:eval:cooperative}).
We also compare Alice with environment-design and curriculum baselines (Sec.~\ref{subsec:eval:ablation}) and show that a cheaper model with Alice outperforms a frontier model alone at one-third the cost (Table~\ref{tab:frontier-model}).
Finally, our refined policies transfer zero-shot from simulation to a physical robot, remaining robust to real-world perturbations and generalizing to new tasks (Sec.~\ref{subsec:eval:real}).

\subsection{Experiment Details}

\begin{table*}[t]
    \begin{minipage}[t]{\columnwidth}
        \caption{\textbf{Ablation for Adversarial Alice.} Success rates (\%) with ASPIRE and Sonnet 5.5, macro-averaged over 12 tasks (tasks 0--2 from each of the four LIBERO suites). All rows start from the same policies of Bob.}
        \label{tab:baselines-ablations}
    \end{minipage}\hfill
    \begin{minipage}[t]{\columnwidth}
        \caption{\textbf{Ablation for Cooperative Alice.} Success rates (\%) with ASPIRE and Sonnet 5.5 on the other nine tasks of each LIBERO suite. All rows start from the policy of task 0 and differ only in what proposes the next task.}
        \label{tab:cooperative-curricula}
    \end{minipage}

    \begin{minipage}[t]{\columnwidth}
        \centering
        \footnotesize
        \setlength{\tabcolsep}{3pt}
        \renewcommand{\arraystretch}{1.05}
        \begin{tabularx}{\linewidth}{@{}l*{4}{>{\centering\arraybackslash}X}@{}}
            \toprule
            \textbf{Training condition} & PRO & PLUS & \shortstack{MAX\\Base} & \shortstack{MAX\\Dyn.} \\
            \midrule
            Bob only
            & 55.0 & 78.0 & 77.2 & 59.3 \\
            Bob + uniform random
            & 55.7 & 77.8 & 72.5 & 61.2 \\
            Bob + SFL~\cite{rutherford2024noregrets}
            & 53.7 & 77.2 & 74.8 & 58.0 \\
            Bob + Eurekaverse~\cite{liang2024eurekaverse}
            & 55.8 & 74.7 & 77.9 & 61.0 \\
            Bob + trajectory-only Alice
            & 56.7 & 78.9 & 79.3 & 66.8 \\
            \textbf{Bob + full Alice (ours)}
            & \textbf{72.8} & \textbf{91.1} & \textbf{86.6} & \textbf{76.4} \\
            \bottomrule
        \end{tabularx}
    \end{minipage}\hfill
    \begin{minipage}[t]{\columnwidth}
        \centering
        \footnotesize
        \setlength{\tabcolsep}{3pt}
        \renewcommand{\arraystretch}{1.05}
        \begin{tabularx}{\linewidth}{@{}l*{3}{>{\centering\arraybackslash}X}c@{}}
            \toprule
            \textbf{Next task proposed by} & \shortstack{\phantom{X}\\Spatial} & Object & Goal & LIBERO-10 \\
            \midrule
            None (policy of task 0)
            & 40.0 & 10.4 & 0.0 & 0.0 \\
            Random
            & 53.3 & 37.8 & 14.1 & 5.9 \\
            TSCL~\cite{matiisen2019teacher}
            & 65.9 & 48.9 & 8.1 & 0.0 \\
            OMNI-EPIC~\cite{faldor2025omniepic}
            & 60.0 & 37.8 & 27.4 & 22.2 \\
            Trajectory-only Alice
            & 56.3 & 41.5 & 24.4 & 8.1 \\
            \textbf{Cooperative Alice (ours)}
            & \textbf{91.9} & \textbf{68.9} & \textbf{60.7} & \textbf{57.0} \\
            \bottomrule
        \end{tabularx}
    \end{minipage}
\end{table*}

\label{subsec:eval:setup}

\minisection{Benchmarks}
All code policies are refined in LIBERO simulation~\cite{liu2023libero}, where a robot arm follows a language instruction in a tabletop scene and succeeds when every predicate of its goal holds.
We use four of its suites, each with ten tasks: Spatial, Object, Goal, and LIBERO-10.
Robustness is measured on LIBERO-Pro, LIBERO-Plus, and LIBERO-MAX~\cite{liberopro2025,liberoplus,liberomax}, which perturb the scenes of the same LIBERO tasks. For more info, see Appendix~\ref{app:robustness_benchmarks}. Adversarial Alice builds scene variations and cooperative Alice builds new related tasks with the scene functions of Sec.~\ref{subsec:method:alice}, without access to the perturbed test scenes or the held-out tasks.

\minisection{Harnesses}
The three existing harnesses, which we use as-is, differ in how they organize a policy.
ASPIRE and RHO write free-form code: ASPIRE keeps validated repairs as reusable skills and RHO searches over multi-file repositories.
GaP builds a graph of perception and action nodes.
We also evaluate our \method{}-Bob, which writes explicit if-else decisions for Alice to inspect and target (Sec.~\ref{subsec:method:structured}).
Unless otherwise stated, Alice and Bob both use Sonnet 5.5.
We report success rates of the final frozen policies on held-out test cases.
Appendices~\ref{sec:appendix:method_details}, \ref{sec:appendix:benchmarks}, and~\ref{sec:appendix:baselines} detail the scene functions, optimization settings, test sets, baselines, and API costs.

\subsection{Adversarial Alice}
\label{subsec:eval:adversarial}
In the physical world, a frozen policy should remain robust to unseen perturbations without further optimization. We test whether the scenes Alice proposes during refinement improve the policy's performance on held-out test scenes. ASPIRE reports results on LIBERO-Pro after refining each policy on the perturbed tasks themselves~\cite{lu2026aspire}, which does not test unseen perturbations, and no harness reports results on LIBERO-Plus or LIBERO-MAX. We therefore run every baseline ourselves with refinement on the original LIBERO scenes only, so every variation is unseen.

\minisection{Experimental protocol}
For each LIBERO task, Bob first refines a policy on 15 development scenes of the original task in up to eight iterations, which gives the Bob-only baseline.
Bob then plays eight cycles with Alice (Sec.~\ref{subsec:method:alice}) and refines on each proposed scene in at most two attempts.
Alice constructs these scenes using generic scene functions (Table~\ref{tab:scene_changes}).
We evaluate the frozen policies on held-out test cases from LIBERO-Pro, LIBERO-Plus, and LIBERO-MAX (Appendix~\ref{app:robustness_benchmarks}).

\minisection{Robustness across harnesses}
Alice raises overall success by 19.9 percentage points with ASPIRE, 15.0 with RHO, 13.7 with GaP, and 17.7 with \method{}-Bob (Table~\ref{tab:harness-performance}).
On LIBERO-MAX Dynamic, where scenes change during execution, success improves by 14.7--24.4 percentage points across the four harnesses.
Alice therefore makes the policies of all four harnesses more robust to unseen variations.

\minisection{Robustness of \method{}-Bob}
\method{}-Bob makes each decision of a policy explicit, so that Alice can read and target it in the code and Bob can refine it alone.
Across all 40 tasks of LIBERO, \method{}-Bob achieves the highest overall success among the four harnesses, both alone (65.5\%) and with Alice (83.2\%).

\begin{table}[t]
    \centering
    \caption{\textbf{A frontier model as Bob alone.} Success rates (\%) with ASPIRE as Bob’s harness, evaluated on the robustness test sets for the 30 tasks of Spatial, Object, and Goal.
    Fable 5.2 (Bob only) receives additional robustness guidance; Sonnet 5.5 is evaluated as Bob only and Bob + Alice.
    Results are macro-averaged over tasks.
    Cost is in USD per task.}
    \label{tab:frontier-model}
    \begingroup
    \footnotesize
    \setlength{\tabcolsep}{2.5pt}
    \renewcommand{\arraystretch}{1.1}
    \begin{tabularx}{\columnwidth}{@{}ll*{4}{>{\centering\arraybackslash}X}cc@{}}
        \toprule
        \textbf{Model} & \textbf{Setup} & PRO & PLUS & \shortstack{MAX\\Base} & \shortstack{MAX\\Dyn.} & Overall & \shortstack{Cost\\(\$)} \\
        \midrule
        Fable 5.2 & Bob only & 62.5 & 78.5 & 76.9 & 67.9 & 71.1 & 22.82 \\
        \midrule
        \rowcolor{black!6}
        Sonnet 5.5 & Bob only & 55.0 & 73.1 & 67.6 & 50.1 & 62.3 & \\
        & + Alice & \textbf{74.7} & \textbf{92.3} & \textbf{85.6} & \textbf{76.2} & \textbf{82.6} & 7.62 \\
        \bottomrule
    \end{tabularx}
    \endgroup
\end{table}

\minisection{Comparison with Frontier Models}
\method{} enables Sonnet 5.5 to outperform the frontier model Fable 5.2 at a total API cost of \$7.62 per task (Table~\ref{tab:frontier-model}).
Fable 5.2 runs ASPIRE as Bob alone.
For a fairer comparison, we augment the Bob prompt of Fable 5.2 with the generic perturbation guidelines (Appendix~\ref{sec:appendix:frontier-prompt}) provided to Alice and explicitly request robust behavior.
On the 30 tasks of Spatial, Object, and Goal, Alice raises Sonnet 5.5's overall success by 20.3 percentage points, enabling it to outperform Fable 5.2 on every benchmark.
Structuring refinement with \method{} thus improves robustness more than switching to a frontier model.

\subsection{Cooperative Alice}
\label{subsec:eval:cooperative}
We consider a deployment setting in which the task family is known but the future requests are not.
For example, a robot may be expected to place objects into containers, with the specific object and container chosen only when a request arrives.
A reusable code policy should handle these related requests without a separate refinement for each one.
Starting from a policy refined on an initial task, we test whether Alice can propose practice tasks that improve Bob's performance on related tasks excluded from refinement.

\minisection{Experimental protocol}
We call the ten tasks in each LIBERO suite a task family.
They share sub-goals, such as grasping or placing objects, but differ in the objects or targets (Appendix~\ref{app:libero_suites}).
For each suite, Bob refines a policy on 15 development scenes of task 0, which gives the Bob-only baseline.
Then Bob plays up to eight cycles with Alice for Spatial and Object and sixteen for Goal and LIBERO-10 (Sec.~\ref{subsec:method:alice}), refining the same policy to handle task 0 and Alice's proposed practice tasks.
Alice builds these tasks from generic task-construction tools (Table~\ref{tab:scene_changes}), without access to the nine test tasks or their results.
We evaluate the frozen policy on the other nine tasks of each suite, with 15 initial scenes per task, where it receives each task's instruction and runs without further refinement.

\minisection{Task generalization across harnesses}
We show that cooperative Alice can be applied to different harnesses without changing how they write or refine policies.
For each of the four harnesses, we compare the policy refined with cooperative Alice against the Bob-only baseline on the other nine tasks of each suite.
Alice raises success for every harness in Spatial, Object, and Goal (Table~\ref{tab:cooperative-main}).
For example, with ASPIRE, success rises by 51.9 percentage points in Spatial, 58.5 in Object, and 60.7 in Goal, where the policy of task 0 solves none of the other tasks.
A single refined policy therefore generalizes to related tasks excluded from refinement.

\minisection{Long-horizon tasks}
Long-horizon tasks chain several sub-tasks, each a short-horizon task like those of Spatial, Object, and Goal, so a policy must also generalize to new combinations of sub-tasks.
In LIBERO-10, the policy is refined on task 0, which places two objects in a basket, and tested on the other nine tasks, which combine different sub-tasks.
Alice proposes practice tasks with one or several sub-tasks.
When a task with several sub-tasks fails, the execution trace shows which already succeed, so Alice can propose a task with only the failing sub-task before returning to the combined task.
Starting from 0.0\%, success rises by 57.0 percentage points with ASPIRE, 47.4 with RHO, 17.8 with GaP, and 60.0 with \method{}-Bob.
Alice's practice tasks therefore also improve generalization to long-horizon tasks.

\minisection{Task generalization with \method{}-Bob}
\method{}-Bob writes decisions that can be reused for other objects and positions, so a policy can apply them to related tasks (Sec.~\ref{subsec:method:structured}).
Cooperative Alice can also compare candidate tasks with these decisions and propose the one that needs the fewest new or changed decisions, so that Bob can solve it within one cycle.
This makes refinement more efficient: \method{}-Bob exceeds ASPIRE, the best existing harness, by 10.4 percentage points in Goal, by 15.5 in Object, and by 3.0 in LIBERO-10.

\subsection{Ablations}
\label{subsec:eval:ablation}

\minisection{Ablation for Adversarial Alice}
We first ablate adversarial Alice on 12 tasks, using tasks 0--2 from each of Spatial, Object, Goal, and LIBERO-10 (Table~\ref{tab:baselines-ablations}).
We use three ablations: Bob alone, random scenes drawn uniformly from the scene functions, and a trajectory-only Alice that sees execution traces but not the code.
Relative to Bob alone, overall success rises by 14.8 percentage points with full Alice and 2.5 with trajectory-only Alice, while it decreases by 0.3 with random scenes.
Unlike random sampling and trajectory-only Alice, full Alice reads the code behind the policy's actions and proposes adversarial scene variations that expose weaknesses for Bob to address.

We further compare Alice with two existing environment-design methods that propose scenes to Bob: SFL~\cite{rutherford2024noregrets} keeps random scenes on which Bob succeeds about half of the time, and Eurekaverse~\cite{liang2024eurekaverse} asks a language model to evolve scenes from Bob's success rates.
See Appendix~\ref{app:baseline_proposers} for details.
Overall success with Alice is 16.1 percentage points higher than with SFL and 15.2 higher than with Eurekaverse.
Unlike SFL and Eurekaverse, which choose scenes from execution outcomes, \method{} gives adversarial Alice access to the policy code.
The code reveals how the policy chooses actions, so Alice can identify decisions that may fail and propose scenes that test them, giving Bob failures to address during refinement.

\minisection{Ablation for Cooperative Alice}
For cooperative Alice, we use random-task and trajectory-only ablations on the four suites (Table~\ref{tab:cooperative-curricula}) and compare with two curriculum methods, TSCL~\cite{matiisen2019teacher} and OMNI-EPIC~\cite{faldor2025omniepic}, which choose tasks from Bob's past results (Appendix~\ref{app:baseline_proposers}). Alice improves average success by 42 percent over random tasks, 37 over trajectory-only Alice, 39 over TSCL, and 33 over OMNI-EPIC. Unlike these prior methods, Cooperative Alice refers to Bob's code policy to choose related tasks that build on what it already handles, so Bob can extend the same policy across tasks.

\subsection{Simulation-to-Real Transfer}
\begin{table}[t]
    \centering
    \caption{\textbf{Simulation-to-real transfer.} Successes out of 20 trials on a Franka arm, for the policy of Bob alone and the policy after the cycles with Alice. Every policy is refined in simulation only and runs frozen on the robot. \textit{Adversarial}: the task on its original scene and under two changes of the scene. \textit{Cooperative}: tasks that differ from the first task, picking the can and placing it in the box, in the object, in the target, or in both.}
    \label{tab:real-world}
    \begingroup
    \small
    \setlength{\tabcolsep}{4pt}
    \renewcommand{\arraystretch}{1.12}
    \begin{tabularx}{\columnwidth}{@{}l>{\raggedright\arraybackslash}Xcc@{}}
        \toprule
        \textbf{Task} & \textbf{Condition} & \textbf{Bob only} & \textbf{+ Alice} \\
        \midrule
        \multicolumn{4}{@{}l}{\textit{Adversarial Alice}} \\
        Can in box & Original task & 8/20 & \textbf{20/20} \\
        & Occluded & 0/20 & \textbf{14/20} \\
        & Moving target & 0/20 & \textbf{20/20} \\
        \midrule
        \multicolumn{4}{@{}l}{\textit{Cooperative Alice}} \\
        Different object & Strawberry in box & 0/20 & \textbf{19/20} \\
        Different target & Can in a bowl & 0/20 & \textbf{20/20} \\
        \addlinespace[2pt]
        Different target/object & Cup in a bowl & 0/20 & \textbf{20/20} \\
        Different target/object & Cup in another cup & 0/20 & \textbf{15/20} \\
        \bottomrule
    \end{tabularx}
    \endgroup
\end{table}

\begin{figure}[t]
    \centering
    \includegraphics[width=\columnwidth]{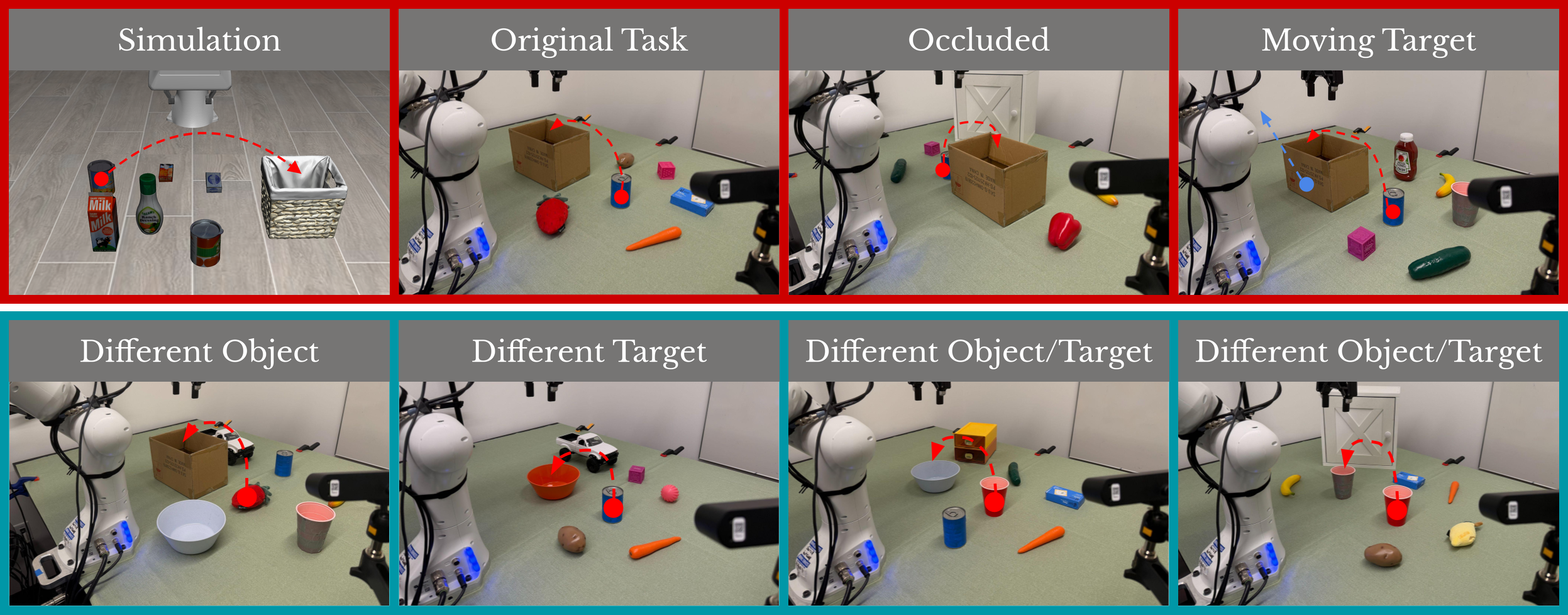}
    \caption{\textbf{From simulation to the real world.} Every policy starts from task 0 of LIBERO-Object in simulation (first image) and runs on the Franka arm under the conditions of Table~\ref{tab:real-world}. The red dot marks the object, and the dashed red arrow points to its target. The blue arrow shows the box moving during execution. Red frames: adversarial Alice. Teal frames: cooperative Alice.}
    \label{fig:real-world}
\end{figure}

\label{subsec:eval:real}

Refining policies in simulation can reduce the need for physical trials, but this benefit depends on sim-to-real transfer.
The frozen policy must handle new tasks and variations in object appearance and location, camera views, and backgrounds.
In this paper, we show that policies refined with \method{} can remain robust to variations and generalize across tasks in the real world.
In our scenario, Bob refines every policy only in simulation, starting from task 0 of LIBERO-Object, and the frozen policy then runs on a Franka arm with a scene camera and a wrist camera (Figure~\ref{fig:real-world}). For more info, see Appendix~\ref{sec:appendix:hardware_setup}.
Each experiment has 20 trials, and a trial succeeds when the object rests in its target at the end (Table~\ref{tab:real-world}).

\minisection{Adversarial Alice in the real world}
We test whether \method{} produces code policies that are more robust to perturbations in the real world. Bob refines the policy, alone or with Alice, only in simulation on LIBERO-Object task 0 (Appendix~\ref{sec:appendix:real}). At test time, we introduce a static perturbation, occluding the manipulated object from the scene camera, and a dynamic perturbation, moving the target object mid-execution. Compared with Bob alone, refining with Alice raises success from 40\% to 100\% on the original task and from 0\% to 85\% under the two perturbations. These results show that the robustness Alice adds in simulation also holds in the real world.

\minisection{Cooperative Alice in the real world}
We test whether a single code policy refined with \method{} in simulation generalizes to related tasks on a real robot.
As in the previous experiment, Bob starts from the policy of LIBERO-Object task 0, and cooperative Alice proposes related tasks that change the manipulated object, the target, or their positions, so that Bob extends the same policy to all of them (Appendix~\ref{sec:appendix:real}).
On the robot, we test four new tasks: a strawberry placed in a box (new object), a can placed in a bowl (new target), and a cup placed in a bowl or stacked in another cup (both).
Compared with Bob alone, refining with Alice raises average success across the four tasks from 0\% to 92.5\%.
These results show that the generalization Alice adds in simulation also holds in the real world.

\section{Conclusion}
\label{sec:conclusion}

In this work, we show that frozen code policies can become robust to unseen variations or reusable across related tasks without modifying their harness. By reading the policy of a harness, Bob, and proposing the variations or tasks for its next refinement, an additional agent, Alice, outperforms environment-design and curriculum methods that choose from past outcomes alone. 
As an adversary, Alice raises overall success by 14 to 20 percent across different harnesses and lets a cheaper model outperform a frontier model at a third of its cost; as a collaborator, Alice extends one policy to the other tasks of its suite, raising success by up to 73.3 points. We further show that these policies transfer from simulation to a physical robot, where success rises from 0\% to 85\% under scene changes and to 92.5\% on new tasks. Overall, our results suggest that reading a policy's code is a practical way to decide what a robot should practice.

\minisection{Limitations} \method{} has three limitations. First, Alice needs guidelines to propose scene variations and related tasks. Second, the simulator’s objects and dynamics further limit which proposals Alice can build. Finally, the policy and Alice both rely on the harness’s perception and control tools. Refinement does not modify these tools, and undetected tool errors can mislead Alice.

{
    \small
    \bibliographystyle{ieeenat_fullname}
    \bibliography{main}
}

\clearpage
\appendix
\setcounter{dbltopnumber}{4}
\renewcommand{\dbltopfraction}{0.95}
\renewcommand{\dblfloatpagefraction}{0.8}
\setcounter{topnumber}{4}
\setcounter{totalnumber}{8}
\renewcommand{\topfraction}{0.95}
\renewcommand{\textfraction}{0.05}
\renewcommand{\floatpagefraction}{0.8}
\makeatletter
\setlength{\@dblfptop}{0pt}
\setlength{\@dblfpbot}{0pt plus 1fil}
\setlength{\@fptop}{0pt}
\setlength{\@fpbot}{0pt plus 1fil}
\setlength{\@dblfpsep}{\dblfloatsep}
\setlength{\@fpsep}{\floatsep}
\makeatother
\onecolumn
\begin{center}
    \Large
    \textbf{\thetitle}\\
    \vspace{0.5em}Appendix
\end{center}
\vspace{1.0em}

Here, we provide additional method details (Sec.~\ref{sec:appendix:method_details}), further experimental results (Sec.~\ref{sec:appendix:additional_exp}), and details on the benchmarks (Sec.~\ref{sec:appendix:benchmarks}), the baselines (Sec.~\ref{sec:appendix:baselines}), and our hardware setup (Sec.~\ref{sec:appendix:hardware_setup}).

\titlecontents{section}[0pt]{\addvspace{12pt}\bfseries}{\thecontentslabel\hspace{1em}}{}{\hfill\contentspage}
\titlecontents{subsection}[3.4em]{\addvspace{3pt}\rightskip=1.55em plus 4em}{\hspace*{-2.4em}\thecontentslabel\hspace{1em}}{}{\titlerule*[0.6pc]{.}\contentspage}
\startcontents[appendix]
{\hypersetup{linkcolor=black}\printcontents[appendix]{}{1}{\section*{Contents}\setcounter{tocdepth}{2}}}
\twocolumn

\section{Additional Method Details}
\label{sec:appendix:method_details}

\subsection{Tools of Alice}
\label{sec:appendix:tools}
\label{sec:appendix:scene-functions}
Table~\ref{tab:alice-tools} lists the tools that Alice calls in each cycle (Sec.~\ref{subsec:method:alice}).
Alice writes every scene and task as a list of calls to scene functions.
Table~\ref{tab:task_parts} lists what Alice can change with them, and Table~\ref{tab:scene-function-examples} gives representative examples.
\begin{table*}[t]
    \centering
    \caption{\textbf{Tools of Alice.} In each cycle, Alice calls these tools in a session of limited length before proposing. The tools read Bob's policy and its recorded executions, query the perception model that Bob's policy calls, and run the policy in the simulator. No tool returns the state of the simulator. Both roles of Alice use the same tools; only the adversary runs a scene without its dynamic perturbation.}
    \label{tab:alice-tools}
    \begingroup
    \footnotesize
    \setlength{\tabcolsep}{5pt}
    \renewcommand{\arraystretch}{1.08}
    \begin{tabularx}{\textwidth}{@{}>{\raggedright\arraybackslash}p{0.30\textwidth}>{\raggedright\arraybackslash}X@{}}
        \toprule
        \textbf{Tool and parameters} & \textbf{Description} \\
        \midrule
        \texttt{read\_code(start\_line, end\_line)} & Numbered lines of Bob's current policy. \\
        \texttt{inspect\_episode(episode\_id, keyframes)} & One recorded execution of the current policy: the calls it executed and their outcomes, the outputs of perception, policy errors, the point of failure, and the camera images at chosen moments. \\
        \texttt{perceive(scene, queries)} & Builds a scene that Alice writes and runs Bob's own perception call on its first camera images, for text queries that Alice chooses. Returns the top detections with their scores and boxes for each camera, and the images. The policy is not run. \\
        \texttt{validate\_scene(scene)} & Checks a scene that Alice writes before it is built, and rejects one that cannot be built, such as one with overlapping objects, with the reason why it is invalid. \\
        \texttt{probe(scenes, without\_event)} & Runs Bob's current policy in the simulator on scenes that Alice writes, in one parallel batch. Returns the outcome, the executed calls, the point of failure, and images. With \texttt{without\_event}, the same scene runs without its dynamic perturbation. \\
        \texttt{screen(scenes)} & Runs the current policy on the instances of each scene that {\method{}} uses to accept a proposal. Returns an identifier that a proposal cites and, for each failed execution, how far it got. \\
        \texttt{submit\_proposals(cards)} & Up to three ranked proposals, each with its scene or task and the screening result that supports it. Ends the session. \\
        \bottomrule
    \end{tabularx}
    \endgroup
\end{table*}

\begin{table}[t]
    \centering
    \caption{\textbf{What Alice can change.} A change is static, holding for the whole execution, or dynamic, happening once during the execution without any action of the robot.}
    \label{tab:task_parts}
    \label{tab:scene_changes}
    \begingroup
    \footnotesize
    \setlength{\tabcolsep}{5pt}
    \begin{tabularx}{\columnwidth}{@{}>{\raggedright\arraybackslash}p{0.22\columnwidth}>{\raggedright\arraybackslash}p{0.36\columnwidth}>{\raggedright\arraybackslash}X@{}}
        \toprule
        \textbf{Alice can change} & \textbf{Options} & \textbf{Examples} \\
        \midrule
        Camera & Where the camera is placed and how clean its images are \\
        Look of the scene & The light, the surfaces, the colours, the sizes, or shape of objects & A thematic change in the scene \\
        Object to move & Which object the robot moves, and its size & A mug, a smaller cube \\
        Start poses and Final poses & Where the objects and the arm begin and end & Inside a drawer, an item that slides during the run \\
        Object state & Doors, drawers, and knobs at the start and at the end & A drawer open or closed \\
        Extra objects & Objects that the task does not need & A look-alike of the object to move \\
        Instruction & What the task instruction is & Different instruction for task \\
        Goal & Where the object must end: on or in an object or a fixture, or beside an object & In the basket, to the left of the basket \\
        Goal state of a fixture & A fixture that must end open or closed & Close the microwave \\
        Fixtures & Which fixtures the scene has, and where they stand & A microwave, a shelf \\
        Number of goals & One goal, or multiple reached in the order written & A cookie in the basket, then a ketchup on the plate \\
        \bottomrule
    \end{tabularx}
    \endgroup
\end{table}

\begin{table}[t]
    \centering
    \caption{\textbf{Representative scene functions.} Alice writes a scene or a task as a list of such calls; the names and parameters are simplified. A dynamic function triggers once during the execution, and a goal function sets what must hold at the end.}
    \label{tab:scene-function-examples}
    \begingroup
    \footnotesize
    \setlength{\tabcolsep}{5pt}
    \begin{tabularx}{\columnwidth}{@{}>{\raggedright\arraybackslash}p{0.48\columnwidth}>{\raggedright\arraybackslash}X@{}}
        \toprule
        \textbf{Scene function} & \textbf{What it does} \\
        \midrule
        \texttt{camera(\allowbreak azimuth,\allowbreak  elevation,\allowbreak  distance)} & Places the exterior camera elsewhere around the workspace. \\
        \texttt{light(\allowbreak brightness,\allowbreak  angle,\allowbreak  tint)} & Sets how bright the scene is, where the light comes from, and its colour. \\
        \texttt{place(\allowbreak object,\allowbreak  x,\allowbreak  y,\allowbreak  yaw)} & Puts a movable object at a pose on the table. \\
        \texttt{shift\_when(\allowbreak object,\allowbreak  dx,\allowbreak  dy)} & Dynamic. Moves an object by an offset. \\
        \texttt{goal\_on(\allowbreak object,\allowbreak  support)} & Goal. The object must end on another object or on a fixture. \\
        \bottomrule
    \end{tabularx}
    \endgroup
\end{table}

\subsection{Task Success Conditions}
\label{sec:appendix:success}
Following LIBERO~\cite{liu2023libero}, the success condition $G$ of a task (Sec.~\ref{subsec:method:overview}) is a conjunction of predicates.
A predicate is a property of one object, or a relation between two objects, that is either true or false in a given state of the scene.
The tasks of the four suites use five predicates: the relations \texttt{On} (an object rests on another object or on a fixture) and \texttt{In} (an object is inside a container), and the properties \texttt{Open}, \texttt{Close}, and \texttt{TurnOn} (a drawer or a door is open or closed, the stove is on).
Alice can also set \texttt{TurnOff} (the stove is off).
For example, \texttt{On(bowl, plate)} is true when the bowl rests on the plate.
The simulator checks these predicates, and a run succeeds if all of them hold before the step limit (1000 simulator steps, 4000 in LIBERO-10).
A variation of adversarial Alice keeps the predicates of the original task.
A task of cooperative Alice sets its own predicates with goal functions such as \texttt{goal\_on} (Table~\ref{tab:scene-function-examples}); a task of LIBERO-10 can have two goals, which must be reached in the order written.
A run that cannot succeed for a reason outside the policy, such as an object that starts off its support, is marked infeasible and removed from the count.
The success rate is the fraction of the remaining runs that succeed.

\subsection{\method{}-Bob Policy Representation}
\label{app:structured_bob}

Alice can inspect code policies from any harness.
We design \method{}-Bob to make this inspection easier: named actions identify the physical stage, explicit if-else decisions expose its conditions and responses, and execution records show what happened when those decisions were used (Sec.~\ref{subsec:method:structured}).
Together, these let Alice connect a physical failure to the relevant code and choose a scene that tests its weakness.

\minisection{Actions and if-else decisions}
A policy is one Python module with a list of actions, such as acquiring, transporting, and placing an object.
Each entry gives the action's function, its dependencies, and its physical completion condition.
Each action takes a shared context and returns whether that condition holds; a coordinator calls the actions in order.
Within an action, each if-else decision states a check of the robot or its surroundings and the response to that check.
Every retry is bounded, and Bob's refinements preserve this structure.

Figure~\ref{fig:rallybob-example} shows a cooperative LIBERO-Spatial policy.
Its acquisition code selects the bowl on a detected landmark; if the landmark is not found, it selects the highest bowl.
Alice can therefore inspect both the condition and the fallback's assumption that the target bowl is the highest one.
The named actions also show which later behaviors, such as transport and placement, a related task could reuse.

\minisection{Execution traces}
Each run records perception and control calls and their results, robot states, commands, and keyframe images.
The policy logs action transitions and reports its measurements and decisions.
The action names and branch labels connect these records to the code: Alice can locate the stage that failed and inspect its checks and responses.
In Figure~\ref{fig:rallybob-example}, the log identifies the landmark-selection branch used in the recorded run.
\begin{figure*}[t]
    \centering
    \begin{minipage}[t]{0.49\textwidth}
    \small\textbf{Actions and their physical conditions}\par\smallskip
    \footnotesize
    \textit{Action list (completion conditions summarized)}\par\smallskip
    \begingroup
    \setlength{\tabcolsep}{3pt}
    \renewcommand{\arraystretch}{1.12}
    \begin{tabularx}{\linewidth}{@{}l>{\raggedright\arraybackslash}X@{}}
        \toprule
        Action & Physical completion condition \\
        \midrule
        \texttt{acquire\_bowl} & Bowl held between the fingers and lifted above its support. \\
        \texttt{transport\_bowl} & Held bowl above the plate at transit height. \\
        \texttt{place\_bowl} & Bowl on the plate, gripper open, and arm lifted clear. \\
        \bottomrule
    \end{tabularx}
    \endgroup

    \par\medskip
    \textbf{Coordinator (code excerpt)}
    \scriptsize
\begin{verbatim}
if action_acquire_bowl(context):
    if action_transport_bowl(context):
        action_place_bowl(context)
result = context["report"]
\end{verbatim}
    \footnotesize
    Each action returns whether its completion condition holds.
    The coordinator proceeds to the next action only when the preceding one succeeds.

    \par\medskip
    \textbf{What Alice can inspect}\par\smallskip
    The action names locate the physical stage.
    Their completion conditions state what must have happened before the policy proceeds.
    Within an action, the explicit branches expose the conditions and responses that Alice can test.

    \par\medskip
    \textbf{Assumption exposed by the branch at right}\par\smallskip
    If the landmark is not found, the policy chooses the highest bowl.
    This makes the fallback's assumption explicit: the target bowl is the highest one.
    \end{minipage}\hfill
    \begin{minipage}[t]{0.49\textwidth}
    \small\textbf{A decision inside \texttt{acquire\_bowl}}\par\smallskip
    \footnotesize
    \textit{Code excerpt: selecting the bowl on a landmark}
    \scriptsize
\begin{verbatim}
elif mode == "on":
    lm = landmark_point(ctx, landmark_text, bowls)
    if lm is not None:
        best = select_bowl_on(bowls, lm)
        used = "on:" + landmark_text
    else:
        # landmark not found: the target bowl starts
        # raised, so take the highest one
        for b in bowls:
            if (best is None
                    or b["z_low"] > best["z_low"]):
                best = b
        used = "on:highest"
\end{verbatim}
    \footnotesize
    \textbf{Condition:} was the landmark found?\par
    \textbf{Yes:} select the bowl on that landmark.\par
    \textbf{No:} select the highest bowl.

    \par\medskip
    \textbf{Recorded run: which branch was used?}\par
    \textit{Stdout excerpts (line-wrapped)}
    \scriptsize
\begin{verbatim}
acquire_bowl: held, gripper=0.058 side=1
  mode=on:ramekin
transport_bowl: above plate, grip=0.057 z=0.240
\end{verbatim}
    \footnotesize
    The label \texttt{on:ramekin} identifies the landmark-selection branch.
    The next log entry records transport above the plate.
    The native task succeeds after 425 simulator steps.
    \end{minipage}
    \caption{\textbf{The structure that Alice can inspect in \method{}-Bob.} A LIBERO-Spatial policy after the cooperative game. Left: a summary of its declared actions and physical completion conditions, with the actual coordinator. Right: an explicit condition and its two responses inside acquisition, followed by logs from a run on task 5. The action names identify the stage, the branches expose its decisions and assumptions, and the logs identify the branch used in the run. Code excerpts omit unrelated lines; the surrounding notes explain the structure for the reader.}
    \label{fig:rallybob-example}
\end{figure*}

\minisection{Alice's reasoning and selected challenge}
Figure~\ref{fig:rallybob-adversarial-example} shows Alice's reasoning about a transport weakness in LIBERO-Spatial task 9.
The incumbent policy re-localizes the plate, then retries transport at three heights with one tool orientation.
The transport code therefore exposes a specific limitation: its fallback changes height but keeps orientation fixed.
Alice identifies a reach-related weakness and selects a plate relocation in $+x$ as the challenge.

The recorded results support this selection: the $+x$ relocation fails in all three runs, each reporting three inverse-kinematics failures.
The named action and explicit retry code connect the physical reachability problem to the part of the policy that needs attention.
Bob's accepted repair adds a search over tool orientations and heights.
Appendix~\ref{app:qual:reach} gives the repair and execution details.

\begin{figure*}[t]
    \centering
    \begin{minipage}[t]{0.49\textwidth}
    \small\textbf{Policy available to Alice (P9)}\par\smallskip
    \footnotesize
    \textbf{Physical action: transport the held bowl to the plate.}
    The policy re-localizes the plate before the excerpt below, then computes the carrying target from that estimate.

    \par\smallskip
    \textit{Incumbent transport code (excerpt)}
    \scriptsize
\begin{verbatim}
tx = plate["xc"] - ctx["held_dx"]
ty = plate["yc"] - ctx["held_dy"]
ctx["place_xy"] = (tx, ty)
...
arrived = False
for dz in (0.0, -0.04, -0.08):
    z = ctx["lift_z"] + dz
    if mv(ctx, tx, ty, z, 0.02, 260):
        ctx["lift_z"] = z
        arrived = True
        break
    log(ctx, sid, "transport at z=%.3f failed" % z)
    if ctx["steps"] > 800:
        break
if not arrived:
    return False
\end{verbatim}
    \footnotesize
    \textbf{Condition:} can the robot reach the carrying pose?\par
    \textbf{On success:} record arrival and stop retrying.\par
    \textbf{On failure:} try a lower height with the same tool orientation; return failure if no attempt succeeds.

    \end{minipage}\hfill
    \begin{minipage}[t]{0.49\textwidth}
    \small\textbf{Alice's reasoning and selected challenge}\par\smallskip
    \footnotesize
    \textbf{Alice's diagnosis}\par\smallskip
    Alice identifies a reach-related transport weakness even though the policy re-observes the plate.
    \par\smallskip
    \textit{Recorded excerpt:}
    \lq\lq \texttt{move\_to\_pose} gives repeated \texttt{ik\_failed} and the placement fails, with no fallback to a reachable pose/height.\rq\rq

    \par\medskip
    \textbf{Challenge selected:} move the destination plate 10\,cm in $+x$ while the robot approaches the bowl, keeping the bowl-on-plate goal unchanged.

    \par\medskip
    \textbf{Recorded results for this challenge (summary)}\par\smallskip
    \begingroup
    \setlength{\tabcolsep}{3pt}
    \renewcommand{\arraystretch}{1.12}
    \begin{tabularx}{\linewidth}{@{}l c >{\raggedright\arraybackslash}X@{}}
        \toprule
        Challenge & Success & Feedback \\
        \midrule
        Plate moved in $+x$ & 0/3 & Three IK failures per run \\
        \bottomrule
    \end{tabularx}
    \endgroup
    \par\smallskip
    The failed calls are reported as \texttt{move\_to\_pose: ik\_failed}.

    \par\medskip
    \textbf{Link from the physical failure to the code}\par\smallskip
    The plate estimate is refreshed, so the policy does not simply reuse its old position.
    The transport action exposes the available fallback: change the height while keeping orientation fixed.
    Alice can connect reach failures to this action and its limited responses.
    \end{minipage}

    \par\medskip
    \begin{minipage}{\textwidth}
    \footnotesize
    \textbf{Subsequent repair.}
    Bob's accepted P11 adds a search over tool orientations and heights.
    \end{minipage}
    \caption{\textbf{Alice's reasoning links a transport weakness to the selected challenge.} In LIBERO-Spatial task 9, Bob must move the bowl from the wooden cabinet to the plate. Left: the incumbent transport code and its condition and fallback. Right: Alice's reachability diagnosis, a recorded excerpt, the selected $+x$ relocation, and the compact execution feedback supplied to Alice. All three runs fail, each reporting three inverse-kinematics failures. The named action and explicit retry loop link this physical failure to the transport code. The repair is the intermediate P9-to-P11 change in cycle 7, accepted after P10 was rejected. Appendix~\ref{app:qual:reach} gives the repair details.}
    \label{fig:rallybob-adversarial-example}
\end{figure*}

\subsection{Role Guidelines and Candidate Checks}
\label{app:refinement_cycle}
The guideline of adversarial Alice lists the kinds of variation to look for; it is the guideline summarized in Appendix~\ref{sec:appendix:frontier-prompt}, which Fable 5.2 also receives.

\minisection{Guideline of cooperative Alice}
Cooperative Alice is told that the policy solves one task and should grow into one policy for the tasks of its scene, which is then run unchanged on tasks that neither player sees.
In each cycle, Alice maps what the policy can do: starting near the task that the policy solves, Alice writes tasks that change one factor at a time and runs the current policy on them.
These tasks are sorted into solved ones, tasks at the edge, where some instances succeed or the failed runs get most of the way, and far tasks, which fail at the first step.
Alice then proposes the smallest extension at the edge that one refinement can achieve, preferring the one whose refinement should carry over to the most other tasks.
After a failed refinement, Alice does not repeat the task but proposes a simpler one that isolates part of it, and returns to the original goal later.
The guideline of each suite describes the scene and names the factors that can change and what a simpler task is.

\minisection{Adversarial payoff objective}
Alice's score of a variation (Eq.~\ref{eq:alice_score}) compares two success rates of the policy on the same instances of the variation.
The first, $s_0$, is measured when the variation is accepted, on the instances that Alice's test did not run; the second, $s_{\mathrm{end}}$, is measured two cycles later.
Alice's prompt lists every earlier variation with its type of change and its score.
A type of change is retired, so that Alice can no longer propose it, once two of its variations have been scored and both their score and Bob's best refinement attempt on them are below 0.1.
A run stops before its eighth cycle if, for two cycles in a row, the policy does not improve and no new variation scores above 0.1 (Table~\ref{tab:settings-adversarial}).

\minisection{Cooperative payoff objective}
The shared score (Eq.~\ref{eq:shared_score}) is measured again after every kept refinement: each accepted task is run on its five instances with the new policy, which gives its new success rate $s_k$.
An instance that the policy appears to lose is run twice more before it counts as lost.
Alice's prompt lists every accepted task with $s_{0,k}$ and its current $s_k$, so Alice sees which earlier gains the policy keeps (Table~\ref{tab:settings-cooperative}).

\minisection{Candidate acceptance checks}
Each cycle has two checks: one on Alice's proposal before Bob refines on it, and one on Bob's candidate policy after the refinement.
For the first, Alice ranks at most three candidate proposals and tests each on three instances with the current policy; the highest-ranked candidate that passes is accepted, and Bob receives it with five instances.
An adversarial variation passes if the policy succeeds on at most 60\% of the test instances; a dynamic variation must also succeed on every instance when it is run without its change, so that the failure comes from the change.
A cooperative task passes if its objects, success condition, or sub-tasks, or in Spatial its layout, differ from those of each of the other nine tasks of its suite (Appendix~\ref{sec:appendix:settings}), at least 10 of its 12 start states build into valid scenes, and the policy fails on at least one of its five instances.
For the second check, Bob's candidate policy is kept only if it succeeds on at least four of the five instances of the proposal and on more of them than the current policy (condition (i) of Sec.~\ref{subsec:method:overview}), and if it loses none of the scenes of the original task and of earlier proposals on which it is checked (condition (ii)).
Otherwise the current policy stays, and Bob can make a second attempt in the same cycle.

\subsection{Experiment Settings and Budgets}
\label{sec:appendix:settings}
\label{app:refinement_budgets}
This subsection gives every budget and threshold of the experiments (Sec.~\ref{subsec:eval:setup}).
Table~\ref{tab:settings-shared} lists what both settings share, and Tables~\ref{tab:settings-adversarial} and~\ref{tab:settings-cooperative} list what is specific to each cycle.
Table~\ref{tab:settings-suites} gives the cooperative setting per suite, Table~\ref{tab:settings-curricula} the curricula that it is compared with, and Table~\ref{tab:settings-evaluation} the evaluation sets.

\minisection{Unseen at test time: adversarial setting in simulation}
Using the scene functions of Table~\ref{tab:scene_changes} and a generic guideline (Appendix~\ref{sec:appendix:frontier-prompt}) that names general categories of perturbations, Alice constructs variations based on the weaknesses of Bob's current policy.
Alice builds these variations only on the development scenes of a task, and has no access to the test cases of LIBERO-Pro, LIBERO-Plus, or LIBERO-MAX (Table~\ref{tab:settings-evaluation}), their perturbation configurations, or benchmark-specific assets.
Only the general categories of perturbations, such as a change of the camera view or of the lighting, are common to Alice and the benchmarks, because these are the ways in which a physical scene varies and therefore what a robust policy has to handle.
Within a category, however, the change that Alice builds can differ completely from the one that a benchmark applies.
Beyond this, a task is sometimes tested on categories of perturbations that are completely unseen during its refinement.
Alice also causes substantial improvement on these unseen categories of perturbations.
For example, with ASPIRE as Bob, on the tasks where no cycle uses a scene with a camera change, the success on the camera perturbations of the benchmarks rises by 19.6 points; in the same way, it rises by 27.5 points for image noise and by 20.7 points for occlusion or added objects.

\minisection{Unseen at test time: cooperative setting in simulation}
Related tasks share most of their sub-goals, such as finding, grasping, and placing an object, but they can still differ in many ways: which object is manipulated, where the objects start, what the success condition is, and which sub-tasks are chained.
These differences combine into far too many tasks to refine on one by one, and in many cases the tasks that a robot will meet are not known in advance.
To be reused on a new task without further refinement, a frozen policy must therefore generalize to unseen tasks, that is, related tasks that it was never refined on.
Alice cannot propose all of these tasks, so the aim of the cooperative game is a policy that generalizes from the few related tasks that Alice proposes to the many tasks that remain unseen.
For this reason, we evaluate only on unseen tasks, to test whether the policy would also work when a new task arrives that was unknown during refinement.
In each LIBERO suite, the ten tasks are related, as they share most of their sub-goals (Appendix~\ref{app:libero_suites}).
Within a suite, Bob starts from the policy refined on task 0 and extends it to solve each proposed task of Alice (Sec.~\ref{subsec:method:alice}).
We evaluate the frozen policy on the nine unseen tasks of the suite, called the test tasks (Table~\ref{tab:settings-evaluation}), and report success rates on these tasks only, since Bob refines on task 0.
Using a separate checker (Appendix~\ref{app:refinement_cycle}), we ensure that each proposed task differs from every test task in at least one of the following: the objects involved, the success condition, and the sub-tasks.
In Spatial, where these three are the same for all ten tasks, each proposed task differs from every test task in the layout.
Tables~\ref{tab:coop-tasks-object} to~\ref{tab:coop-tasks-long} show examples of the proposed tasks next to the test tasks, and Figure~\ref{fig:coop-spatial-layout} shows where the target bowl starts in Spatial.
Each suite isolates one of the ways in which related tasks differ: its ten tasks differ mainly along one axis, and the suite tests whether the policy generalizes along this axis.
Even along a single axis, far more tasks are possible than the ten of a suite.
For stronger generalization, we also give Alice assets from outside the suite to vary this axis, such as other objects in Object.
Here we show the axis of each suite:
\begin{itemize}
    \item \textbf{Object.} The manipulated object.
    \item \textbf{Spatial.} Where the two bowls start in the scene.
    \item \textbf{Goal.} The success condition $G$, one of the predicates \texttt{On}, \texttt{In}, \texttt{Open}, or \texttt{TurnOn} applied to objects~\cite{liu2023libero,liberopro2025}, as in \texttt{On(bowl, plate)} or \texttt{TurnOn(stove)} (Appendix~\ref{sec:appendix:success}).
    \item \textbf{LIBERO-10.} The combination of sub-tasks that a task chains.
\end{itemize}

\minisection{Unseen at test time: real robot}
\label{sec:appendix:real}
Bob refines every policy only in simulation, so no real scene or trial is used in a cycle (Sec.~\ref{subsec:eval:real}).
The frozen policy therefore meets the real objects, camera views, and backgrounds for the first time at test time.

\begin{table}[t]
    \centering
    \caption{\textbf{Settings shared by both settings.}}
    \label{tab:settings-shared}
    \begingroup
    \footnotesize
    \setlength{\tabcolsep}{4pt}
    \renewcommand{\arraystretch}{1.08}
    \begin{tabularx}{\linewidth}{@{}>{\raggedright\arraybackslash}p{0.36\linewidth}>{\raggedright\arraybackslash}X@{}}
        \toprule
        \textbf{Setting} & \textbf{Value} \\
        \midrule
        Model of Alice and of Bob & Sonnet 5.5 (GPT-6.1 Sol in Tables~\ref{tab:adversarial-sol} and~\ref{tab:adversarial-sol-suite}; Fable 5.2 as Bob alone in Table~\ref{tab:frontier-model} and as Alice and Bob in Table~\ref{tab:cooperative-fable}) \\
        Harness of Bob & ASPIRE, RHO, GaP, or \method{}-Bob, each with its own generation and refinement steps \\
        Training tasks & LIBERO: Spatial, Object, Goal, LIBERO-10, ten tasks each \\
        Alice, per cycle & at most 12 tool turns, 48 test executions, 8 perception queries \\
        Proposals per cycle & at most 3, ranked by Alice; one is accepted \\
        Test of a proposal & 3 instances \\
        Accepted scene & 5 instances for Bob \\
        Bob, per cycle & at most 2 refinement attempts \\
        Refinement target & at least 4 of the 5 instances of the new scene, and more of them than the current policy \\
        Original task in every refinement & 5 scenes that the policy must keep \\
        Length of an execution & 1000 simulator steps (4000 in LIBERO-10) \\
        Reported number & success of the frozen final policy on start states that neither player saw \\
        \bottomrule
    \end{tabularx}
    \endgroup
\end{table}

\begin{table}[t]
    \centering
    \caption{\textbf{Settings of the adversarial cycle.} One policy per task.}
    \label{tab:settings-adversarial}
    \begingroup
    \footnotesize
    \setlength{\tabcolsep}{4pt}
    \renewcommand{\arraystretch}{1.08}
    \begin{tabularx}{\linewidth}{@{}>{\raggedright\arraybackslash}p{0.36\linewidth}>{\raggedright\arraybackslash}X@{}}
        \toprule
        \textbf{Setting} & \textbf{Value} \\
        \midrule
        Start & the policy that Bob refined on the original scenes of the task \\
        Cycles & at most 8 per task \\
        Early stop & 2 cycles in a row with no improvement and no new variation that scores above 0.1 \\
        Test of a dynamic variation & the same 3 instances are also run without the change \\
        Acceptance & the policy succeeds on at most 60\% of the test instances, and on every instance without the change \\
        Success at acceptance & measured on the instances that the test did not run \\
        Refinement attempt & 20 minutes; at most 16 attempts per run \\
        Score of a variation & measured 2 cycles after its acceptance \\
        Retired perturbation type & 2 scored variations whose score and best attempt are both below 0.1 \\
        \bottomrule
    \end{tabularx}
    \endgroup
\end{table}

\begin{table}[t]
    \centering
    \caption{\textbf{Settings of the cooperative cycle.} One policy per suite.}
    \label{tab:settings-cooperative}
    \begingroup
    \footnotesize
    \setlength{\tabcolsep}{4pt}
    \renewcommand{\arraystretch}{1.08}
    \begin{tabularx}{\linewidth}{@{}>{\raggedright\arraybackslash}p{0.36\linewidth}>{\raggedright\arraybackslash}X@{}}
        \toprule
        \textbf{Setting} & \textbf{Value} \\
        \midrule
        Start & the policy that Bob refined on task 0 of the suite, with the harness unchanged \\
        Cycles & 8 in Spatial and Object, 16 in Goal and LIBERO-10; no early stop \\
        New task & 12 start states drawn by the simulator's own placement sampler; at least 10 must build \\
        Acceptance & the task differs from every test task (objects, success condition, or sub-tasks; in Spatial, the layout), it is a valid scene, and the policy fails on at least 1 of its 5 instances \\
        Refinement attempt & 20 minutes in Spatial and Object, 40 minutes in Goal and LIBERO-10 \\
        Attempts per run & at most 16 with 8 cycles, 32 with 16 cycles \\
        Lost scene & run twice more before it counts as lost \\
        Score & every accepted task is measured again on its 5 instances after each kept refinement \\
        Evaluation & the other 9 tasks of the suite, start states 25 to 39 (135 executions per suite) \\
        RHO as Bob & one search of two mutations per cycle in place of the two attempts \\
        \bottomrule
    \end{tabularx}
    \endgroup
\end{table}

\begin{table*}[t]
    \centering
    \caption{\textbf{The cooperative setting per suite.} The first task is task 0 of the suite, the only LIBERO task that Bob refines on. The last column gives what Alice may vary when building a task (Table~\ref{tab:task_parts}).}
    \label{tab:settings-suites}
    \begingroup
    \footnotesize
    \setlength{\tabcolsep}{5pt}
    \renewcommand{\arraystretch}{1.08}
    \begin{tabularx}{\textwidth}{@{}l>{\raggedright\arraybackslash}p{0.27\textwidth}cc>{\raggedright\arraybackslash}X@{}}
        \toprule
        \textbf{Suite} & \textbf{First task} & \textbf{Cycles} & \textbf{Minutes per attempt} & \textbf{What a task of Alice varies} \\
        \midrule
        Spatial & Pick up the black bowl between the plate and the ramekin and place it on the plate & 8 & 20 & Where the target bowl and a second bowl start (table, landmark, drawer), the landmarks, and how the instruction refers to the target \\
        Object & Pick up the alphabet soup and place it in the basket & 8 & 20 & The target object, the five other objects of the scene, and all positions \\
        Goal & Open the middle drawer of the cabinet & 16 & 40 & The fixtures and objects of the scene, the start state of the fixtures, and one goal \\
        LIBERO-10 & Put both the alphabet soup and the tomato sauce in the basket & 16 & 40 & The room, the fixtures and objects, the start state of the fixtures, and one or two goals in order \\
        \bottomrule
    \end{tabularx}
    \endgroup
\end{table*}

\begin{table*}[t]
    \centering
    \caption{\textbf{Curricula compared with the cooperative Alice} (Table~\ref{tab:cooperative-curricula}). All run in the same cycle with the same Bob, the same acceptance and the same number of cycles; only the choice of the next task differs.}
    \label{tab:settings-curricula}
    \begingroup
    \footnotesize
    \setlength{\tabcolsep}{5pt}
    \renewcommand{\arraystretch}{1.08}
    \begin{tabularx}{\textwidth}{@{}l>{\raggedright\arraybackslash}X>{\raggedright\arraybackslash}p{0.2\textwidth}c@{}}
        \toprule
        \textbf{Curriculum} & \textbf{How the next task is chosen} & \textbf{What it reads} \\
        \midrule
        Random & A task is drawn at random from the scene functions of the suite. & Nothing \\
        TSCL~\cite{matiisen2019teacher} & The sampling teacher. Each category of tasks keeps its last 10 rewards, where a reward is the change of Bob's success on the task of a cycle. One reward is sampled per category, and a random task of the category with the largest absolute sample is next. & The success of past cycles \\
        OMNI-EPIC~\cite{faldor2025omniepic} & A language model sees up to 5 learned and 5 failed tasks of the archive and writes a next task that is new and learnable. A rejected task is returned with the error up to 5 times. A second call judges whether the task is interesting next to 10 tasks of the archive. A failed task is modified once. & The archive of tasks and their outcome, and the scene functions \\
        Trajectory-only Alice & The cooperative Alice with the same tools, without the policy: no source, no tool that reads it. & Recorded executions, camera images, outputs of perception, success \\
        Cooperative Alice & Sec.~\ref{subsec:method:alice}. & The policy, recorded executions, perception, success \\
        \bottomrule
    \end{tabularx}
    \endgroup
\end{table*}

\begin{table*}[t]
    \centering
    \caption{\textbf{Evaluation sets.} No scene of these sets is used in a cycle. In the adversarial setting each task is evaluated with its own policy; in the cooperative setting one policy is evaluated on the nine tasks that it was not refined on.}
    \label{tab:settings-evaluation}
    \begingroup
    \footnotesize
    \setlength{\tabcolsep}{5pt}
    \renewcommand{\arraystretch}{1.08}
    \begin{tabularx}{\textwidth}{@{}ll>{\raggedright\arraybackslash}X@{}}
        \toprule
        \textbf{Setting} & \textbf{Set} & \textbf{Scenes} \\
        \midrule
        Adversarial & LIBERO-Pro~\cite{liberopro2025} & Four original perturbations (object, position, semantic, task) and six extensions (camera view angle, initial pose, object shape, object texture, view occlusion, visual noise and glare) \\
        Adversarial & LIBERO-Plus~\cite{liberoplus} & Seven perturbations at five difficulty levels \\
        Adversarial & LIBERO-MAX~\cite{liberomax} & Eight dynamic perturbations; every scene is run without its change (Base) and with it (Dynamic) \\
        Cooperative & LIBERO held-out & Tasks 1 to 9 of the suite on start states 25 to 39 \\
        \bottomrule
    \end{tabularx}
    \endgroup
\end{table*}

\section{Additional Experimental Results}
\label{sec:appendix:additional_exp}

\subsection{\method{} on RoboTwin 2.0}
\label{sec:appendix:robotwin}
\begin{table}[t]
    \centering
    \caption{\textbf{\method{} on RoboTwin 2.0.} Success rates (\%) of ASPIRE policies on the randomized test scenes of RoboTwin 2.0, 100 scenes per task. Bob refines each policy on clean scenes only, alone or with two further cycles with the adversarial Alice. Average is over the 14 tasks.}
    \label{tab:robotwin}
    \begingroup
    \small
    \setlength{\tabcolsep}{4pt}
    \renewcommand{\arraystretch}{1.1}
    \begin{tabularx}{\columnwidth}{@{}X*{3}{>{\centering\arraybackslash}p{0.15\columnwidth}}@{}}
        \toprule
        \textbf{Task} & \textbf{Bob alone} & \textbf{+ Alice} & \textbf{$\Delta$} \\
        \midrule
        Adjust Bottle & 63 & \textbf{72} & $+9$ \\
        Beat Block Hammer & 57 & \textbf{86} & $+29$ \\
        Click Bell & 93 & \textbf{98} & $+5$ \\
        Handover Block & 81 & \textbf{89} & $+8$ \\
        Move Can Pot & 69 & \textbf{71} & $+2$ \\
        Move Pillbottle Pad & 24 & \textbf{25} & $+1$ \\
        Open Laptop & 98 & 98 & $0$ \\
        Place Container Plate & 40 & \textbf{87} & $+47$ \\
        Place Mouse Pad & 61 & \textbf{87} & $+26$ \\
        Place Object Scale & 39 & \textbf{61} & $+22$ \\
        Place Phone Stand & 69 & \textbf{76} & $+7$ \\
        Place Shoe & 100 & 100 & $0$ \\
        Press Stapler & 78 & \textbf{79} & $+1$ \\
        Stack Blocks Two & \textbf{100} & 98 & $-2$ \\
        \midrule
        Average & 69.4 & \textbf{80.5} & $+11.1$ \\
        \bottomrule
    \end{tabularx}
    \endgroup
\end{table}

We test whether the gains of the adversarial Alice carry over to another simulator, robot, and set of tasks, on 14 tasks of RoboTwin 2.0 (Appendix~\ref{app:robotwin}).

\minisection{Experimental protocol}
The protocol follows Sec.~\ref{subsec:eval:adversarial}, with ASPIRE as the harness and Sonnet 5.5 as Alice and Bob.
For each task, Bob first refines a policy on 15 clean development scenes, which gives the Bob-only baseline.
Starting from this policy, Bob then plays two cycles with Alice, instead of eight on LIBERO, under the same acceptance checks.
We port Alice's scene functions to RoboTwin 2.0; they draw only from the seen textures and instruction templates of RoboTwin 2.0, never from the unseen ones reserved for testing, and Alice never sees a test scene.
We evaluate the frozen policies on the randomized test scenes of Appendix~\ref{app:robotwin}.

\minisection{Results}
Alice raises the average success from 69.4\% to 80.5\%, an increase of 11.1 percentage points (Table~\ref{tab:robotwin}).
Over the same test scenes, the policies after Alice succeed on 194 scenes where the policies of Bob alone fail, and fail on 39 where they succeed.
The largest gains are on Place Container Plate ($+47$), Beat Block Hammer ($+29$), Place Mouse Pad ($+26$), and Place Object Scale ($+22$).
For example, on Place Container Plate, the policy of Bob alone targets a distractor instead of the container in every test scene that only the policy after Alice solves.
No task loses more than two points, and the tasks that Bob alone already solves in at least 93\% of the scenes stay near that level.
Under the step limits of RoboTwin 2.0, success rises from 68.7\% to 79.0\%.
The two cycles cost \$1.16 per task on average.
The adversarial Alice therefore also makes code policies more robust on another benchmark with its own domain randomization, after only two cycles.

\subsection{\method{} with Another Model}
\label{sec:appendix:model}

\begin{table}[t]
    \centering
    \caption{\textbf{Adversarial Alice with GPT-6.1 Sol.} Success rates (\%) with ASPIRE as the harness and GPT-6.1 Sol as both Alice and Bob, on the robustness test sets of 29 tasks of Spatial, Object, and Goal (all but Goal task 5), macro-averaged over tasks. The shaded row is Bob alone, and the row below is the same policy after the cycles with the adversarial Alice. Overall is computed as in Table~\ref{tab:main-results}.}
    \label{tab:adversarial-sol}
    \begingroup
    \small
    \setlength{\tabcolsep}{3pt}
    \renewcommand{\arraystretch}{1.18}
    \begin{tabularx}{\columnwidth}{@{}p{0.18\columnwidth}*{5}{>{\centering\arraybackslash}X}@{}}
        \toprule
        \textbf{Policy} & PRO & PLUS & \shortstack{MAX\\Base} & \shortstack{MAX\\Dyn.} & Overall \\
        \midrule
        \rowcolor{black!6}
        Bob only & 56.3 & 77.5 & 73.9 & 54.3 & 66.0 \\
        + Alice & 82.6 & 92.5 & 90.4 & 81.6 & 87.0 \\
        \bottomrule
    \end{tabularx}
    \endgroup
\end{table}

\begin{table}[t]
    \centering
    \caption{\textbf{Adversarial Alice with GPT-6.1 Sol, per suite.} The results of Table~\ref{tab:adversarial-sol} for each suite, macro-averaged over its tasks (10 in Spatial and Object, 9 in Goal).}
    \label{tab:adversarial-sol-suite}
    \begingroup
    \small
    \setlength{\tabcolsep}{3pt}
    \renewcommand{\arraystretch}{1.18}
    \begin{tabularx}{\columnwidth}{@{}p{0.14\columnwidth}p{0.14\columnwidth}*{5}{>{\centering\arraybackslash}X}@{}}
        \toprule
        \textbf{Suite} & \textbf{Policy} & PRO & PLUS & \shortstack{MAX\\Base} & \shortstack{MAX\\Dyn.} & Overall \\
        \midrule
        \rowcolor{black!6}
        \textbf{Spatial} & Bob only & 54.4 & 72.5 & 68.6 & 45.6 & 61.3 \\
        & + Alice & 78.4 & 93.5 & 86.6 & 78.1 & 84.7 \\
        \midrule
        \rowcolor{black!6}
        \textbf{Object} & Bob only & 50.8 & 80.8 & 78.1 & 57.5 & 66.5 \\
        & + Alice & 83.4 & 93.1 & 92.5 & 81.6 & 87.9 \\
        \midrule
        \rowcolor{black!6}
        \textbf{Goal} & Bob only & 64.7 & 79.6 & 75.0 & 60.2 & 70.6 \\
        & + Alice & 86.4 & 90.7 & 92.4 & 85.6 & 88.7 \\
        \bottomrule
    \end{tabularx}
    \endgroup
\end{table}

\minisection{Adversarial Alice}
Table~\ref{tab:adversarial-sol} repeats the adversarial game with GPT-6.1 Sol as both Alice and Bob, with ASPIRE as the harness and the test sets of Table~\ref{tab:frontier-model}; Table~\ref{tab:adversarial-sol-suite} gives the results per suite.
Bob alone with GPT-6.1 Sol produced no policy for Goal task 5, so we average over the other 29 tasks of Spatial, Object, and Goal.
The cycles with Alice raise the policy of Bob from 66.0\% to 87.0\% overall (+21.0).
As with Sonnet 5.5, the largest gains are under PRO (+26.3) and under the dynamic perturbations of MAX (+27.3).
Every suite gains between 18.1 and 23.4 points.
On seven of the 29 tasks the run ended before the last cycle, and we evaluate the last accepted policy.
Adversarial Alice therefore also makes the policy more robust with another model.

\minisection{Cooperative Alice}
\begin{table}[t]
    \centering
    \caption{\textbf{Cooperative Alice with another model.} Success rates (\%) on the other nine tasks of each suite, with ASPIRE as the harness and one model as both Alice and Bob. The setting is that of Table~\ref{tab:cooperative-main}: the shaded row is the policy refined on task 0, and the row below is the same policy after the cycles with the cooperative Alice. Average is over the three suites.}
    \label{tab:cooperative-fable}
    \begingroup
    \small
    \setlength{\tabcolsep}{3.5pt}
    \renewcommand{\arraystretch}{1.18}
    \begin{tabularx}{\columnwidth}{@{}p{0.22\columnwidth}p{0.18\columnwidth}*{4}{>{\centering\arraybackslash}X}@{}}
        \toprule
        \textbf{Model} & \textbf{Policy} & \textbf{Spatial} & \textbf{Object} & \textbf{Goal} & \textbf{Avg.} \\
        \midrule
        \rowcolor{black!6}
        \textbf{Sonnet 5.5} & Task 0 & 40.0 & 10.4 & 0.0 & 16.8 \\
        & + Alice & 91.9 & 68.9 & 60.7 & 73.8 \\
        \midrule
        \rowcolor{black!6}
        \textbf{Fable 5.2} & Task 0 & 38.5 & 11.1 & 0.0 & 16.5 \\
        & + Alice & 98.5 & 83.7 & 67.4 & 83.2 \\
        \bottomrule
    \end{tabularx}
    \endgroup
\end{table}

Table~\ref{tab:cooperative-fable} repeats the cooperative game with Fable 5.2 as both Alice and Bob, with ASPIRE as the harness.
The policy of task 0 starts at the same level as with Sonnet 5.5: it solves 38.5\% of the other 9 episodes in Spatial, 11.1\% in Object, and none in Goal.
After the cycles with Alice, the policy reaches 98.5\% in Spatial, 83.7\% in Object, and 67.4\% in Goal, against 91.9\%, 68.9\%, and 60.7\% with Sonnet 5.5.
Cooperative Alice therefore grows the policy of one task into a policy for its suite with either model.
Table~\ref{tab:coop-pertask-fable} gives these results per task.

\subsection{Additional Real-World Results}
\label{sec:appendix:real-replace}
\begin{table}[t]
    \centering
    \caption{\textbf{Simulation-to-real transfer on a long-horizon replace task.} Successes out of 20 trials on a Franka arm, for the policy of Bob alone and the policy after the cycles with Alice. Every policy is refined in simulation only and runs frozen on the robot. In a replace task, the robot removes the object that is in a container and places a second object there. The last row starts from another task, task 9 of LIBERO-10 (put the mug in the microwave and close it).}
    \label{tab:real-world-replace}
    \begingroup
    \small
    \setlength{\tabcolsep}{4pt}
    \renewcommand{\arraystretch}{1.12}
    \begin{tabularx}{\columnwidth}{@{}l>{\raggedright\arraybackslash}Xcc@{}}
        \toprule
        \textbf{Condition} & \textbf{Task} & \textbf{Bob only} & \textbf{+ Alice} \\
        \midrule
        Original task & Replace strawberry with can in a drawer & 8/20 & \textbf{18/20} \\
        \midrule
        \multicolumn{4}{@{}l}{\textit{Adversarial Alice}} \\
        Occluded & & 0/20 & \textbf{15/20} \\
        Moving target & & 0/20 & \textbf{16/20} \\
        \midrule
        \multicolumn{4}{@{}l}{\textit{Cooperative Alice}} \\
        Different target & Replace strawberry with can in a bowl & 3/20 & \textbf{17/20} \\
        Different object & Replace can with green tea box in a drawer & 0/20 & \textbf{14/20} \\
        \midrule
        \multicolumn{4}{@{}l}{\textit{Cooperative Alice}, starting from task 9 of LIBERO-10} \\
        Different object and container & Put strawberry in a closet and close the door & 0/20 & \textbf{12/20} \\
        \bottomrule
    \end{tabularx}
    \endgroup
\end{table}

\begin{figure*}[t]
    \centering
    \includegraphics[width=0.7\textwidth]{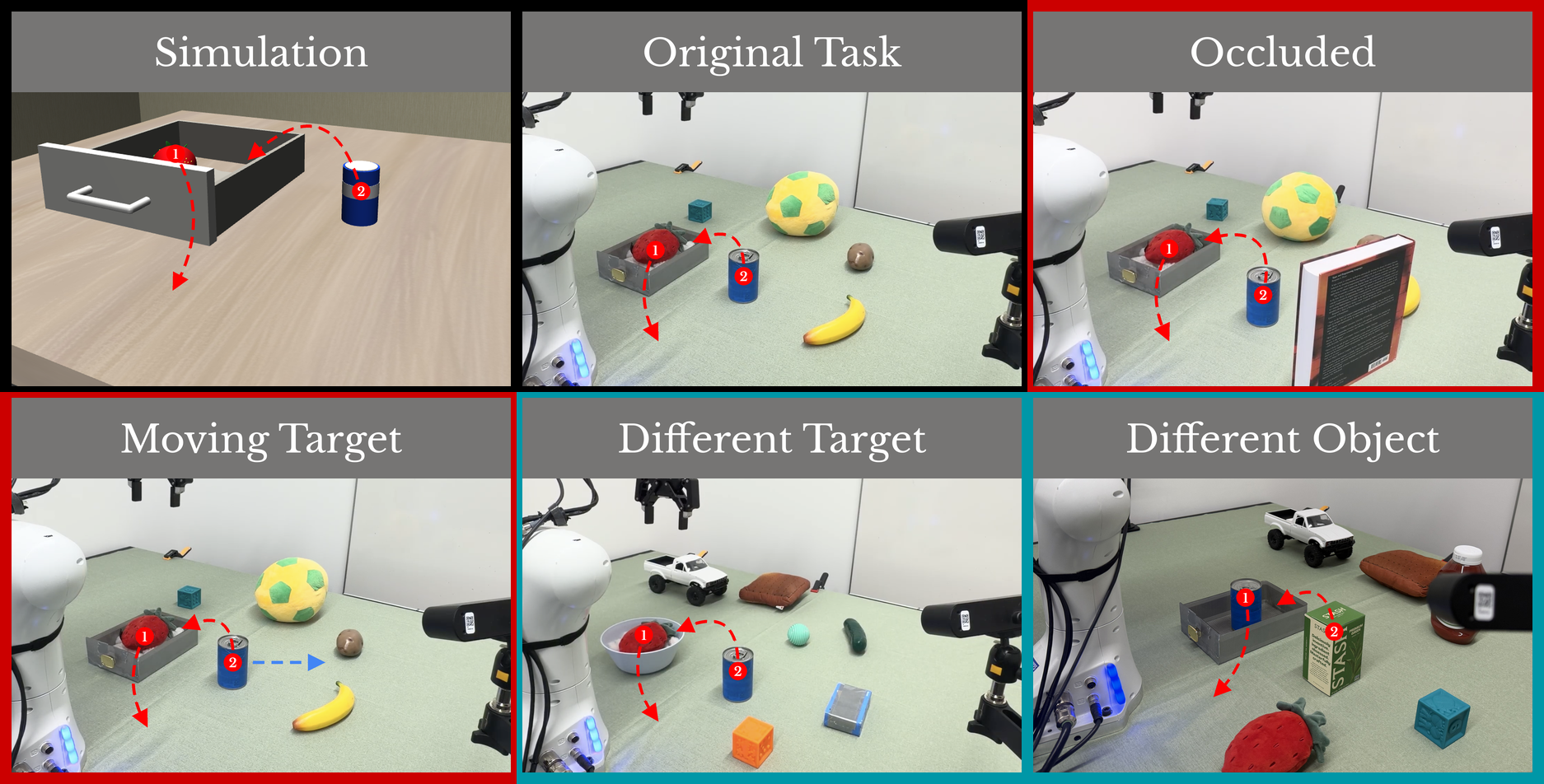}
    \caption{\textbf{From simulation to the real world on a long-horizon replace task.} Every policy starts from the replace task in simulation (first image) and runs on the Franka arm under the conditions of Table~\ref{tab:real-world-replace}. The numbered red dots mark the two objects in the order in which the robot moves them, and each dashed red arrow points to where the object goes. The blue arrow shows the can moving during execution. Black frames: the original task, in simulation and on the robot. Red frames: adversarial Alice. Teal frames: cooperative Alice.}
    \label{fig:real-world-replace}
\end{figure*}
We repeat the real-world experiment of Sec.~\ref{subsec:eval:real} on a long-horizon task.
The tasks of Sec.~\ref{subsec:eval:real} have one sub-task, placing an object in a target, while a replace task chains two: the robot removes the object that is in a container and then places a second object there.
Bob is ASPIRE with Sonnet 5.5 and refines every policy only in simulation, on a task that we built for this experiment and that is not part of LIBERO: replace the strawberry with the can in a drawer, where the drawer is taken out of its cabinet and lies on the table.
Because the task is long-horizon, adversarial and cooperative Alice each play 16 cycles, the number that cooperative Alice plays on LIBERO-10 (Table~\ref{tab:settings-cooperative}).
The frozen policy then runs on the same robot (Appendix~\ref{sec:appendix:hardware_setup}), with 20 trials per experiment (Figure~\ref{fig:real-world-replace}, Table~\ref{tab:real-world-replace}).

\minisection{Adversarial Alice}
As in Sec.~\ref{subsec:eval:real}, we introduce a static perturbation, an occlusion in the view of the scene camera, and a dynamic perturbation, in which the can moves when the arm approaches it.
The policy of Bob alone fails in every trial under both perturbations, while the policy refined with Alice remains robust to them (Table~\ref{tab:real-world-replace}).

\minisection{Cooperative Alice}
In simulation, cooperative Alice generalizes the policy by proposing replace tasks with different manipulated objects, targets, and layouts that the current policy cannot yet solve, and Bob extends the same policy to cover each of them.
On the robot, we test this policy on two new tasks: replace the strawberry with the can in a bowl (new target), and replace the can with a green tea box in the drawer (new object).
The policy of Bob alone rarely succeeds on these tasks, while the single policy refined with Alice generalizes across the manipulated objects and targets (Table~\ref{tab:real-world-replace}).

\begin{figure}[t]
    \centering
    \includegraphics[width=\columnwidth]{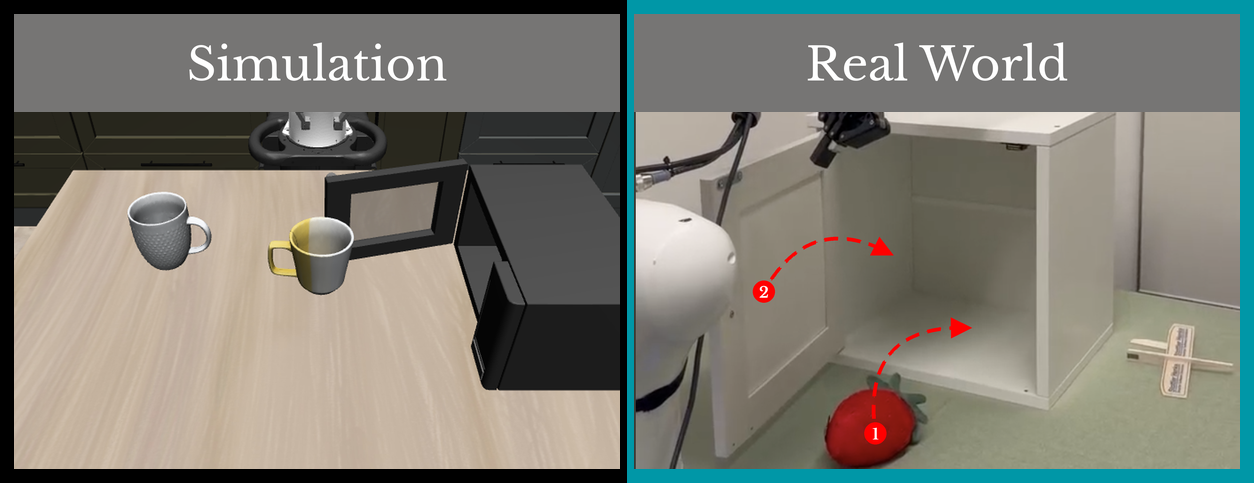}
    \caption{\textbf{From task 9 of LIBERO-10 to the real world.} Left: the task in simulation, put the yellow and white mug in the microwave and close it. Right: the task on the Franka arm, put the strawberry in a closet and close the door. On the right, the numbered red dots mark the object and the door in the order in which the robot moves them, and each dashed red arrow points to where they go.}
    \label{fig:real-world-microwave}
\end{figure}

\minisection{Cooperative Alice from a LIBERO task}
We also start from a task of LIBERO: task 9 of LIBERO-10, put the yellow and white mug in the microwave and close it.
Here Bob first refines the policy on task 9 itself, and, as in the prior experiments, cooperative Alice then proposes related tasks on which Bob extends this policy.
This differs from Table~\ref{tab:coop-pertask-long}, where Bob first refines on task 0 before Alice proposes tasks; here Bob starts from task 9.
On the robot, we test the frozen policy on a task with a different object and a different container: put the strawberry in a closet and close the door.
The policy of Bob alone succeeds in none of the 20 trials, and the policy refined with Alice in 12 of 20 (Figure~\ref{fig:real-world-microwave}, Table~\ref{tab:real-world-replace}).

\subsection{Standalone Evaluation of \method{}-Bob on LIBERO-Pro, RoboDojo and Robosuite}
\label{sec:appendix:bob_standalone}

A core advantage of code-as-policy is that policies are explicit programs, allowing us to directly define their structure.
\method{}-Bob organizes policies around physically grounded decision conditions and reusable responses.
This structure shares behavior across situations that present the same decision problem while preserving distinctions that require different actions.
For example, different object positions may require the same grasp strategy with newly observed coordinates, whereas successful, failed, and uncertain grasps require different responses.
Section~\ref{subsec:method:structured} and Appendix~\ref{app:structured_bob} describe this representation in detail.

This structure creates two opportunities for efficient policy refinement.
\emph{Localized improvement:} physical execution evidence can be connected to identifiable conditions and responses, supporting localized error diagnosis, focused repair, and targeted checks of the resulting behavior.
\emph{Reuse across trajectories:} experience from different trajectories can inform a shared response strategy, so repairing that strategy can improve behavior across multiple rollouts.
These properties support both players in \method{}: explicit decisions help Alice inspect Bob's policy and propose tasks that target its limitations, and give Bob a structured representation for improving its policy without Alice.
Below, we evaluate \method{}-Bob without Alice, measuring success on held-out scenes and the API cost of policy development.

\subsubsection{Direct optimization on LIBERO-Pro}
\label{app:bob:direct_optimization}

\minisection{Setting}
Here, Bob optimizes a separate policy directly on the development scenes of each LIBERO-Pro task variant, then freezes that policy for evaluation on separate test scenes of the same variant.
The variant itself is therefore known during optimization.
In the adversarial setting of Table~\ref{tab:main-results}, Bob instead optimizes on original LIBERO scenes, followed by Alice-proposed variations, and is evaluated on benchmark perturbations.
\method{}-Bob uses Claude Opus 5.5.
Following ASPIRE~\cite{lu2026aspire}, we optimize on 15 development seeds (51--65) and evaluate the frozen policy on 50 test seeds (1--50).

\minisection{Benchmark coverage}
Table~\ref{tab:libero-pro} covers 80 task variants: ten tasks from each of Object, Goal, Spatial, and Long, with position and task perturbations.
The table retains the reported results of learned policies, hybrid systems, and code-policy harnesses; methods without Long results are averaged over their 60 available task variants.

\begin{table*}[t]
    \centering
    \caption{LIBERO-Pro~\cite{liberopro2025} success (\%) under position (Pos.) and task perturbations, with API cost per task. Overall success averages the evaluated sub-suites. Dashes denote unavailable or unreported values.}
    \label{tab:libero-pro}
    \begingroup
    \footnotesize
    \setlength{\tabcolsep}{2.5pt}
    \renewcommand{\arraystretch}{1.12}
    \begin{tabularx}{\textwidth}{ll*{8}{>{\raggedleft\arraybackslash}X}rr}
        \toprule
        Method & Agent model & \multicolumn{2}{c}{Object} & \multicolumn{2}{c}{Goal} & \multicolumn{2}{c}{Spatial} & \multicolumn{2}{c}{Long} & Overall & API USD/task \\
        \cmidrule(lr){3-4}\cmidrule(lr){5-6}\cmidrule(lr){7-8}\cmidrule(lr){9-10}
        & & Pos. & Task & Pos. & Task & Pos. & Task & Pos. & Task & & \\
        \midrule
        OpenVLA~\cite{kim2024openvla} & --- & 0.0 & 0.0 & 0.0 & 0.0 & 0.0 & 0.0 & 0.0 & 0.0 & 0.0 & --- \\
        $\pi_0$~\cite{black2024pi0} & --- & 0.0 & 0.0 & 0.0 & 0.0 & 0.0 & 0.0 & 0.0 & 0.0 & 0.0 & --- \\
        $\pi_{0.5}$~\cite{black2025pi05} & --- & 17.0 & 1.0 & 38.0 & 0.0 & 20.0 & 1.0 & 8.0 & 1.0 & 10.8 & --- \\
        MolmoAct~\cite{lee2025molmoact} & --- & 6.0 & 0.0 & 0.0 & 0.0 & 0.0 & 0.0 & 0.0 & 6.0 & 1.5 & --- \\
        NORA~\cite{hung2025nora} & --- & 0.0 & 0.0 & 0.0 & 0.0 & 0.0 & 0.0 & 0.0 & 0.0 & 0.0 & --- \\
        x-VLA~\cite{zheng2025xvla} & --- & 2.0 & 8.0 & 1.0 & 9.0 & 0.0 & 0.0 & 0.0 & 10.0 & 3.8 & --- \\
        AtomVLA~\cite{sun2026atomvla} & --- & 10.0 & 0.0 & 2.0 & 11.0 & 16.0 & 1.0 & 1.0 & 9.0 & 6.3 & --- \\
        \midrule
        Harness VLA~\cite{zhang2026harnessvla} & GPT-5.5 & 91.0 & 94.0 & 66.0 & 75.0 & 69.0 & 81.0 & 49.0 & 52.0 & 72.1 & --- \\
        Harness VLA & Claude Opus 4.7 & 90.0 & 88.0 & 87.0 & 87.0 & 80.0 & 94.0 & 62.0 & 71.0 & 82.4 & --- \\
        \midrule
        CaP-Agent0~\cite{fu2026capx} & Gemini 3.1 Pro Preview & 27.0 & 31.0 & 29.0 & 16.0 & 13.0 & 23.0 & \cellcolor{black!10}--- & \cellcolor{black!10}--- & 23.2\rlap{\textsuperscript{*}} & 42.25\rlap{\textsuperscript{\textdagger}} \\
        RATs~\cite{zhang2026rats} & Gemini 3.1 Pro Preview & 61.0 & 63.0 & 43.0 & 36.0 & 29.0 & 31.0 & \cellcolor{black!10}--- & \cellcolor{black!10}--- & 43.8\rlap{\textsuperscript{*}} & 70.48\rlap{\textsuperscript{\textdagger}} \\
        \midrule
        RHO~\cite{rho2026} & GPT-5.5 & 12.2 & 37.2 & 50.6 & 55.6 & 53.0 & 61.6 & \cellcolor{black!10}--- & \cellcolor{black!10}--- & 45.0\rlap{\textsuperscript{*}} & --- \\
        ASPIRE~\cite{lu2026aspire} & Claude Opus 4.6 & 98.0 & 95.0 & 81.0 & 45.0 & 51.0 & 60.0 & 43.2 & 52.0 & 65.7 & 56.85 \\
        ASPIRE & Claude Opus 5 & 98.6 & 100.0 & 97.0 & 86.8 & 97.8 & 98.0 & 87.0 & 90.4 & 94.5 & 50.00 \\
        ASPIRE & GPT-6 Astra & 99.0 & 99.4 & 97.0 & 96.8 & 98.6 & 97.6 & 84.6 & 95.0 & 96.0 & 20.21 \\
        ASPIRE & Claude Opus 5.5 & 98.2 & 100.0 & 98.0 & 99.4 & 99.4 & 99.2 & 97.0 & 98.8 & 98.8 & 14.96 \\
        \addlinespace[2pt]
        \rowcolor{cvprblue!10}
        \textbf{\method{}-Bob} & Claude Opus 5.5 & 100.0 & 100.0 & 97.6 & 98.6 & 98.8 & 96.8 & 97.3 & 97.2 & 98.3 & 3.12 \\
        \bottomrule
    \end{tabularx}
    \par\vspace{3pt}
    \begin{minipage}{\textwidth}
        \footnotesize
        \textsuperscript{*}Overall success covers 60 tasks (Object, Goal, Spatial); all other overall scores cover 80 tasks, including Long.
        \textsuperscript{\textdagger}CaP-Agent0 and RATs success uses 10 trials/task; costs project execution to 50 trials/task over 60 tasks, with RATs play charged once.
        \par\smallskip
        \textit{Sources:} OpenVLA, $\pi_0$, $\pi_{0.5}$, MolmoAct, NORA, and x-VLA results are from the \href{https://github.com/Zxy-MLlab/LIBERO-PRO\#-libero-pro-model-leaderboard}{official LIBERO-Pro leaderboard}.
        CaP-Agent0 results are the reproduction reported by RATs~\cite{zhang2026rats}; other external results are from the cited method papers.
    \end{minipage}
    \endgroup
\end{table*}

\minisection{Results}
\method{}-Bob achieves \textbf{98.3\% overall success}, compared with 98.8\% for ASPIRE with the same Claude Opus 5.5 model, but at a much lower reported API cost: \textbf{\$3.12 per task versus \$14.96}, a reduction of approximately 79\%.
Its reported cost is also substantially below CaP-Agent0 (\$42.25 per task) and RATs (\$70.48 per task), which use LLM calls during task execution~\cite{zhang2026rats}.
\method{}-Bob requires no runtime LLM calls.

\subsubsection{LIBERO-Pro performance--cost frontier across models}
\label{app:bob:single_task}

\minisection{Comparison design}
To examine learning efficiency with fewer differences in task coverage and prior learned knowledge, we follow ASPIRE's ordinary-debugging study on the ten Goal tasks with position perturbations~\cite{lu2026aspire}.
Each task is learned independently, without carrying learned knowledge across tasks.
Our runs use fixed strategy-format starter guidance and the same development/test seed split as Appendix~\ref{app:bob:direct_optimization}, giving 500 test episodes per model setting.
The comparisons with Haiku 5.5, Sonnet 5.5, and GPT-6 Luna use the same base model on both sides; ASPIRE's other published models provide additional reference points.
Fixing the task set, testing protocol, and independent task learning makes the comparison more focused on single-task policy generation and repair.

\begin{figure*}[t]
    \centering
    \includegraphics[width=\textwidth]{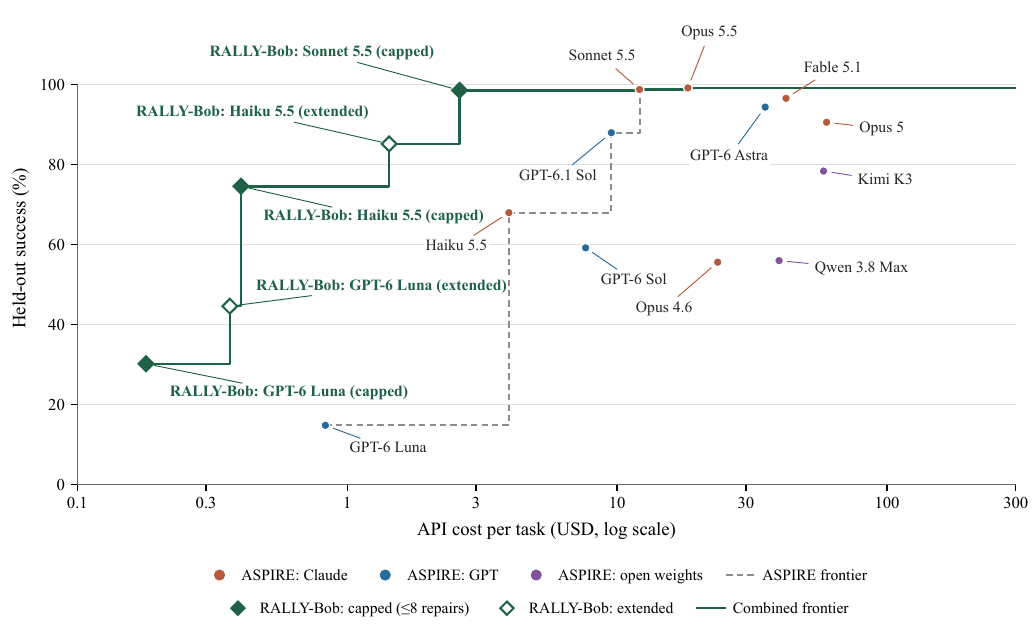}
    \caption{\textbf{Standalone single-task learning on LIBERO-Pro Goal/position.}
    Each point reports success over ten tasks and 50 test seeds per task, with each task learned independently.
    ASPIRE values are published ordinary-debugging results~\cite{lu2026aspire}.
    Filled diamonds allow at most eight repairs after initialization; open diamonds use extended budgets.
    The dashed curve is the ASPIRE frontier and the solid curve is the frontier over all plotted configurations: each gives the highest reported success available at or below a cost budget.}
    \label{fig:goal-position-pareto}
\end{figure*}

\minisection{Performance--cost frontier}
Figure~\ref{fig:goal-position-pareto} plots success against API cost per task.
With at most eight repairs, \method{}-Bob with Haiku achieves 74.6\% at \$0.40 per task, compared with ASPIRE's 68.0\% at \$3.97; with Luna, it achieves 30.2\% at \$0.18, compared with 14.8\% at \$0.83.
With Sonnet, \method{}-Bob achieves 98.6\% at \$2.61, compared with 98.8\% at \$12.11 for ASPIRE.
Across these three models, \method{}-Bob reduces reported API cost by factors of \textbf{4.6--9.8}.
Extended runs reach 85.2\% at \$1.43/task with Haiku and 44.6\% at \$0.37/task with Luna.

\subsubsection{Direct optimization on RoboDojo}
\label{app:bob:robodojo}

\minisection{Setting}
\method{}-Bob uses Claude Sonnet 5.5 to optimize each task policy on 15 development layouts, with up to eight repairs.
Development uses RoboDojo evaluation set~1, layouts 0--14; testing uses the separate evaluation set~0.
The selected program is frozen for evaluation.

\minisection{Benchmark coverage}
We compare the same ten RoboDojo tasks~\cite{chen2026robodojo}: putting bottles into a dustbin, classifying objects, arranging the largest number, building a tower, classifying objects by language, forming a kong, imitating a sorting sequence, folding clothes, organizing a table, and packing objects into a box.
Success means complete task success under the native evaluator, rather than its partial-credit \emph{Score}; we average the ten task success rates equally.
Our test uses 50 episodes per task: layouts 0--49 for ordinary tasks, or layouts 0--24 from each of the standard and randomized variants for number arrangement, folding, and packing.\footnote{For \texttt{imitate\_sorting\_sequence}, six native-invalid test layouts (4, 8, 17, 22, 36, 48) were replaced by layouts 50--55, preserving all other outcomes.}
PyRUA-Lean RGB~\cite{si2026pyrua} and Agent-as-Policy~\cite{agent2026policy} also use 50 episodes per task, with 25 standard and 25 randomized episodes for these three generalization tasks.
The GPT-as-Policy Direct and Hybrid results~\cite{su2026gptpolicy} use five episodes per task, with two standard and three randomized layouts for these three tasks.
We use published reports and released evaluation records for the baselines, including the public GPT-as-Policy report\footnote{\url{https://anonymous-report-421.github.io/public-website/?view=1}}; we do not rerun them.

\begin{table*}[t]
    \centering
    \caption{\textbf{Direct optimization on RoboDojo.} Success is the equally weighted mean of complete-task success over ten tasks. API costs are per task for 50 deployment episodes: \method{}-Bob includes one-time learning and zero runtime LLM usage; the other rows include deployment usage only, since separate learning costs are unavailable.}
    \label{tab:bob-robodojo}
    \begingroup
    \small
    \setlength{\tabcolsep}{6pt}
    \renewcommand{\arraystretch}{1.12}
    \begin{tabularx}{\textwidth}{Xlrr}
        \toprule
        Method & Agent model & Success (\%) $\uparrow$ & API USD/task $\downarrow$ \\
        \midrule
        GPT-as-Policy Direct~\cite{su2026gptpolicy} & GPT-6 Astra & 26.0 & $\approx$1,440.75\textsuperscript{*} \\
        GPT-as-Policy Hybrid~\cite{su2026gptpolicy} & GPT-6 Astra + $\pi_{0.5}$ & 48.0 & $\approx$831.83\textsuperscript{*} \\
        PyRUA-Lean (RGB)~\cite{si2026pyrua} & GPT-6 Astra & \textbf{58.8} & $\approx$120.00\textsuperscript{\textdagger} \\
        Agent-as-Policy~\cite{agent2026policy} & GPT-6 Astra & 55.8 & $\approx$162.99 \\
        \addlinespace[2pt]
        \rowcolor{cvprblue!10}
        \textbf{\method{}-Bob} & Claude Sonnet 5.5 & 46.2 & \textbf{15.21} \\
        \bottomrule
    \end{tabularx}
    \par\vspace{3pt}
    \begin{minipage}{\textwidth}
        \footnotesize
        \textsuperscript{*}Projected from five to 50 episodes per task, assuming unchanged mean token usage and caching.
        \textsuperscript{\textdagger}Uses the authors' benchmark-wide gateway estimate of \$2.40/episode.
        All values cover LLM API usage only; non-LLM compute is excluded.
    \end{minipage}
    \endgroup
\end{table*}

\minisection{Results}
\method{}-Bob achieves \textbf{46.2\% success} at a one-time learning API cost of \textbf{\$15.21 per task}, with no runtime LLM calls.
GPT-as-Policy, PyRUA-Lean, and Agent-as-Policy all invoke an LLM during task execution, incurring API costs on each deployment episode~\cite{su2026gptpolicy,si2026pyrua,agent2026policy}.
\method{}-Bob instead optimizes a program once and reuses the frozen program across episodes.
For 50 episodes per task, Table~\ref{tab:bob-robodojo} reports estimated baseline API costs of \$1,440.75 for Direct, \$831.83 for Hybrid, \$120.00 for PyRUA-Lean, and \$162.99 for Agent-as-Policy.
This corresponds to approximately \textbf{8--95$\times$ lower API cost}, approaching two orders of magnitude for \mbox{GPT-as-Policy Direct}.

For GPT-as-Policy and Agent-as-Policy, we estimate API cost from token records using
\begin{equation}
    C_{\mathrm{API}} = \frac{10T_{\mathrm{uncached}} + T_{\mathrm{cached}} + 50T_{\mathrm{output}}}{10^6},
    \label{eq:bob:robodojo-api-cost}
\end{equation}
where output includes reasoning tokens, and cached input is separated from total input to avoid double counting.
These reference-price estimates are not provider invoices and do not reconstruct long-context premiums or separate cache-write charges.
Direct and Hybrid are projected from 50 recorded runs to 500 episodes, assuming unchanged average token usage and caching; earlier retries are excluded.
Agent-as-Policy's estimate uses token records for all 500 selected episodes, without extrapolation.
PyRUA-Lean's approximately \$2.40 per episode is a gateway-price average across its full 42-task RGB evaluation; \$120 per task is therefore a cost proxy, without normalization to Eq.~\ref{eq:bob:robodojo-api-cost}.
All costs exclude simulator, perception, motion-planning, hardware, and VLA inference, including Hybrid's $\pi_{0.5}$ compute.

\subsubsection{Direct optimization on Robosuite}
\label{app:bob:robosuite}

\begin{table*}[t]
    \centering
    \caption{\textbf{Direct optimization on Robosuite.} Native task success averaged over seven tasks, with 100 evaluation episodes per task. API USD/task divides each reported learning/search total by seven. RHO is the paper's main result, trained on the first ten of its 100 evaluation trial IDs. Both methods deploy fixed code without LLM code-generation calls.}
    \label{tab:bob-robosuite}
    \begingroup
    \small
    \setlength{\tabcolsep}{6pt}
    \renewcommand{\arraystretch}{1.12}
    \begin{tabularx}{\textwidth}{Xlrr}
        \toprule
        Method & Agent model & Success (\%) $\uparrow$ & API USD/task $\downarrow$ \\
        \midrule
        RHO~\cite{rho2026} & GPT-5.5 & 70.00 & 403.00\textsuperscript{*} \\
        \addlinespace[2pt]
        \rowcolor{cvprblue!10}
        \textbf{\method{}-Bob} & Claude Opus 5.5 & \textbf{99.14} & \textbf{7.72} \\
        \bottomrule
    \end{tabularx}
    \par\vspace{3pt}
    \begin{minipage}{\textwidth}
        \footnotesize
        \textsuperscript{*}RHO's published \$2,821 estimate covers mutator tokens and excludes judge and visual-difference-model API charges.
    \end{minipage}
    \endgroup
\end{table*}

\minisection{Setting}
\method{}-Bob uses Claude Opus 5.5, with initial optimization on 25 development seeds (101--125) and evaluation on 100 test seeds (1--100).
The selected program is frozen for evaluation.

\minisection{Benchmark coverage}
We evaluate seven CaP-Bench Robosuite tasks~\cite{robosuite,fu2026capx}: cube lifting, cube stacking, cube restacking, square-nut assembly, spill wiping, two-arm lifting, and two-arm handover.
Success is native complete-task success, averaged equally across the seven tasks and 100 episodes per task.
We compare with RHO's published main-method result using GPT-5.5~\cite{rho2026}.

\minisection{Results}
Table~\ref{tab:bob-robosuite} reports \textbf{99.14\% success} for \method{}-Bob, compared with 70.00\% for RHO.
\method{}-Bob uses \textbf{\$7.72 per task}, compared with \$403.00 for RHO: approximately \textbf{52$\times$ lower learning API cost}.
Costs divide total learning/search API usage by seven tasks: \$54.02 for \method{}-Bob, including all repairs, and \$2,821 for RHO.
Both methods execute fixed programs without runtime LLM code generation.

\clearpage

\subsection{Adversarial Alice: Results per Perturbation}
\label{sec:appendix:adversarial}
This subsection breaks the results of Table~\ref{tab:harness-performance} down by perturbation, with one group of three tables per harness.
The first table of a group gives LIBERO-Pro and LIBERO-Plus per perturbation, the second gives LIBERO-MAX per perturbation without (Base) and with (Dynamic) the change, and the third gives LIBERO-Plus per perturbation and difficulty level.
ASPIRE is in Tables~\ref{tab:adv-pro-plus-aspire} to~\ref{tab:adv-plus-levels-aspire}, RHO in Tables~\ref{tab:adv-pro-plus-rho} to~\ref{tab:adv-plus-levels-rho}, GaP in Tables~\ref{tab:adv-pro-plus-gap} to~\ref{tab:adv-plus-levels-gap}, and \method{}-Bob in Tables~\ref{tab:adv-pro-plus-rally} to~\ref{tab:adv-plus-levels-rally}.
Dashes denote a perturbation that has no scenes in that suite.

\begin{table*}[t]
    \centering
    \caption{\textbf{ASPIRE: LIBERO-Pro and LIBERO-Plus per perturbation.} Success (\%) of the policy of ASPIRE alone (Bob) and after the adversarial game (+Alice), per suite. LIBERO-Pro has four original perturbations and six extensions added by the community to its official repository. Each value is the mean over the tasks of the suite. Avg.\ is the mean over the suites that have a value, as in Table~\ref{tab:main-results}.}
    \label{tab:adv-pro-plus-aspire}
    \begingroup
    \footnotesize
    \setlength{\tabcolsep}{4pt}
    \renewcommand{\arraystretch}{1.06}
    \begin{tabular}{@{}lrrrrrrrrrr@{}}
        \toprule
        \textbf{Perturbation} & \multicolumn{2}{c}{Spatial} & \multicolumn{2}{c}{Object} & \multicolumn{2}{c}{Goal} & \multicolumn{2}{c}{LIBERO-10} & \multicolumn{2}{c}{Avg.} \\
        \cmidrule(lr){2-3}\cmidrule(lr){4-5}\cmidrule(lr){6-7}\cmidrule(lr){8-9}\cmidrule(lr){10-11}
         & Bob & +Alice & Bob & +Alice & Bob & +Alice & Bob & +Alice & Bob & +Alice \\
        \midrule
        \multicolumn{11}{@{}l}{\textit{LIBERO-Pro}} \\
        \quad Object & 98.0 & 98.0 & 44.0 & 70.0 & 68.0 & 80.0 & 44.0 & 68.0 & 63.5 & 79.0 \\
        \quad Position & 78.0 & 92.0 & 66.0 & 90.0 & 54.0 & 72.0 & 34.0 & 56.0 & 58.0 & 77.5 \\
        \quad Semantic & 94.0 & 94.0 & 100.0 & 100.0 & 94.0 & 94.0 & 82.0 & 100.0 & 92.5 & 97.0 \\
        \quad Task & 0.0 & 2.0 & 0.0 & 0.0 & 0.0 & 0.0 & 10.0 & 10.0 & 2.5 & 3.0 \\
        \quad All original & 67.5 & 71.5 & 52.5 & 65.0 & 54.0 & 61.5 & 42.5 & 58.5 & 54.1 & 64.1 \\
        \addlinespace[2pt]
        \quad Camera view angle & 0.0 & 78.0 & 0.0 & 58.0 & 44.0 & 86.0 & 0.0 & 58.0 & 11.0 & 70.0 \\
        \quad Initial pose & 12.0 & 56.0 & 4.0 & 60.0 & 40.0 & 54.0 & 24.0 & 50.0 & 20.0 & 55.0 \\
        \quad Object shape & 92.0 & 94.0 & 98.0 & 100.0 & 94.0 & 98.0 & 82.0 & 94.0 & 91.5 & 96.5 \\
        \quad Object texture & 90.0 & 90.0 & 100.0 & 100.0 & 82.0 & 90.0 & 74.0 & 94.0 & 86.5 & 93.5 \\
        \quad View occlusion & 12.0 & 48.0 & 10.0 & 82.0 & 36.0 & 64.0 & 4.0 & 38.0 & 15.5 & 58.0 \\
        \quad Visual noise and glare & 58.0 & 96.0 & 92.0 & 98.0 & 90.0 & 98.0 & 76.0 & 92.0 & 79.0 & 96.0 \\
        \quad All extensions & 44.0 & 77.0 & 50.7 & 83.0 & 64.3 & 81.7 & 43.3 & 71.0 & 50.6 & 78.2 \\
        \addlinespace[2pt]
        All & 53.4 & 74.8 & 51.4 & 75.8 & 60.2 & 73.6 & 43.0 & 66.0 & 52.0 & 72.6 \\
        \midrule
        \multicolumn{11}{@{}l}{\textit{LIBERO-Plus}} \\
        \quad Background textures & 70.3 & 90.0 & 100.0 & 100.0 & 80.0 & 98.8 & 83.8 & 97.1 & 83.5 & 96.5 \\
        \quad Camera viewpoints & 72.9 & 92.0 & 93.1 & 96.1 & 75.0 & 97.9 & 78.8 & 92.8 & 79.9 & 94.7 \\
        \quad Language instructions & 90.0 & 100.0 & 100.0 & 100.0 & 80.0 & 100.0 & 90.0 & 100.0 & 90.0 & 100.0 \\
        \quad Light conditions & 82.2 & 100.0 & 100.0 & 100.0 & 81.3 & 100.0 & 87.5 & 100.0 & 87.7 & 100.0 \\
        \quad Objects layout & 91.1 & 94.4 & 93.3 & 90.2 & 58.3 & 72.3 & 73.7 & 83.1 & 79.1 & 85.0 \\
        \quad Robot initial states & 67.4 & 85.8 & 79.0 & 93.0 & 70.8 & 88.4 & 63.3 & 73.8 & 70.1 & 85.2 \\
        \quad Sensor noise & 10.6 & 89.7 & 27.0 & 83.0 & 45.9 & 76.9 & 36.7 & 62.4 & 30.0 & 78.0 \\
        \addlinespace[2pt]
        All & 68.9 & 93.4 & 82.1 & 94.0 & 68.4 & 89.4 & 72.3 & 85.4 & 72.9 & 90.5 \\
        \bottomrule
    \end{tabular}
    \endgroup
\end{table*}

\begin{table*}[t]
    \centering
    \caption{\textbf{ASPIRE: LIBERO-MAX per perturbation.} Success (\%) of the policy of ASPIRE alone (Bob) and after the adversarial game (+Alice). Every scene is run twice, without its change (Base) and with it (Dynamic). A perturbation row pools the executions of all tasks of the suite, since a task has few scenes of one perturbation; the rows All are means over the tasks. Avg.\ is the mean over the suites that have a value, as in Table~\ref{tab:main-results}.}
    \label{tab:adv-max-aspire}
    \begingroup
    \footnotesize
    \setlength{\tabcolsep}{4pt}
    \renewcommand{\arraystretch}{1.06}
    \begin{tabular}{@{}lrrrrrrrrrr@{}}
        \toprule
        \textbf{Perturbation} & \multicolumn{2}{c}{Spatial} & \multicolumn{2}{c}{Object} & \multicolumn{2}{c}{Goal} & \multicolumn{2}{c}{LIBERO-10} & \multicolumn{2}{c}{Avg.} \\
        \cmidrule(lr){2-3}\cmidrule(lr){4-5}\cmidrule(lr){6-7}\cmidrule(lr){8-9}\cmidrule(lr){10-11}
         & Bob & +Alice & Bob & +Alice & Bob & +Alice & Bob & +Alice & Bob & +Alice \\
        \midrule
        \multicolumn{11}{@{}l}{\textit{Base: the scene of each perturbation before the change happens}} \\
        \quad Camera shift & 72.2 & 88.9 & 55.6 & 77.8 & 63.6 & 93.9 & 72.7 & 86.4 & 66.0 & 86.7 \\
        \quad Distractor burst & 100.0 & 100.0 & 75.3 & 89.4 & --- & --- & 71.4 & 78.6 & 82.2 & 89.3 \\
        \quad Illumination switch & 66.7 & 85.7 & 89.5 & 94.7 & 60.0 & 80.0 & 60.0 & 93.3 & 69.0 & 88.4 \\
        \quad Obstacle insertion & 42.9 & 71.4 & 81.0 & 97.6 & 70.0 & 80.0 & 36.6 & 73.2 & 57.6 & 80.6 \\
        \quad Receptacle relocation & 67.4 & 82.6 & 66.7 & 86.7 & 60.0 & 90.0 & 64.3 & 71.4 & 64.6 & 82.7 \\
        \quad Sensor noise onset & 37.0 & 85.2 & 73.3 & 80.0 & 69.2 & 80.8 & 62.5 & 81.3 & 60.5 & 81.8 \\
        \quad Target relocation & 87.5 & 87.5 & 86.7 & 96.7 & 53.3 & 70.0 & 70.8 & 91.7 & 74.6 & 86.5 \\
        \quad Visual theme switch & 57.1 & 85.7 & 69.6 & 82.6 & 76.0 & 88.0 & 75.0 & 83.3 & 69.4 & 84.9 \\
        \addlinespace[2pt]
        All & 63.8 & 84.7 & 74.9 & 89.2 & 64.0 & 82.8 & 62.2 & 81.7 & 66.2 & 84.6 \\
        \midrule
        \multicolumn{11}{@{}l}{\textit{Dynamic: the change happens during the execution}} \\
        \quad Camera shift & 72.2 & 88.9 & 55.6 & 77.8 & 66.7 & 84.8 & 68.2 & 86.4 & 65.7 & 84.5 \\
        \quad Distractor burst & 0.0 & 0.0 & 57.6 & 75.3 & --- & --- & 42.9 & 50.0 & 33.5 & 41.8 \\
        \quad Illumination switch & 57.1 & 85.7 & 89.5 & 94.7 & 60.0 & 77.8 & 60.0 & 93.3 & 66.7 & 87.9 \\
        \quad Obstacle insertion & 42.9 & 85.7 & 81.0 & 97.6 & 70.0 & 70.0 & 17.1 & 36.6 & 52.7 & 72.5 \\
        \quad Receptacle relocation & 0.0 & 50.0 & 33.3 & 80.0 & 0.0 & 70.0 & 14.3 & 57.1 & 11.9 & 64.3 \\
        \quad Sensor noise onset & 40.7 & 74.1 & 73.3 & 66.7 & 69.2 & 73.1 & 56.3 & 68.8 & 59.9 & 70.6 \\
        \quad Target relocation & 25.0 & 75.0 & 3.3 & 70.0 & 6.7 & 43.3 & 0.0 & 37.5 & 8.8 & 56.5 \\
        \quad Visual theme switch & 60.7 & 89.3 & 69.6 & 78.3 & 76.0 & 88.0 & 79.2 & 79.2 & 71.4 & 83.7 \\
        \addlinespace[2pt]
        All & 40.9 & 75.0 & 55.2 & 79.3 & 54.3 & 74.3 & 42.5 & 61.9 & 48.2 & 72.6 \\
        \bottomrule
    \end{tabular}
    \endgroup
\end{table*}

\begin{table*}[t]
    \centering
    \caption{\textbf{ASPIRE: LIBERO-Plus per perturbation and level.} Success (\%) of the policy of ASPIRE alone (Bob) and after the adversarial game (+Alice) at each of the five difficulty levels of LIBERO-Plus. The table covers the 40 tasks of the four suites, and each value is the mean over the tasks that have a scene of that perturbation and level.}
    \label{tab:adv-plus-levels-aspire}
    \begingroup
    \footnotesize
    \setlength{\tabcolsep}{4pt}
    \renewcommand{\arraystretch}{1.06}
    \begin{tabular}{@{}lrrrrrrrrrrrr@{}}
        \toprule
        \textbf{Perturbation} & \multicolumn{2}{c}{Level 1} & \multicolumn{2}{c}{Level 2} & \multicolumn{2}{c}{Level 3} & \multicolumn{2}{c}{Level 4} & \multicolumn{2}{c}{Level 5} & \multicolumn{2}{c}{All levels} \\
        \cmidrule(lr){2-3}\cmidrule(lr){4-5}\cmidrule(lr){6-7}\cmidrule(lr){8-9}\cmidrule(lr){10-11}\cmidrule(lr){12-13}
         & Bob & +Alice & Bob & +Alice & Bob & +Alice & Bob & +Alice & Bob & +Alice & Bob & +Alice \\
        \midrule
        Background textures & 84.8 & 92.4 & 83.3 & 97.2 & 81.9 & 100.0 & 89.7 & 98.3 & 79.4 & 100.0 & 83.5 & 96.5 \\
        Camera viewpoints & 91.9 & 96.8 & 88.9 & 94.4 & 80.6 & 90.3 & 71.0 & 95.2 & 71.2 & 96.2 & 79.9 & 94.7 \\
        Language instructions & 89.2 & 100.0 & 92.3 & 100.0 & 89.7 & 100.0 & 90.0 & 100.0 & 88.9 & 100.0 & 90.0 & 100.0 \\
        Light conditions & 89.1 & 100.0 & 83.3 & 100.0 & 84.8 & 100.0 & 85.5 & 100.0 & 93.5 & 100.0 & 87.7 & 100.0 \\
        Objects layout & 79.7 & 84.4 & 83.3 & 88.9 & 84.2 & 97.4 & 66.7 & 73.3 & 68.3 & 73.3 & 79.1 & 85.0 \\
        Robot initial states & 82.9 & 94.7 & 78.8 & 87.5 & 81.3 & 83.8 & 61.5 & 82.1 & 46.3 & 78.8 & 70.1 & 85.2 \\
        Sensor noise & 48.6 & 80.6 & 36.3 & 83.8 & 25.0 & 81.3 & 23.8 & 76.3 & 12.1 & 60.3 & 30.0 & 78.0 \\
        \addlinespace[2pt]
        All perturbations & 80.0 & 92.2 & 77.0 & 92.3 & 74.8 & 92.3 & 67.4 & 88.8 & 62.9 & 84.5 & 72.9 & 90.5 \\
        \bottomrule
    \end{tabular}
    \endgroup
\end{table*}

\begin{table*}[t]
    \centering
    \caption{\textbf{RHO: LIBERO-Pro and LIBERO-Plus per perturbation.} Success (\%) of the policy of RHO alone (Bob) and after the adversarial game (+Alice), per suite. LIBERO-Pro has four original perturbations and six extensions added by the community to its official repository. Each value is the mean over the tasks of the suite.}
    \label{tab:adv-pro-plus-rho}
    \begingroup
    \footnotesize
    \setlength{\tabcolsep}{4pt}
    \renewcommand{\arraystretch}{1.06}
    \begin{tabular}{@{}lrrrrrrrrrr@{}}
        \toprule
        \textbf{Perturbation} & \multicolumn{2}{c}{Spatial} & \multicolumn{2}{c}{Object} & \multicolumn{2}{c}{Goal} & \multicolumn{2}{c}{LIBERO-10} & \multicolumn{2}{c}{Avg.} \\
        \cmidrule(lr){2-3}\cmidrule(lr){4-5}\cmidrule(lr){6-7}\cmidrule(lr){8-9}\cmidrule(lr){10-11}
         & Bob & +Alice & Bob & +Alice & Bob & +Alice & Bob & +Alice & Bob & +Alice \\
        \midrule
        \multicolumn{11}{@{}l}{\textit{LIBERO-Pro}} \\
        \quad Object & 96.0 & 98.0 & 62.0 & 68.0 & 58.0 & 78.0 & 20.0 & 26.0 & 59.0 & 67.5 \\
        \quad Position & 66.0 & 80.0 & 34.0 & 86.0 & 22.0 & 60.0 & 2.0 & 6.0 & 31.0 & 58.0 \\
        \quad Semantic & 96.0 & 98.0 & 84.0 & 98.0 & 76.0 & 98.0 & 36.0 & 40.0 & 73.0 & 83.5 \\
        \quad Task & 2.0 & 0.0 & 0.0 & 0.0 & 0.0 & 0.0 & 4.0 & 10.0 & 1.5 & 2.5 \\
        \quad All original & 65.0 & 69.0 & 45.0 & 63.0 & 39.0 & 59.0 & 15.5 & 20.5 & 41.1 & 52.9 \\
        \addlinespace[2pt]
        \quad Camera view angle & 0.0 & 10.0 & 0.0 & 30.0 & 20.0 & 86.0 & 0.0 & 18.0 & 5.0 & 36.0 \\
        \quad Initial pose & 34.0 & 42.0 & 16.0 & 30.0 & 28.0 & 46.0 & 12.0 & 8.0 & 22.5 & 31.5 \\
        \quad Object shape & 98.0 & 100.0 & 86.0 & 98.0 & 80.0 & 98.0 & 32.0 & 42.0 & 74.0 & 84.5 \\
        \quad Object texture & 98.0 & 98.0 & 86.0 & 98.0 & 78.0 & 92.0 & 22.0 & 44.0 & 71.0 & 83.0 \\
        \quad View occlusion & 14.0 & 32.0 & 0.0 & 36.0 & 14.0 & 60.0 & 10.0 & 26.0 & 9.5 & 38.5 \\
        \quad Visual noise and glare & 66.0 & 94.0 & 54.0 & 80.0 & 60.0 & 90.0 & 22.0 & 24.0 & 50.5 & 72.0 \\
        \quad All extensions & 51.7 & 62.7 & 40.3 & 62.0 & 46.7 & 78.7 & 16.3 & 27.0 & 38.8 & 57.6 \\
        \addlinespace[2pt]
        All & 57.0 & 65.2 & 42.2 & 62.4 & 43.6 & 70.8 & 16.0 & 24.4 & 39.7 & 55.7 \\
        \midrule
        \multicolumn{11}{@{}l}{\textit{LIBERO-Plus}} \\
        \quad Background textures & 87.5 & 97.5 & 78.9 & 90.8 & 75.0 & 100.0 & 33.6 & 42.7 & 68.7 & 82.8 \\
        \quad Camera viewpoints & 69.0 & 87.2 & 50.8 & 77.3 & 59.4 & 93.0 & 31.6 & 35.3 & 52.7 & 73.2 \\
        \quad Language instructions & 100.0 & 100.0 & 90.0 & 100.0 & 80.0 & 100.0 & 40.0 & 50.0 & 77.5 & 87.5 \\
        \quad Light conditions & 92.6 & 98.3 & 78.9 & 95.8 & 78.3 & 98.3 & 42.3 & 55.0 & 73.0 & 86.9 \\
        \quad Objects layout & 78.1 & 82.0 & 70.5 & 79.5 & 39.9 & 67.2 & 30.7 & 39.3 & 54.8 & 67.0 \\
        \quad Robot initial states & 86.8 & 93.0 & 78.0 & 79.0 & 71.9 & 88.6 & 40.0 & 39.3 & 69.2 & 75.0 \\
        \quad Sensor noise & 16.9 & 26.1 & 17.0 & 46.0 & 19.1 & 48.4 & 6.9 & 21.0 & 15.0 & 35.4 \\
        \addlinespace[2pt]
        All & 76.6 & 83.5 & 64.3 & 79.2 & 58.5 & 83.7 & 31.5 & 39.7 & 57.7 & 71.5 \\
        \bottomrule
    \end{tabular}
    \endgroup
\end{table*}

\begin{table*}[t]
    \centering
    \caption{\textbf{RHO: LIBERO-MAX per perturbation.} Success (\%) of the policy of RHO alone (Bob) and after the adversarial game (+Alice). Every scene is run twice, without its change (Base) and with it (Dynamic). A perturbation row pools the executions of all tasks of the suite, since a task has few scenes of one perturbation; the rows All are means over the tasks.}
    \label{tab:adv-max-rho}
    \begingroup
    \footnotesize
    \setlength{\tabcolsep}{4pt}
    \renewcommand{\arraystretch}{1.06}
    \begin{tabular}{@{}lrrrrrrrrrr@{}}
        \toprule
        \textbf{Perturbation} & \multicolumn{2}{c}{Spatial} & \multicolumn{2}{c}{Object} & \multicolumn{2}{c}{Goal} & \multicolumn{2}{c}{LIBERO-10} & \multicolumn{2}{c}{Avg.} \\
        \cmidrule(lr){2-3}\cmidrule(lr){4-5}\cmidrule(lr){6-7}\cmidrule(lr){8-9}\cmidrule(lr){10-11}
         & Bob & +Alice & Bob & +Alice & Bob & +Alice & Bob & +Alice & Bob & +Alice \\
        \midrule
        \multicolumn{11}{@{}l}{\textit{Base: the scene of each perturbation before the change happens}} \\
        \quad Camera shift & 88.9 & 88.9 & 33.3 & 66.7 & 60.6 & 84.8 & 18.2 & 22.7 & 50.3 & 65.8 \\
        \quad Distractor burst & 100.0 & 100.0 & 60.0 & 87.1 & --- & --- & 21.4 & 35.7 & 60.5 & 74.3 \\
        \quad Illumination switch & 52.4 & 71.4 & 52.6 & 78.9 & 40.0 & 77.8 & 33.3 & 53.3 & 44.6 & 70.4 \\
        \quad Obstacle insertion & 71.4 & 85.7 & 69.0 & 83.3 & 90.0 & 90.0 & 4.9 & 7.3 & 58.8 & 66.6 \\
        \quad Receptacle relocation & 76.1 & 78.3 & 46.7 & 73.3 & 60.0 & 70.0 & 42.9 & 57.1 & 56.4 & 69.7 \\
        \quad Sensor noise onset & 59.3 & 59.3 & 40.0 & 80.0 & 57.7 & 76.9 & 50.0 & 56.3 & 51.7 & 68.1 \\
        \quad Target relocation & 81.3 & 93.8 & 56.7 & 73.3 & 53.3 & 70.0 & 25.0 & 33.3 & 54.1 & 67.6 \\
        \quad Visual theme switch & 57.1 & 71.4 & 56.5 & 69.6 & 60.0 & 80.0 & 37.5 & 41.7 & 52.8 & 65.7 \\
        \addlinespace[2pt]
        All & 70.9 & 77.7 & 59.2 & 79.6 & 54.9 & 78.2 & 28.6 & 38.0 & 53.4 & 68.4 \\
        \midrule
        \multicolumn{11}{@{}l}{\textit{Dynamic: the change happens during the execution}} \\
        \quad Camera shift & 88.9 & 88.9 & 22.2 & 66.7 & 54.5 & 60.6 & 18.2 & 13.6 & 46.0 & 57.4 \\
        \quad Distractor burst & 100.0 & 100.0 & 52.9 & 64.7 & --- & --- & 21.4 & 21.4 & 58.1 & 62.0 \\
        \quad Illumination switch & 52.4 & 71.4 & 57.9 & 78.9 & 40.0 & 75.6 & 33.3 & 53.3 & 45.9 & 69.8 \\
        \quad Obstacle insertion & 57.1 & 71.4 & 69.0 & 83.3 & 80.0 & 80.0 & 2.4 & 2.4 & 52.2 & 59.3 \\
        \quad Receptacle relocation & 0.0 & 47.8 & 16.7 & 43.3 & 0.0 & 20.0 & 0.0 & 28.6 & 4.2 & 34.9 \\
        \quad Sensor noise onset & 59.3 & 59.3 & 40.0 & 60.0 & 61.5 & 61.5 & 37.5 & 46.9 & 49.6 & 56.9 \\
        \quad Target relocation & 0.0 & 37.5 & 6.7 & 26.7 & 0.0 & 20.0 & 0.0 & 8.3 & 1.7 & 23.1 \\
        \quad Visual theme switch & 57.1 & 71.4 & 60.9 & 73.9 & 60.0 & 76.0 & 33.3 & 41.7 & 52.8 & 65.8 \\
        \addlinespace[2pt]
        All & 44.2 & 64.5 & 46.1 & 59.7 & 42.1 & 60.7 & 17.7 & 26.4 & 37.5 & 52.8 \\
        \bottomrule
    \end{tabular}
    \endgroup
\end{table*}

\begin{table*}[t]
    \centering
    \caption{\textbf{RHO: LIBERO-Plus per perturbation and level.} Success (\%) of the policy of RHO alone (Bob) and after the adversarial game (+Alice) at each of the five difficulty levels of LIBERO-Plus.}
    \label{tab:adv-plus-levels-rho}
    \begingroup
    \footnotesize
    \setlength{\tabcolsep}{4pt}
    \renewcommand{\arraystretch}{1.06}
    \begin{tabular}{@{}lrrrrrrrrrrrr@{}}
        \toprule
        \textbf{Perturbation} & \multicolumn{2}{c}{Level 1} & \multicolumn{2}{c}{Level 2} & \multicolumn{2}{c}{Level 3} & \multicolumn{2}{c}{Level 4} & \multicolumn{2}{c}{Level 5} & \multicolumn{2}{c}{All levels} \\
        \cmidrule(lr){2-3}\cmidrule(lr){4-5}\cmidrule(lr){6-7}\cmidrule(lr){8-9}\cmidrule(lr){10-11}\cmidrule(lr){12-13}
         & Bob & +Alice & Bob & +Alice & Bob & +Alice & Bob & +Alice & Bob & +Alice & Bob & +Alice \\
        \midrule
        Background textures & 74.2 & 89.4 & 70.8 & 86.1 & 69.4 & 81.9 & 72.4 & 82.8 & 50.0 & 67.6 & 68.7 & 82.8 \\
        Camera viewpoints & 61.3 & 83.9 & 55.6 & 77.8 & 52.8 & 70.8 & 48.4 & 67.7 & 40.4 & 65.4 & 52.7 & 73.2 \\
        Language instructions & 78.4 & 89.2 & 76.9 & 87.2 & 76.9 & 87.2 & 77.5 & 87.5 & 75.0 & 86.1 & 77.5 & 87.5 \\
        Light conditions & 80.4 & 100.0 & 71.7 & 85.0 & 69.7 & 86.4 & 67.7 & 85.5 & 69.6 & 80.4 & 73.0 & 86.9 \\
        Objects layout & 51.6 & 62.5 & 56.9 & 69.4 & 57.9 & 76.3 & 55.0 & 63.3 & 46.7 & 65.0 & 54.8 & 67.0 \\
        Robot initial states & 78.9 & 82.9 & 70.0 & 76.3 & 68.8 & 77.5 & 69.2 & 75.6 & 63.8 & 67.5 & 69.2 & 75.0 \\
        Sensor noise & 27.8 & 45.8 & 15.0 & 33.8 & 10.0 & 28.8 & 11.3 & 40.0 & 6.9 & 25.9 & 15.0 & 35.4 \\
        \addlinespace[2pt]
        All perturbations & 61.4 & 74.9 & 58.1 & 71.2 & 57.8 & 71.7 & 56.0 & 71.0 & 54.2 & 68.3 & 57.7 & 71.5 \\
        \bottomrule
    \end{tabular}
    \endgroup
\end{table*}

\begin{table*}[t]
    \centering
    \caption{\textbf{GaP: LIBERO-Pro and LIBERO-Plus per perturbation.} Success (\%) of the policy of GaP alone (Bob) and after the adversarial game (+Alice), per suite. LIBERO-Pro has four original perturbations and six extensions added by the community to its official repository. Each value is the mean over the tasks of the suite.}
    \label{tab:adv-pro-plus-gap}
    \begingroup
    \footnotesize
    \setlength{\tabcolsep}{4pt}
    \renewcommand{\arraystretch}{1.06}
    \begin{tabular}{@{}lrrrrrrrrrr@{}}
        \toprule
        \textbf{Perturbation} & \multicolumn{2}{c}{Spatial} & \multicolumn{2}{c}{Object} & \multicolumn{2}{c}{Goal} & \multicolumn{2}{c}{LIBERO-10} & \multicolumn{2}{c}{Avg.} \\
        \cmidrule(lr){2-3}\cmidrule(lr){4-5}\cmidrule(lr){6-7}\cmidrule(lr){8-9}\cmidrule(lr){10-11}
         & Bob & +Alice & Bob & +Alice & Bob & +Alice & Bob & +Alice & Bob & +Alice \\
        \midrule
        \multicolumn{11}{@{}l}{\textit{LIBERO-Pro}} \\
        \quad Object & 80.0 & 88.0 & 34.0 & 82.0 & 46.0 & 52.0 & 18.0 & 42.0 & 44.5 & 66.0 \\
        \quad Position & 70.0 & 86.0 & 60.0 & 80.0 & 32.0 & 46.0 & 28.0 & 42.0 & 47.5 & 63.5 \\
        \quad Semantic & 84.0 & 92.0 & 92.0 & 98.0 & 60.0 & 68.0 & 54.0 & 64.0 & 72.5 & 80.5 \\
        \quad Task & 0.0 & 2.0 & 0.0 & 0.0 & 0.0 & 0.0 & 2.0 & 2.0 & 0.5 & 1.0 \\
        \quad All original & 58.5 & 67.0 & 46.5 & 65.0 & 34.5 & 41.5 & 25.5 & 37.5 & 41.3 & 52.8 \\
        \addlinespace[2pt]
        \quad Camera view angle & 6.0 & 32.0 & 0.0 & 56.0 & 36.0 & 52.0 & 0.0 & 20.0 & 10.5 & 40.0 \\
        \quad Initial pose & 28.0 & 30.0 & 24.0 & 24.0 & 22.0 & 42.0 & 16.5 & 18.5 & 22.6 & 28.6 \\
        \quad Object shape & 80.0 & 88.0 & 96.0 & 96.0 & 66.0 & 78.0 & 50.0 & 54.0 & 73.0 & 79.0 \\
        \quad Object texture & 80.0 & 74.0 & 94.0 & 90.0 & 66.0 & 78.0 & 50.0 & 52.0 & 72.5 & 73.5 \\
        \quad View occlusion & 4.0 & 8.0 & 8.0 & 48.0 & 18.0 & 42.0 & 4.0 & 20.0 & 8.5 & 29.5 \\
        \quad Visual noise and glare & 44.0 & 58.0 & 58.0 & 86.0 & 44.0 & 56.0 & 46.0 & 58.0 & 48.0 & 64.5 \\
        \quad All extensions & 40.3 & 48.3 & 46.7 & 66.7 & 42.0 & 58.0 & 28.2 & 37.2 & 39.3 & 52.6 \\
        \addlinespace[2pt]
        All & 47.6 & 55.8 & 46.6 & 66.0 & 39.0 & 51.4 & 27.1 & 37.3 & 40.1 & 52.6 \\
        \midrule
        \multicolumn{11}{@{}l}{\textit{LIBERO-Plus}} \\
        \quad Background textures & 77.1 & 82.9 & 89.2 & 97.1 & 56.4 & 71.0 & 44.0 & 62.8 & 66.7 & 78.4 \\
        \quad Camera viewpoints & 62.9 & 75.7 & 66.0 & 85.9 & 44.4 & 64.3 & 43.0 & 54.2 & 54.1 & 70.0 \\
        \quad Language instructions & 80.0 & 80.0 & 96.0 & 98.0 & 70.0 & 80.0 & 40.0 & 63.1 & 71.5 & 80.3 \\
        \quad Light conditions & 81.4 & 72.5 & 91.7 & 98.8 & 68.6 & 80.0 & 50.0 & 60.0 & 72.9 & 77.8 \\
        \quad Objects layout & 74.6 & 77.1 & 83.0 & 84.3 & 28.1 & 40.2 & 48.3 & 52.0 & 58.5 & 63.4 \\
        \quad Robot initial states & 73.7 & 74.4 & 93.0 & 97.0 & 54.6 & 67.6 & 42.0 & 67.0 & 65.8 & 76.5 \\
        \quad Sensor noise & 13.9 & 44.7 & 16.0 & 60.0 & 20.0 & 37.5 & 7.3 & 42.5 & 14.3 & 46.2 \\
        \addlinespace[2pt]
        All & 65.1 & 72.7 & 74.8 & 87.5 & 47.1 & 61.8 & 37.5 & 57.3 & 56.1 & 69.8 \\
        \bottomrule
    \end{tabular}
    \endgroup
\end{table*}

\begin{table*}[t]
    \centering
    \caption{\textbf{GaP: LIBERO-MAX per perturbation.} Success (\%) of the policy of GaP alone (Bob) and after the adversarial game (+Alice). Every scene is run twice, without its change (Base) and with it (Dynamic). A perturbation row pools the executions of all tasks of the suite, since a task has few scenes of one perturbation; the rows All are means over the tasks.}
    \label{tab:adv-max-gap}
    \begingroup
    \footnotesize
    \setlength{\tabcolsep}{4pt}
    \renewcommand{\arraystretch}{1.06}
    \begin{tabular}{@{}lrrrrrrrrrr@{}}
        \toprule
        \textbf{Perturbation} & \multicolumn{2}{c}{Spatial} & \multicolumn{2}{c}{Object} & \multicolumn{2}{c}{Goal} & \multicolumn{2}{c}{LIBERO-10} & \multicolumn{2}{c}{Avg.} \\
        \cmidrule(lr){2-3}\cmidrule(lr){4-5}\cmidrule(lr){6-7}\cmidrule(lr){8-9}\cmidrule(lr){10-11}
         & Bob & +Alice & Bob & +Alice & Bob & +Alice & Bob & +Alice & Bob & +Alice \\
        \midrule
        \multicolumn{11}{@{}l}{\textit{Base: the scene of each perturbation before the change happens}} \\
        \quad Camera shift & 61.1 & 72.2 & 44.4 & 88.9 & 33.3 & 60.6 & 59.1 & 59.1 & 49.5 & 70.2 \\
        \quad Distractor burst & 0.0 & 0.0 & 72.9 & 89.4 & --- & --- & 85.7 & 78.6 & 52.9 & 56.0 \\
        \quad Illumination switch & 47.6 & 66.7 & 52.6 & 78.9 & 42.2 & 55.6 & 26.7 & 40.0 & 42.3 & 60.3 \\
        \quad Obstacle insertion & 57.1 & 57.1 & 83.3 & 95.2 & 60.0 & 90.0 & 4.9 & 7.3 & 51.3 & 62.4 \\
        \quad Receptacle relocation & 65.2 & 69.6 & 70.0 & 86.7 & 60.0 & 70.0 & 21.4 & 42.9 & 54.2 & 67.3 \\
        \quad Sensor noise onset & 63.0 & 63.0 & 80.0 & 86.7 & 46.2 & 65.4 & 43.8 & 62.5 & 58.2 & 69.4 \\
        \quad Target relocation & 75.0 & 81.3 & 63.3 & 76.7 & 26.7 & 56.7 & 25.0 & 45.8 & 47.5 & 65.1 \\
        \quad Visual theme switch & 53.6 & 67.9 & 56.5 & 73.9 & 40.0 & 60.0 & 45.8 & 50.0 & 49.0 & 62.9 \\
        \addlinespace[2pt]
        All & 57.9 & 68.6 & 65.7 & 85.5 & 42.4 & 60.6 & 34.2 & 44.1 & 50.0 & 64.7 \\
        \midrule
        \multicolumn{11}{@{}l}{\textit{Dynamic: the change happens during the execution}} \\
        \quad Camera shift & 61.1 & 72.2 & 44.4 & 88.9 & 33.3 & 60.6 & 66.7 & 50.0 & 51.4 & 67.9 \\
        \quad Distractor burst & 0.0 & 0.0 & 64.7 & 78.8 & --- & --- & 85.7 & 78.6 & 50.1 & 52.5 \\
        \quad Illumination switch & 47.6 & 66.7 & 63.2 & 84.2 & 42.2 & 53.3 & 26.7 & 40.0 & 44.9 & 61.1 \\
        \quad Obstacle insertion & 42.9 & 42.9 & 85.7 & 90.5 & 60.0 & 80.0 & 4.9 & 9.8 & 48.4 & 55.8 \\
        \quad Receptacle relocation & 2.2 & 34.8 & 33.3 & 83.3 & 0.0 & 50.0 & 0.0 & 14.3 & 8.9 & 45.6 \\
        \quad Sensor noise onset & 63.0 & 48.1 & 86.7 & 80.0 & 46.2 & 65.4 & 46.9 & 46.9 & 60.7 & 60.1 \\
        \quad Target relocation & 0.0 & 18.8 & 3.3 & 33.3 & 10.0 & 33.3 & 8.3 & 4.2 & 5.4 & 22.4 \\
        \quad Visual theme switch & 53.6 & 67.9 & 60.9 & 73.9 & 40.0 & 60.0 & 41.7 & 54.2 & 49.0 & 64.0 \\
        \addlinespace[2pt]
        All & 36.2 & 51.9 & 54.2 & 75.7 & 36.4 & 55.8 & 31.4 & 33.8 & 39.6 & 54.3 \\
        \bottomrule
    \end{tabular}
    \endgroup
\end{table*}

\begin{table*}[t]
    \centering
    \caption{\textbf{GaP: LIBERO-Plus per perturbation and level.} Success (\%) of the policy of GaP alone (Bob) and after the adversarial game (+Alice) at each of the five difficulty levels of LIBERO-Plus.}
    \label{tab:adv-plus-levels-gap}
    \begingroup
    \footnotesize
    \setlength{\tabcolsep}{4pt}
    \renewcommand{\arraystretch}{1.06}
    \begin{tabular}{@{}lrrrrrrrrrrrr@{}}
        \toprule
        \textbf{Perturbation} & \multicolumn{2}{c}{Level 1} & \multicolumn{2}{c}{Level 2} & \multicolumn{2}{c}{Level 3} & \multicolumn{2}{c}{Level 4} & \multicolumn{2}{c}{Level 5} & \multicolumn{2}{c}{All levels} \\
        \cmidrule(lr){2-3}\cmidrule(lr){4-5}\cmidrule(lr){6-7}\cmidrule(lr){8-9}\cmidrule(lr){10-11}\cmidrule(lr){12-13}
         & Bob & +Alice & Bob & +Alice & Bob & +Alice & Bob & +Alice & Bob & +Alice & Bob & +Alice \\
        \midrule
        Background textures & 66.7 & 80.3 & 66.7 & 81.9 & 73.6 & 84.7 & 58.6 & 77.6 & 58.8 & 64.7 & 66.7 & 78.4 \\
        Camera viewpoints & 66.1 & 74.2 & 58.3 & 75.0 & 50.0 & 72.2 & 37.1 & 64.5 & 46.2 & 59.6 & 54.1 & 70.0 \\
        Language instructions & 71.6 & 79.7 & 74.4 & 80.8 & 70.5 & 80.8 & 71.3 & 81.3 & 68.1 & 79.2 & 71.5 & 80.3 \\
        Light conditions & 71.7 & 80.4 & 70.0 & 71.7 & 71.2 & 78.8 & 71.0 & 77.4 & 69.6 & 78.3 & 72.9 & 77.8 \\
        Objects layout & 57.8 & 62.5 & 56.9 & 66.7 & 67.1 & 64.5 & 50.0 & 56.7 & 53.3 & 58.3 & 58.5 & 63.4 \\
        Robot initial states & 71.1 & 76.3 & 70.0 & 81.3 & 62.5 & 77.5 & 66.7 & 79.5 & 61.3 & 70.0 & 65.8 & 76.5 \\
        Sensor noise & 25.0 & 52.8 & 11.3 & 41.3 & 10.0 & 43.8 & 13.8 & 45.0 & 10.3 & 50.0 & 14.3 & 46.2 \\
        \addlinespace[2pt]
        All perturbations & 60.1 & 70.8 & 56.6 & 70.4 & 56.3 & 70.9 & 52.2 & 68.3 & 55.2 & 66.9 & 56.1 & 69.8 \\
        \bottomrule
    \end{tabular}
    \endgroup
\end{table*}

\begin{table*}[t]
    \centering
    \caption{\textbf{\method{}-Bob: LIBERO-Pro and LIBERO-Plus per perturbation.} Success (\%) of the policy of \method{}-Bob alone (Bob) and after the adversarial game (+Alice), per suite. LIBERO-Pro has four original perturbations and six extensions added by the community to its official repository. Each value is the mean over the tasks of the suite.}
    \label{tab:adv-pro-plus-rally}
    \begingroup
    \footnotesize
    \setlength{\tabcolsep}{4pt}
    \renewcommand{\arraystretch}{1.06}
    \begin{tabular}{@{}lrrrrrrrrrr@{}}
        \toprule
        \textbf{Perturbation} & \multicolumn{2}{c}{Spatial} & \multicolumn{2}{c}{Object} & \multicolumn{2}{c}{Goal} & \multicolumn{2}{c}{LIBERO-10} & \multicolumn{2}{c}{Avg.} \\
        \cmidrule(lr){2-3}\cmidrule(lr){4-5}\cmidrule(lr){6-7}\cmidrule(lr){8-9}\cmidrule(lr){10-11}
         & Bob & +Alice & Bob & +Alice & Bob & +Alice & Bob & +Alice & Bob & +Alice \\
        \midrule
        \multicolumn{11}{@{}l}{\textit{LIBERO-Pro}} \\
        \quad Object & 94.0 & 100.0 & 64.0 & 70.0 & 76.0 & 94.0 & 42.0 & 56.0 & 69.0 & 80.0 \\
        \quad Position & 76.0 & 90.0 & 76.0 & 90.0 & 56.0 & 70.0 & 34.0 & 68.0 & 60.5 & 79.5 \\
        \quad Semantic & 94.0 & 98.0 & 98.0 & 100.0 & 100.0 & 100.0 & 100.0 & 98.0 & 98.0 & 99.0 \\
        \quad Task & 0.0 & 2.0 & 0.0 & 0.0 & 10.0 & 10.0 & 10.0 & 10.0 & 5.0 & 5.5 \\
        \quad All original & 66.0 & 72.5 & 59.5 & 65.0 & 60.5 & 68.5 & 46.5 & 58.0 & 58.1 & 66.0 \\
        \addlinespace[2pt]
        \quad Camera view angle & 0.0 & 92.0 & 0.0 & 72.0 & 60.0 & 86.0 & 6.0 & 64.0 & 16.5 & 78.5 \\
        \quad Initial pose & 18.0 & 52.0 & 12.0 & 48.0 & 28.0 & 78.0 & 16.0 & 56.0 & 18.5 & 58.5 \\
        \quad Object shape & 94.0 & 100.0 & 98.0 & 100.0 & 92.0 & 100.0 & 96.0 & 98.0 & 95.0 & 99.5 \\
        \quad Object texture & 94.0 & 100.0 & 100.0 & 100.0 & 82.0 & 96.0 & 84.0 & 100.0 & 90.0 & 99.0 \\
        \quad View occlusion & 14.0 & 34.0 & 10.0 & 78.0 & 30.0 & 90.0 & 16.0 & 84.0 & 17.5 & 71.5 \\
        \quad Visual noise and glare & 90.0 & 100.0 & 86.0 & 96.0 & 80.0 & 96.0 & 76.0 & 98.0 & 83.0 & 97.5 \\
        \quad All extensions & 51.7 & 79.7 & 51.0 & 82.3 & 62.0 & 91.0 & 49.0 & 83.3 & 53.4 & 84.1 \\
        \addlinespace[2pt]
        All & 57.4 & 76.8 & 54.4 & 75.4 & 61.4 & 82.0 & 48.0 & 73.2 & 55.3 & 76.9 \\
        \midrule
        \multicolumn{11}{@{}l}{\textit{LIBERO-Plus}} \\
        \quad Background textures & 98.6 & 100.0 & 98.3 & 100.0 & 97.0 & 99.0 & 87.9 & 95.6 & 95.5 & 98.6 \\
        \quad Camera viewpoints & 80.9 & 96.5 & 87.4 & 96.1 & 88.1 & 93.5 & 76.3 & 89.2 & 83.2 & 93.8 \\
        \quad Language instructions & 100.0 & 100.0 & 100.0 & 100.0 & 100.0 & 100.0 & 90.0 & 100.0 & 97.5 & 100.0 \\
        \quad Light conditions & 100.0 & 100.0 & 100.0 & 100.0 & 95.0 & 100.0 & 81.6 & 100.0 & 94.2 & 100.0 \\
        \quad Objects layout & 91.1 & 100.0 & 95.0 & 91.0 & 53.0 & 64.3 & 86.8 & 88.7 & 81.5 & 86.0 \\
        \quad Robot initial states & 74.4 & 84.0 & 59.0 & 70.0 & 80.6 & 85.8 & 74.5 & 85.0 & 72.1 & 81.2 \\
        \quad Sensor noise & 33.7 & 89.9 & 21.0 & 76.0 & 42.8 & 78.6 & 25.6 & 76.0 & 30.8 & 80.1 \\
        \addlinespace[2pt]
        All & 80.9 & 95.2 & 76.9 & 88.8 & 77.4 & 87.2 & 73.5 & 90.6 & 77.2 & 90.5 \\
        \bottomrule
    \end{tabular}
    \endgroup
\end{table*}

\begin{table*}[t]
    \centering
    \caption{\textbf{\method{}-Bob: LIBERO-MAX per perturbation.} Success (\%) of the policy of \method{}-Bob alone (Bob) and after the adversarial game (+Alice). Every scene is run twice, without its change (Base) and with it (Dynamic). A perturbation row pools the executions of all tasks of the suite, since a task has few scenes of one perturbation; the rows All are means over the tasks.}
    \label{tab:adv-max-rally}
    \begingroup
    \footnotesize
    \setlength{\tabcolsep}{4pt}
    \renewcommand{\arraystretch}{1.06}
    \begin{tabular}{@{}lrrrrrrrrrr@{}}
        \toprule
        \textbf{Perturbation} & \multicolumn{2}{c}{Spatial} & \multicolumn{2}{c}{Object} & \multicolumn{2}{c}{Goal} & \multicolumn{2}{c}{LIBERO-10} & \multicolumn{2}{c}{Avg.} \\
        \cmidrule(lr){2-3}\cmidrule(lr){4-5}\cmidrule(lr){6-7}\cmidrule(lr){8-9}\cmidrule(lr){10-11}
         & Bob & +Alice & Bob & +Alice & Bob & +Alice & Bob & +Alice & Bob & +Alice \\
        \midrule
        \multicolumn{11}{@{}l}{\textit{Base: the scene of each perturbation before the change happens}} \\
        \quad Camera shift & 86.1 & 94.4 & 55.6 & 88.9 & 63.6 & 90.9 & 81.8 & 95.5 & 71.8 & 92.4 \\
        \quad Distractor burst & 100.0 & 0.0 & 76.5 & 91.8 & --- & --- & 50.0 & 85.7 & 75.5 & 59.2 \\
        \quad Illumination switch & 57.1 & 85.7 & 84.2 & 94.7 & 73.3 & 91.1 & 73.3 & 93.3 & 72.0 & 91.2 \\
        \quad Obstacle insertion & 57.1 & 100.0 & 64.3 & 81.0 & 100.0 & 100.0 & 63.4 & 78.0 & 71.2 & 89.8 \\
        \quad Receptacle relocation & 71.7 & 80.4 & 66.7 & 86.7 & 50.0 & 80.0 & 64.3 & 71.4 & 63.2 & 79.6 \\
        \quad Sensor noise onset & 63.0 & 85.2 & 86.7 & 86.7 & 76.9 & 96.2 & 75.0 & 87.5 & 75.4 & 88.9 \\
        \quad Target relocation & 87.5 & 93.8 & 83.3 & 96.7 & 63.3 & 76.7 & 62.5 & 91.7 & 74.2 & 89.7 \\
        \quad Visual theme switch & 64.3 & 85.7 & 69.6 & 73.9 & 68.0 & 80.0 & 66.7 & 87.5 & 67.1 & 81.8 \\
        \addlinespace[2pt]
        All & 71.7 & 85.9 & 74.2 & 87.0 & 70.8 & 88.3 & 67.7 & 86.2 & 71.1 & 86.9 \\
        \midrule
        \multicolumn{11}{@{}l}{\textit{Dynamic: the change happens during the execution}} \\
        \quad Camera shift & 83.3 & 94.4 & 55.6 & 55.6 & 66.7 & 84.8 & 72.7 & 86.4 & 69.6 & 80.3 \\
        \quad Distractor burst & 100.0 & 0.0 & 62.4 & 82.4 & --- & --- & 50.0 & 85.7 & 70.8 & 56.0 \\
        \quad Illumination switch & 57.1 & 81.0 & 84.2 & 100.0 & 75.6 & 91.1 & 80.0 & 93.3 & 74.2 & 91.3 \\
        \quad Obstacle insertion & 42.9 & 100.0 & 64.3 & 85.7 & 100.0 & 100.0 & 34.1 & 53.7 & 60.3 & 84.8 \\
        \quad Receptacle relocation & 2.2 & 54.3 & 40.0 & 90.0 & 0.0 & 50.0 & 0.0 & 42.9 & 10.5 & 59.3 \\
        \quad Sensor noise onset & 63.0 & 77.8 & 80.0 & 80.0 & 73.1 & 92.3 & 56.3 & 81.3 & 68.1 & 82.8 \\
        \quad Target relocation & 37.5 & 56.3 & 50.0 & 76.7 & 36.7 & 50.0 & 45.8 & 62.5 & 42.5 & 61.4 \\
        \quad Visual theme switch & 67.9 & 89.3 & 69.6 & 78.3 & 68.0 & 84.0 & 75.0 & 83.3 & 70.1 & 83.7 \\
        \addlinespace[2pt]
        All & 48.9 & 74.8 & 61.5 & 82.3 & 65.5 & 81.8 & 52.4 & 72.2 & 57.1 & 77.8 \\
        \bottomrule
    \end{tabular}
    \endgroup
\end{table*}

\begin{table*}[t]
    \centering
    \caption{\textbf{\method{}-Bob: LIBERO-Plus per perturbation and level.} Success (\%) of the policy of \method{}-Bob alone (Bob) and after the adversarial game (+Alice) at each of the five difficulty levels of LIBERO-Plus.}
    \label{tab:adv-plus-levels-rally}
    \begingroup
    \footnotesize
    \setlength{\tabcolsep}{4pt}
    \renewcommand{\arraystretch}{1.06}
    \begin{tabular}{@{}lrrrrrrrrrrrr@{}}
        \toprule
        \textbf{Perturbation} & \multicolumn{2}{c}{Level 1} & \multicolumn{2}{c}{Level 2} & \multicolumn{2}{c}{Level 3} & \multicolumn{2}{c}{Level 4} & \multicolumn{2}{c}{Level 5} & \multicolumn{2}{c}{All levels} \\
        \cmidrule(lr){2-3}\cmidrule(lr){4-5}\cmidrule(lr){6-7}\cmidrule(lr){8-9}\cmidrule(lr){10-11}\cmidrule(lr){12-13}
         & Bob & +Alice & Bob & +Alice & Bob & +Alice & Bob & +Alice & Bob & +Alice & Bob & +Alice \\
        \midrule
        Background textures & 93.9 & 98.5 & 98.6 & 100.0 & 95.8 & 98.6 & 96.6 & 100.0 & 88.2 & 97.1 & 95.5 & 98.6 \\
        Camera viewpoints & 90.3 & 95.2 & 87.5 & 94.4 & 83.3 & 97.2 & 74.2 & 91.9 & 73.1 & 94.2 & 83.2 & 93.8 \\
        Language instructions & 97.3 & 100.0 & 97.4 & 100.0 & 97.4 & 100.0 & 97.5 & 100.0 & 97.2 & 100.0 & 97.5 & 100.0 \\
        Light conditions & 95.7 & 100.0 & 95.0 & 100.0 & 95.5 & 100.0 & 93.5 & 100.0 & 91.3 & 100.0 & 94.2 & 100.0 \\
        Objects layout & 82.8 & 85.9 & 87.5 & 87.5 & 89.5 & 93.4 & 65.0 & 78.3 & 68.3 & 75.0 & 81.5 & 86.0 \\
        Robot initial states & 86.8 & 94.7 & 81.3 & 82.5 & 78.8 & 82.5 & 59.0 & 78.2 & 55.0 & 70.0 & 72.1 & 81.2 \\
        Sensor noise & 52.8 & 83.3 & 33.8 & 82.5 & 23.8 & 78.8 & 21.3 & 78.8 & 19.0 & 74.1 & 30.8 & 80.1 \\
        \addlinespace[2pt]
        All perturbations & 84.5 & 92.4 & 81.5 & 91.4 & 79.5 & 92.2 & 69.5 & 89.1 & 69.1 & 85.6 & 77.2 & 90.5 \\
        \bottomrule
    \end{tabular}
    \endgroup
\end{table*}

\subsection{Comparison with Vision-Language-Action Policies}
\label{sec:appendix:vla}
We compare the robustness of code policies, before and after refinement with adversarial Alice, with published vision-language-action (VLA) policy results on LIBERO-Pro, LIBERO-Plus, and LIBERO-MAX (Tables~\ref{tab:vla-pro} to~\ref{tab:vla-max}).
The VLA results come from the benchmark authors' evaluations of policies trained on LIBERO; the code-policy results come from our experiments with Sonnet 5.5 in Sec.~\ref{sec:appendix:adversarial}.
The quoted VLA results were obtained under their respective training and evaluation protocols.
As in Table~\ref{tab:main-results}, we report Bob alone and Bob with adversarial Alice for each harness.

\minisection{Results}
After refinement with adversarial Alice, \method{}-Bob achieves higher reported aggregate success than the VLA policies listed here on each benchmark: 66.0\% on the original LIBERO-Pro perturbations, 90.5\% on LIBERO-Plus, and 77.8\% under the dynamic events of LIBERO-MAX.
The strongest listed VLA results are 53.3\%, 69.6\%, and 66.9\%, respectively.
\begin{table*}[t]
    \centering
    \caption{\textbf{Comparison with VLA policies on LIBERO-Pro.} Success (\%) on the four original perturbations. The VLA rows come from the benchmark authors' \href{https://github.com/Zxy-MLlab/LIBERO-PRO\#-libero-pro-model-leaderboard}{official leaderboard}~\cite{liberopro2025}; each perturbation score averages the four reported suite scores. Our rows reproduce the equal-task macro averages in Sec.~\ref{sec:appendix:adversarial}. Mean averages the four original perturbations, whereas the PRO column of Table~\ref{tab:main-results} also includes six community extensions.}
    \label{tab:vla-pro}
    \begingroup
    \footnotesize
    \setlength{\tabcolsep}{4pt}
    \renewcommand{\arraystretch}{1.06}
    \begin{tabularx}{\textwidth}{@{}l*{5}{>{\raggedleft\arraybackslash}X}@{}}
        \toprule
        \textbf{Policy} & Object & Position & Semantic & Task & Mean \\
        \midrule
        \multicolumn{6}{@{}l}{\textit{VLA policies: benchmark authors' evaluations}~\cite{liberopro2025}} \\
        $\pi_{0.5}$ & 96.0 & 20.8 & 95.8 & 0.8 & 53.3 \\
        OpenVLA & 93.0 & 0.0 & 97.3 & 0.0 & 47.6 \\
        $\pi_0$ & 90.5 & 0.0 & 90.5 & 0.0 & 45.3 \\
        \midrule
        \multicolumn{6}{@{}l}{\textit{Code policies: our evaluations with Sonnet 5.5}} \\
        ASPIRE: Bob only & 63.5 & 58.0 & 92.5 & 2.5 & 54.1 \\
        \quad + Alice & \textbf{79.0} & \textbf{77.5} & \textbf{97.0} & \textbf{3.0} & \textbf{64.1} \\
        RHO: Bob only & 59.0 & 31.0 & 73.0 & 1.5 & 41.1 \\
        \quad + Alice & \textbf{67.5} & \textbf{58.0} & \textbf{83.5} & \textbf{2.5} & \textbf{52.9} \\
        GaP: Bob only & 44.5 & 47.5 & 72.5 & 0.5 & 41.3 \\
        \quad + Alice & \textbf{66.0} & \textbf{63.5} & \textbf{80.5} & \textbf{1.0} & \textbf{52.8} \\
        \method{}-Bob: Bob only & 69.0 & 60.5 & 98.0 & 5.0 & 58.1 \\
        \quad + Alice & \textbf{80.0} & \textbf{79.5} & \textbf{99.0} & \textbf{5.5} & \textbf{66.0} \\
        \bottomrule
    \end{tabularx}
    \endgroup
\end{table*}

\begin{table*}[t]
    \centering
    \caption{\textbf{Comparison with VLA policies on LIBERO-Plus.} Success (\%) per perturbation category and overall. VLA category scores and Overall are reproduced directly from Table~2 of LIBERO-Plus~\cite{liberoplus}; OpenVLA-OFT is their suite-specific variant. Our results are from Sec.~\ref{sec:appendix:adversarial} and use equal-task macro averaging. For Overall, we compute each task's success rate over its evaluated cases across all seven categories, then average equally over tasks.}
    \label{tab:vla-plus}
    \begingroup
    \footnotesize
    \setlength{\tabcolsep}{3pt}
    \renewcommand{\arraystretch}{1.06}
    \begin{tabularx}{\textwidth}{@{}l*{8}{>{\raggedleft\arraybackslash}X}@{}}
        \toprule
        \textbf{Policy} & Camera & Robot & Language & Light & Background & Noise & Layout & Overall \\
        \midrule
        \multicolumn{9}{@{}l}{\textit{VLA policies: benchmark authors' evaluations}~\cite{liberoplus}} \\
        OpenVLA-OFT & 56.4 & 31.9 & 79.5 & 88.7 & 93.3 & 75.8 & 74.2 & 69.6 \\
        RIPT-VLA & 55.2 & 31.2 & 77.6 & 88.4 & 91.6 & 73.5 & 74.2 & 68.4 \\
        $\pi_0$-FAST & 65.1 & 21.6 & 61.0 & 73.2 & 73.2 & 74.4 & 68.8 & 61.6 \\
        $\pi_0$ & 13.8 & 6.0 & 58.8 & 85.0 & 81.4 & 79.0 & 68.9 & 53.6 \\
        OpenVLA & 0.8 & 3.5 & 23.0 & 8.1 & 34.8 & 15.2 & 28.5 & 15.6 \\
        \midrule
        \multicolumn{9}{@{}l}{\textit{Code policies: our evaluations with Sonnet 5.5}} \\
        ASPIRE: Bob only & 79.9 & 70.1 & 90.0 & 87.7 & 83.5 & 30.0 & 79.1 & 72.9 \\
        \quad + Alice & \textbf{94.7} & \textbf{85.2} & \textbf{100.0} & \textbf{100.0} & \textbf{96.5} & \textbf{78.0} & \textbf{85.0} & \textbf{90.5} \\
        RHO: Bob only & 52.7 & 69.2 & 77.5 & 73.0 & 68.7 & 15.0 & 54.8 & 57.7 \\
        \quad + Alice & \textbf{73.2} & \textbf{75.0} & \textbf{87.5} & \textbf{86.9} & \textbf{82.8} & \textbf{35.4} & \textbf{67.0} & \textbf{71.5} \\
        GaP: Bob only & 54.1 & 65.8 & 71.5 & 72.9 & 66.7 & 14.3 & 58.5 & 56.1 \\
        \quad + Alice & \textbf{70.0} & \textbf{76.5} & \textbf{80.3} & \textbf{77.8} & \textbf{78.4} & \textbf{46.2} & \textbf{63.4} & \textbf{69.8} \\
        \method{}-Bob: Bob only & 83.2 & 72.1 & 97.5 & 94.2 & 95.5 & 30.8 & 81.5 & 77.2 \\
        \quad + Alice & \textbf{93.8} & \textbf{81.2} & \textbf{100.0} & \textbf{100.0} & \textbf{98.6} & \textbf{80.1} & \textbf{86.0} & \textbf{90.5} \\
        \bottomrule
    \end{tabularx}
    \endgroup
\end{table*}

\begin{table*}[t]
    \centering
    \caption{\textbf{Comparison with VLA policies on LIBERO-MAX.} Success (\%) without (Base) and with (Dynamic) a mid-execution event. The VLA rows are the benchmark authors' results over 8,000 paired cases in Table~2 of LIBERO-MAX~\cite{liberomax}; OpenVLA-OFT uses their checkpoint trained jointly on the four LIBERO suites. Our rows reproduce the task means in Sec.~\ref{sec:appendix:adversarial}.}
    \label{tab:vla-max}
    \begingroup
    \footnotesize
    \setlength{\tabcolsep}{4pt}
    \renewcommand{\arraystretch}{1.06}
    \begin{tabularx}{0.6\textwidth}{@{}l*{2}{>{\raggedleft\arraybackslash}X}@{}}
        \toprule
        \textbf{Policy} & Base & Dynamic \\
        \midrule
        \multicolumn{3}{@{}l}{\textit{VLA policies: benchmark authors' evaluations}~\cite{liberomax}} \\
        $\pi_{0.5}$ & 79.7 & 65.7 \\
        OpenVLA-OFT & 64.3 & 43.2 \\
        MolmoAct2 & 80.3 & 66.9 \\
        X-VLA & 62.6 & 37.7 \\
        \midrule
        \multicolumn{3}{@{}l}{\textit{Code policies: our evaluations with Sonnet 5.5}} \\
        ASPIRE: Bob only & 66.2 & 48.2 \\
        \quad + Alice & \textbf{84.6} & \textbf{72.6} \\
        RHO: Bob only & 53.4 & 37.5 \\
        \quad + Alice & \textbf{68.4} & \textbf{52.8} \\
        GaP: Bob only & 50.0 & 39.6 \\
        \quad + Alice & \textbf{64.7} & \textbf{54.3} \\
        \method{}-Bob: Bob only & 71.1 & 57.1 \\
        \quad + Alice & \textbf{86.9} & \textbf{77.8} \\
        \bottomrule
    \end{tabularx}
    \endgroup
\end{table*}

\subsection{Cooperative Alice: Results per Task}
\label{sec:appendix:cooperative}
Tables~\ref{tab:coop-pertask-spatial} to~\ref{tab:coop-pertask-long} give the held-out success of every policy of Tables~\ref{tab:cooperative-main} and~\ref{tab:cooperative-curricula} per task.
Table~\ref{tab:coop-pertask-fable} does the same for the runs with Fable 5.2 of Table~\ref{tab:cooperative-fable}.

\begin{figure}[t]
    \centering
    \includegraphics[width=\linewidth]{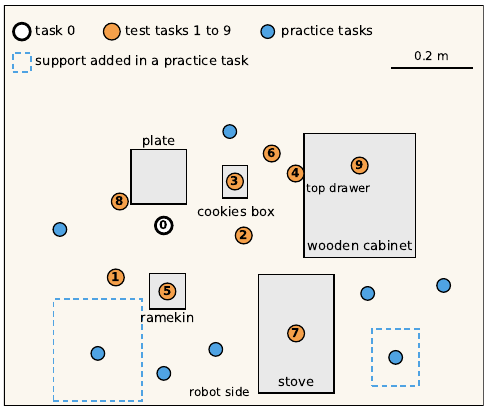}
    \caption{\textbf{LIBERO-Spatial: where the target bowl starts.} Top view of the table, to scale, with the robot at the bottom. Orange: the nine test tasks, numbered as in Table~\ref{tab:coop-pertask-spatial}. White: task 0. Blue: the eight practice tasks. A dashed outline is a support that is added to the scene for a practice task.}
    \label{fig:coop-spatial-layout}
\end{figure}

\minisection{Practice tasks of the cooperative Alice}
Tables~\ref{tab:coop-tasks-object} to~\ref{tab:coop-tasks-long} show, next to the nine test tasks of a suite, examples of the tasks that Alice proposes and that are accepted, with ASPIRE as Bob (Table~\ref{tab:main-results}).
Each proposed task differs from every test task in the objects involved, the success condition, or the sub-tasks, as the checker requires (Appendix~\ref{app:refinement_cycle}).
An object that LIBERO gives a different name counts as a different object.
In Object, the white bottle is a second bottle of wine, with another label than the wine bottle; it is not the salad dressing, the bbq sauce, or the ketchup of the test tasks.
In LIBERO-10, each example chains two sub-tasks, as most test tasks do, and shares at most one sub-task with any test task.
Spatial has no table, since its tasks differ only in the layout (Appendix~\ref{sec:appendix:settings}); Figure~\ref{fig:coop-spatial-layout} shows where the target bowl starts in task 0, in the nine test tasks, and in the practice tasks.

\begin{table*}[t]
    \centering
    \caption{\textbf{LIBERO-Object: test tasks and examples of the tasks proposed by the cooperative Alice.} Left: the nine test tasks of the suite, which no cycle uses. Right: examples of the proposed tasks that are accepted, with ASPIRE as Bob (Table~\ref{tab:main-results}). The two lists are not aligned row by row.}
    \label{tab:coop-tasks-object}
    \begingroup
    \footnotesize
    \setlength{\tabcolsep}{5pt}
    \renewcommand{\arraystretch}{1.06}
    \begin{tabularx}{\textwidth}{@{}r>{\raggedright\arraybackslash}X>{\raggedright\arraybackslash}X@{}}
        \toprule
        \textbf{Task} & \textbf{Test task} & \textbf{Example of a task proposed by Alice} \\
        \midrule
        1 & Pick the cream cheese and place it in the basket & Pick the wine bottle and place it in the basket \\
        2 & Pick the salad dressing and place it in the basket & Pick the cookies and place it in the basket \\
        3 & Pick the bbq sauce and place it in the basket & Pick the white bottle and place it in the basket \\
        4 & Pick the ketchup and place it in the basket & Pick the ramekin and place it in the basket \\
        5 & Pick the tomato sauce and place it in the basket & Pick the popcorn and place it in the basket \\
        6 & Pick the butter and place it in the basket & Pick the macaroni and cheese and place it in the basket \\
        7 & Pick the milk and place it in the basket & Pick the white bowl and place it in the basket \\
        8 & Pick the chocolate pudding and place it in the basket & Pick the moka pot and place it in the basket \\
        9 & Pick the orange juice and place it in the basket &  \\
        \bottomrule
    \end{tabularx}
    \endgroup
\end{table*}

\begin{table*}[t]
    \centering
    \caption{\textbf{LIBERO-Goal: test tasks and examples of the tasks proposed by the cooperative Alice.} Left: the nine test tasks of the suite, which no cycle uses. Right: examples of the proposed tasks that are accepted, with ASPIRE as Bob (Table~\ref{tab:main-results}). The two lists are not aligned row by row.}
    \label{tab:coop-tasks-goal}
    \begingroup
    \footnotesize
    \setlength{\tabcolsep}{5pt}
    \renewcommand{\arraystretch}{1.06}
    \begin{tabularx}{\textwidth}{@{}r>{\raggedright\arraybackslash}X>{\raggedright\arraybackslash}X@{}}
        \toprule
        \textbf{Task} & \textbf{Test task} & \textbf{Example of a task proposed by Alice} \\
        \midrule
        1 & Put the bowl on the stove & Put the plate on the top of the shelf \\
        2 & Put the wine bottle on the top of the drawer & Turn on the faucet \\
        3 & Open the top layer of the drawer and put the bowl inside & Close the top layer of the drawer \\
        4 & Put the bowl on the top of the drawer & Put the black book in the front compartment of the caddy \\
        5 & Push the plate to the front of the stove & Open the microwave \\
        6 & Put the cream cheese on the bowl & Close the microwave \\
        7 & Turn on the stove & Put the bowl on the top of the microwave \\
        8 & Put the bowl on the plate & Put the bowl on the top of the shelf \\
        9 & Put the wine bottle on the rack & Turn off the stove \\
        \bottomrule
    \end{tabularx}
    \endgroup
\end{table*}

\begin{table*}[t]
    \centering
    \caption{\textbf{LIBERO-10: test tasks and examples of the tasks proposed by the cooperative Alice.} Left: the nine test tasks of the suite, which no cycle uses. Right: examples of the proposed tasks that are accepted, with ASPIRE as Bob (Table~\ref{tab:main-results}). The two lists are not aligned row by row. The wording of four long proposed tasks is shortened.}
    \label{tab:coop-tasks-long}
    \begingroup
    \footnotesize
    \setlength{\tabcolsep}{5pt}
    \renewcommand{\arraystretch}{1.06}
    \begin{tabularx}{\textwidth}{@{}r>{\raggedright\arraybackslash}X>{\raggedright\arraybackslash}X@{}}
        \toprule
        \textbf{Task} & \textbf{Test task} & \textbf{Example of a task proposed by Alice} \\
        \midrule
        1 & Put both the cream cheese box and the butter in the basket & Turn off the stove and put the black bowl on top of the cabinet \\
        2 & Turn on the stove and put the moka pot on it & Open the top drawer of the cabinet and put the ketchup inside \\
        3 & Put the black bowl in the bottom drawer of the cabinet and close it & Open the microwave and put the black bowl on top of the microwave \\
        4 & Put the white mug on the left plate and put the yellow and white mug on the right plate & Put the wine bottle on the rack and close the bottom drawer \\
        5 & Pick up the book and place it in the back compartment of the caddy & Put the milk in the basket and put the red mug on the plate \\
        6 & Put the white mug on the plate and put the chocolate pudding to the right of the plate & Put the orange juice in the basket and the ketchup left of the plate \\
        7 & Put both the alphabet soup and the cream cheese box in the basket & Open the middle drawer and put the butter on top of the cabinet \\
        8 & Put both moka pots on the stove & Put the butter on the plate and turn off the stove \\
        9 & Put the yellow and white mug in the microwave and close it & Put the red mug in the microwave and turn on the stove \\
        \bottomrule
    \end{tabularx}
    \endgroup
\end{table*}

\begin{table*}[t]
    \centering
    \caption{\textbf{LIBERO-Spatial: held-out success per task.} Successes of 15 start states (25 to 39) of each task, for the final policy of every curriculum. Task 0 is the only task that Bob refines on and is not counted in the last row. Every instruction is ``Pick the akita black bowl \ldots{} and place it on the plate''; the column gives where the bowl is.}
    \label{tab:coop-pertask-spatial}
    \begingroup
    \footnotesize
    \setlength{\tabcolsep}{4pt}
    \renewcommand{\arraystretch}{1.06}
    \begin{tabularx}{\textwidth}{@{}r>{\raggedright\arraybackslash}Xrrrrrrrrrrrr@{}}
        \toprule
        & & \multicolumn{6}{c}{ASPIRE as Bob} & \multicolumn{2}{c}{RHO as Bob} & \multicolumn{2}{c}{GaP as Bob} & \multicolumn{2}{c}{\method{}-Bob} \\
        \cmidrule(lr){3-8}\cmidrule(lr){9-10}\cmidrule(lr){11-12}\cmidrule(l){13-14}
        \textbf{Task} & \textbf{Instruction} & \textbf{Task 0} & \textbf{Random} & \textbf{TSCL} & \textbf{OMNI-EPIC} & \textbf{Traj.-only} & \textbf{Alice} & \textbf{Task 0} & \textbf{Alice} & \textbf{Task 0} & \textbf{Alice} & \textbf{Task 0} & \textbf{Alice} \\
        \midrule
        0 & Bowl between the plate and the ramekin (first task) & 14 & 14 & 14 & 15 & 15 & 15 & 15 & 15 & 14 & 14 & 15 & 15 \\
        \midrule
        1 & Bowl next to the ramekin & 14 & 14 & 14 & 14 & 14 & 14 & 1 & 14 & 15 & 15 & 15 & 15 \\
        2 & Bowl from table center & 0 & 0 & 7 & 0 & 15 & 15 & 0 & 15 & 0 & 15 & 0 & 15 \\
        3 & Bowl on the cookies box & 11 & 11 & 11 & 15 & 0 & 13 & 15 & 15 & 15 & 15 & 13 & 15 \\
        4 & Bowl in the top layer of the wooden cabinet & 0 & 5 & 15 & 0 & 0 & 14 & 0 & 0 & 0 & 0 & 0 & 15 \\
        5 & Bowl on the ramekin & 10 & 10 & 10 & 10 & 7 & 10 & 1 & 15 & 0 & 15 & 9 & 10 \\
        6 & Bowl next to the cookies box & 4 & 4 & 4 & 2 & 15 & 15 & 0 & 15 & 14 & 14 & 0 & 15 \\
        7 & Bowl on the stove & 1 & 14 & 1 & 14 & 0 & 14 & 0 & 10 & 15 & 15 & 6 & 15 \\
        8 & Bowl next to the plate & 14 & 14 & 13 & 14 & 15 & 15 & 14 & 15 & 9 & 15 & 12 & 15 \\
        9 & Bowl on the wooden cabinet & 0 & 0 & 14 & 12 & 10 & 14 & 0 & 9 & 0 & 4 & 0 & 8 \\
        \midrule
        \multicolumn{2}{@{}l}{Tasks 1 to 9 (\%)} & 40.0 & 53.3 & 65.9 & 60.0 & 56.3 & 91.9 & 23.0 & 80.0 & 50.4 & 80.0 & 40.7 & 91.1 \\
        \bottomrule
    \end{tabularx}
    \endgroup
\end{table*}

\begin{table*}[t]
    \centering
    \caption{\textbf{LIBERO-Object: held-out success per task.} Successes of 15 start states (25 to 39) of each task, for the final policy of every curriculum. Task 0 is the only task that Bob refines on and is not counted in the last row. Every instruction is ``Pick the \ldots{} and place it in the basket''; the column gives the object. Random and OMNI-EPIC are separate runs with different curricula and different final programs that succeed on the same start states.}
    \label{tab:coop-pertask-object}
    \begingroup
    \footnotesize
    \setlength{\tabcolsep}{4pt}
    \renewcommand{\arraystretch}{1.06}
    \begin{tabularx}{\textwidth}{@{}r>{\raggedright\arraybackslash}Xrrrrrrrrrrrr@{}}
        \toprule
        & & \multicolumn{6}{c}{ASPIRE as Bob} & \multicolumn{2}{c}{RHO as Bob} & \multicolumn{2}{c}{GaP as Bob} & \multicolumn{2}{c}{\method{}-Bob} \\
        \cmidrule(lr){3-8}\cmidrule(lr){9-10}\cmidrule(lr){11-12}\cmidrule(l){13-14}
        \textbf{Task} & \textbf{Instruction} & \textbf{Task 0} & \textbf{Random} & \textbf{TSCL} & \textbf{OMNI-EPIC} & \textbf{Traj.-only} & \textbf{Alice} & \textbf{Task 0} & \textbf{Alice} & \textbf{Task 0} & \textbf{Alice} & \textbf{Task 0} & \textbf{Alice} \\
        \midrule
        0 & Alphabet soup (first task) & 15 & 15 & 15 & 15 & 15 & 10 & 15 & 15 & 14 & 13 & 14 & 14 \\
        \midrule
        1 & Cream cheese & 0 & 0 & 0 & 0 & 0 & 0 & 0 & 0 & 0 & 2 & 0 & 0 \\
        2 & Salad dressing & 0 & 11 & 11 & 11 & 15 & 15 & 0 & 15 & 0 & 0 & 0 & 15 \\
        3 & Bbq sauce & 0 & 2 & 2 & 2 & 1 & 0 & 0 & 0 & 0 & 15 & 0 & 15 \\
        4 & Ketchup & 0 & 7 & 7 & 7 & 9 & 15 & 0 & 9 & 0 & 15 & 0 & 15 \\
        5 & Tomato sauce & 14 & 0 & 0 & 0 & 0 & 15 & 0 & 0 & 15 & 15 & 15 & 14 \\
        6 & Butter & 0 & 1 & 1 & 1 & 1 & 11 & 0 & 15 & 0 & 14 & 0 & 15 \\
        7 & Milk & 0 & 15 & 15 & 15 & 15 & 15 & 0 & 15 & 0 & 15 & 0 & 10 \\
        8 & Chocolate pudding & 0 & 0 & 15 & 0 & 0 & 7 & 0 & 15 & 0 & 0 & 0 & 15 \\
        9 & Orange juice & 0 & 15 & 15 & 15 & 15 & 15 & 0 & 15 & 1 & 15 & 0 & 15 \\
        \midrule
        \multicolumn{2}{@{}l}{Tasks 1 to 9 (\%)} & 10.4 & 37.8 & 48.9 & 37.8 & 41.5 & 68.9 & 0.0 & 62.2 & 11.9 & 67.4 & 11.1 & 84.4 \\
        \bottomrule
    \end{tabularx}
    \endgroup
\end{table*}

\begin{table*}[t]
    \centering
    \caption{\textbf{LIBERO-Goal: held-out success per task.} Successes of 15 start states (25 to 39) of each task, for the final policy of every curriculum. Task 0 is the only task that Bob refines on and is not counted in the last row. }
    \label{tab:coop-pertask-goal}
    \begingroup
    \footnotesize
    \setlength{\tabcolsep}{4pt}
    \renewcommand{\arraystretch}{1.06}
    \begin{tabularx}{\textwidth}{@{}r>{\raggedright\arraybackslash}Xrrrrrrrrrrrr@{}}
        \toprule
        & & \multicolumn{6}{c}{ASPIRE as Bob} & \multicolumn{2}{c}{RHO as Bob} & \multicolumn{2}{c}{GaP as Bob} & \multicolumn{2}{c}{\method{}-Bob} \\
        \cmidrule(lr){3-8}\cmidrule(lr){9-10}\cmidrule(lr){11-12}\cmidrule(l){13-14}
        \textbf{Task} & \textbf{Instruction} & \textbf{Task 0} & \textbf{Random} & \textbf{TSCL} & \textbf{OMNI-EPIC} & \textbf{Traj.-only} & \textbf{Alice} & \textbf{Task 0} & \textbf{Alice} & \textbf{Task 0} & \textbf{Alice} & \textbf{Task 0} & \textbf{Alice} \\
        \midrule
        0 & Open the middle layer of the drawer (first task) & 15 & 15 & 15 & 15 & 15 & 15 & 15 & 15 & 15 & 15 & 15 & 15 \\
        \midrule
        1 & Put the bowl on the stove & 0 & 0 & 0 & 15 & 15 & 15 & 0 & 15 & 0 & 14 & 0 & 14 \\
        2 & Put the wine bottle on the top of the drawer & 0 & 0 & 3 & 0 & 0 & 0 & 0 & 15 & 0 & 0 & 0 & 8 \\
        3 & Open the top layer of the drawer and put the bowl inside & 0 & 0 & 0 & 0 & 0 & 0 & 0 & 0 & 0 & 0 & 0 & 0 \\
        4 & Put the bowl on the top of the drawer & 0 & 0 & 0 & 0 & 0 & 15 & 0 & 14 & 0 & 0 & 0 & 15 \\
        5 & Push the plate to the front of the stove & 0 & 0 & 0 & 0 & 0 & 7 & 0 & 0 & 0 & 0 & 0 & 0 \\
        6 & Put the cream cheese on the bowl & 0 & 0 & 0 & 1 & 0 & 15 & 0 & 15 & 0 & 0 & 0 & 15 \\
        7 & Turn on the stove & 0 & 13 & 0 & 15 & 6 & 15 & 0 & 0 & 0 & 0 & 0 & 15 \\
        8 & Put the bowl on the plate & 0 & 0 & 0 & 6 & 12 & 15 & 0 & 15 & 0 & 0 & 0 & 15 \\
        9 & Put the wine bottle on the rack & 0 & 6 & 8 & 0 & 0 & 0 & 0 & 0 & 0 & 0 & 0 & 14 \\
        \midrule
        \multicolumn{2}{@{}l}{Tasks 1 to 9 (\%)} & 0.0 & 14.1 & 8.1 & 27.4 & 24.4 & 60.7 & 0.0 & 54.8 & 0.0 & 10.4 & 0.0 & 71.1 \\
        \bottomrule
    \end{tabularx}
    \endgroup
\end{table*}

\begin{table*}[t]
    \centering
    \caption{\textbf{LIBERO-10: held-out success per task.} Successes of 15 start states (25 to 39) of each task, for the final policy of every curriculum. Task 0 is the only task that Bob refines on and is not counted in the last row. }
    \label{tab:coop-pertask-long}
    \begingroup
    \footnotesize
    \setlength{\tabcolsep}{4pt}
    \renewcommand{\arraystretch}{1.06}
    \begin{tabularx}{\textwidth}{@{}r>{\raggedright\arraybackslash}Xrrrrrrrrrrrr@{}}
        \toprule
        & & \multicolumn{6}{c}{ASPIRE as Bob} & \multicolumn{2}{c}{RHO as Bob} & \multicolumn{2}{c}{GaP as Bob} & \multicolumn{2}{c}{\method{}-Bob} \\
        \cmidrule(lr){3-8}\cmidrule(lr){9-10}\cmidrule(lr){11-12}\cmidrule(l){13-14}
        \textbf{Task} & \textbf{Instruction} & \textbf{Task 0} & \textbf{Random} & \textbf{TSCL} & \textbf{OMNI-EPIC} & \textbf{Traj.-only} & \textbf{Alice} & \textbf{Task 0} & \textbf{Alice} & \textbf{Task 0} & \textbf{Alice} & \textbf{Task 0} & \textbf{Alice} \\
        \midrule
        0 & Put both the alphabet soup and the tomato sauce in the basket (first task) & 11 & 11 & 11 & 11 & 10 & 11 & 6 & 5 & 10 & 11 & 12 & 12 \\
        \midrule
        1 & Put both the cream cheese box and the butter in the basket & 0 & 0 & 0 & 1 & 0 & 11 & 0 & 10 & 0 & 10 & 0 & 15 \\
        2 & Turn on the stove and put the moka pot on it & 0 & 8 & 0 & 0 & 11 & 14 & 0 & 3 & 0 & 0 & 0 & 9 \\
        3 & Put the black bowl in the bottom drawer of the cabinet and close it & 0 & 0 & 0 & 0 & 0 & 0 & 0 & 0 & 0 & 0 & 0 & 15 \\
        4 & Put the white mug on the left plate and put the yellow and white mug on the right plate & 0 & 0 & 0 & 0 & 0 & 15 & 0 & 12 & 0 & 0 & 0 & 15 \\
        5 & Pick up the book and place it in the back compartment of the caddy & 0 & 0 & 0 & 13 & 0 & 14 & 0 & 14 & 0 & 0 & 0 & 0 \\
        6 & Put the white mug on the plate and put the chocolate pudding to the right of the plate & 0 & 0 & 0 & 4 & 0 & 15 & 0 & 11 & 0 & 0 & 0 & 0 \\
        7 & Put both the alphabet soup and the cream cheese box in the basket & 0 & 0 & 0 & 12 & 0 & 8 & 0 & 14 & 0 & 14 & 0 & 13 \\
        8 & Put both moka pots on the stove & 0 & 0 & 0 & 0 & 0 & 0 & 0 & 0 & 0 & 0 & 0 & 14 \\
        9 & Put the yellow and white mug in the microwave and close it & 0 & 0 & 0 & 0 & 0 & 0 & 0 & 0 & 0 & 0 & 0 & 0 \\
        \midrule
        \multicolumn{2}{@{}l}{Tasks 1 to 9 (\%)} & 0.0 & 5.9 & 0.0 & 22.2 & 8.1 & 57.0 & 0.0 & 47.4 & 0.0 & 17.8 & 0.0 & 60.0 \\
        \bottomrule
    \end{tabularx}
    \endgroup

    \vspace{2em}
    \begin{minipage}{0.62\textwidth}
    \centering
    \caption{\textbf{Fable 5.2 as Alice and as Bob: held-out success per task.} Successes of 15 start states (25 to 39) of each task, for the policy that ASPIRE refines on task 0 and for the same policy after the cycles with the cooperative Alice (Table~\ref{tab:cooperative-fable}). The tasks are those of Tables~\ref{tab:coop-pertask-spatial} to~\ref{tab:coop-pertask-goal}.}
    \label{tab:coop-pertask-fable}
    \begingroup
    \footnotesize
    \setlength{\tabcolsep}{4pt}
    \renewcommand{\arraystretch}{1.06}
    \begin{tabular}{@{}lrrrrrr@{}}
        \toprule
        & \multicolumn{2}{c}{LIBERO-Spatial} & \multicolumn{2}{c}{LIBERO-Object} & \multicolumn{2}{c}{LIBERO-Goal} \\
        \cmidrule(lr){2-3}\cmidrule(lr){4-5}\cmidrule(lr){6-7}
        \textbf{Task} & Task 0 & Alice & Task 0 & Alice & Task 0 & Alice \\
        \midrule
        0 (first task) & 15 & 15 & 15 & 15 & 15 & 15 \\
        \midrule
        1 & 15 & 15 & 0 & 0 & 0 & 15 \\
        2 & 0 & 15 & 0 & 15 & 0 & 11 \\
        3 & 15 & 15 & 0 & 15 & 0 & 0 \\
        4 & 0 & 14 & 0 & 15 & 0 & 14 \\
        5 & 0 & 14 & 15 & 8 & 0 & 13 \\
        6 & 0 & 15 & 0 & 15 & 0 & 13 \\
        7 & 10 & 15 & 0 & 15 & 0 & 15 \\
        8 & 12 & 15 & 0 & 15 & 0 & 0 \\
        9 & 0 & 15 & 0 & 15 & 0 & 10 \\
        \midrule
        Tasks 1 to 9 (\%) & 38.5 & 98.5 & 11.1 & 83.7 & 0.0 & 67.4 \\
        \bottomrule
    \end{tabular}
    \endgroup
    \end{minipage}
\end{table*}

\clearpage
\makeatletter\global\@colht\textheight\global\@colroom\textheight\global\vsize\textheight\makeatother
\subsection{Adversarial Alice: Challenge Discovery}
\label{sec:appendix:adversarial_discovery}

\begin{table}[t]
\centering
\small
\setlength{\tabcolsep}{3pt}
\begin{tabular}{@{}lrrrr@{}}
\toprule
& \multicolumn{2}{c}{Initial policy} & \multicolumn{2}{c}{Full history} \\
\cmidrule(lr){2-3}\cmidrule(lr){4-5}
Method & \shortstack{Solved\\3/3 $\downarrow$} & \shortstack{$\geq 2$\\failures $\uparrow$} & \shortstack{Solved\\3/3 $\downarrow$} & \shortstack{$\geq 2$\\failures $\uparrow$} \\
\midrule
Uniform random & 56.3\% & 18.8\% & 90.8\% & 4.0\% \\
SFL & 80.0\% & 15.8\% & 89.9\% & 5.3\% \\
Eurekaverse & 100.0\% & 0.0\% & 78.1\% & 18.8\% \\
Trajectory-only Alice & 65.8\% & 26.4\% & 64.7\% & 25.3\% \\
Full Alice & \textbf{35.9\%} & \textbf{56.0\%} & \textbf{49.8\%} & \textbf{44.8\%} \\
\bottomrule
\end{tabular}
\caption{\textbf{Adversarial proposal quality after Bob-only training.} For challenge discovery, $\downarrow$ indicates lower is better and $\uparrow$ indicates higher is better. Percentages macro-average task-level candidate rates over 12 tasks. Classification uses each candidate's first three recorded trials, before admission filtering. Initial-policy results precede adversarial repairs; full-history results use each method's evolving policy.}
\label{tab:adversarial_discovery}
\end{table}

\minisection{Setup}
We analyze challenge discovery after Bob-only policy development on the 12-task ASPIRE ablation: three tasks each from LIBERO-Object, LIBERO-Goal, LIBERO-Spatial, and LIBERO-10 (Sec.~\ref{subsec:eval:ablation}).
Every method starts from the same task-specific Bob program.
The starting Bob policies already perform well on the original task scenes.
The goal is to find scene changes that expose their remaining weaknesses.

\minisection{Proposal methods}
Full Alice reads Bob's current program and identifies weaknesses testable through scene modifications.
The code shows what Bob observes, which measurements it reuses, and when it invokes alternative actions.
Missing observations, stale coordinates, and absent recovery branches provide hypotheses for targeted challenges.
We compare this mechanism with uniform random perturbations, SFL's success-based selection~\cite{rutherford2024noregrets}, Eurekaverse's feedback-guided environment evolution~\cite{liang2024eurekaverse}, and trajectory-only Alice (Appendix~\ref{app:baseline_proposers}).

\minisection{Measurement}
The initial-policy comparison includes every candidate screened in the first cycle, or first generation for Eurekaverse.
The full history covers eight cycles per task, or four generations with two branches for Eurekaverse.
To compare methods with different screening lengths and acceptance rules, we classify every screened candidate using its first three trials in recorded order, before admission filtering.
A candidate is \emph{already solved} if Bob succeeds in all three trials; we separately report candidates causing at least two failures.
We pool candidates within each original task, compute the task-level rates, and macro-average the 12 task rates.

\begin{samepage}
\minisection{Discovery against the initial policies}
Full Alice produces the lowest fraction of already-solved candidates, 35.9\%, and the highest fraction causing at least two failures, 56.0\% (Table~\ref{tab:adversarial_discovery}).
The latter rate is 18.8\% for uniform random, 15.8\% for SFL, 0.0\% for Eurekaverse, and 26.4\% for trajectory-only Alice.
This comparison uses identical starting programs whose original-scene checks already succeed.\par
\end{samepage}

\minisection{Discovery throughout training}
Full Alice also produces the highest fraction of candidates causing at least two failures across the full history: 44.8\%, compared with 4.0\% for uniform random, 5.3\% for SFL, 18.8\% for Eurekaverse, and 25.3\% for trajectory-only Alice.
Its already-solved fraction is the lowest: 49.8\%, versus 64.7\% for trajectory-only Alice and 78.1--90.8\% for the other methods.

\minisection{Interpretation}
These results show that access to Bob's code helps Alice propose challenges that more reliably expose weaknesses.
The qualitative examples in Appendix~\ref{sec:appendix:adversarial_qualitative} make this connection concrete: they show Alice's reasoning about Bob's policy, identify the code assumptions underlying each proposed challenge, and explain how Bob repairs the relevant behavior.

\begingroup
\raggedbottom
\clubpenalty=10000
\widowpenalty=10000
\subsection{Adversarial Alice: Qualitative Case Studies}
\label{sec:appendix:adversarial_qualitative}
We examine four case studies with ASPIRE and \method{}-Bob. Each pre-repair policy passes all five original checks but fails under Alice's challenge. Each case connects the challenge, policy limitation, physical failure, and repair. \emph{Alice's rationale} blocks are verbatim excerpts from Alice's recorded proposals; the surrounding text explains the code and execution evidence.

Scores report successes/episodes. \emph{Original checks} use five unmodified initial scenes. \emph{Retention checks} test perturbed scenes Alice selected because Bob already solves them. \emph{Earlier-challenge checks} separately retest previously admitted challenges. Each challenge's five episodes share one layout and perturbation configuration. P followed by a number denotes a program version within that run. Code excerpts omit unrelated lines and wrap statements without changing their logic. \method{}-Bob replays preserve observation perturbations and match recorded outcomes and event timing.

\subsubsection{ASPIRE: observing a moving target before contact}
\label{app:qual:book}
\begin{figure*}[t]
\centering
\begin{minipage}[t]{0.49\textwidth}\centering
\small\textbf{10\,cm challenge / before: book left behind}\\[2pt]
\includegraphics[width=\linewidth]{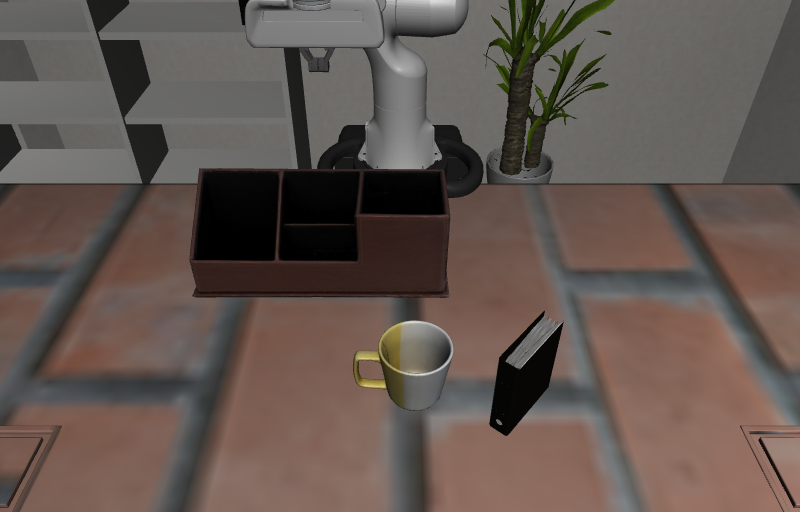}
\end{minipage}\hfill
\begin{minipage}[t]{0.49\textwidth}\centering
\small\textbf{10\,cm challenge / after: book carried}\\[2pt]
\includegraphics[width=\linewidth]{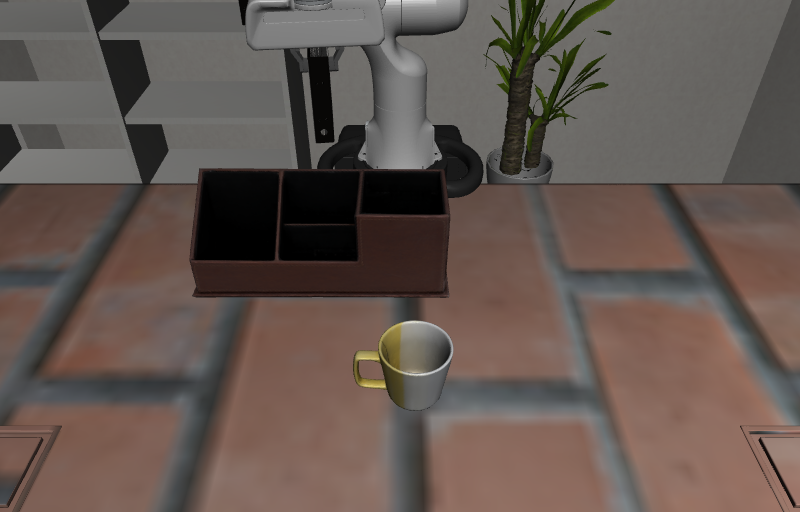}
\end{minipage}
\par\smallskip
\begin{minipage}[t]{0.49\textwidth}\centering
\small\textbf{4\,cm challenge / before: book knocked flat}\\[2pt]
\includegraphics[width=\linewidth]{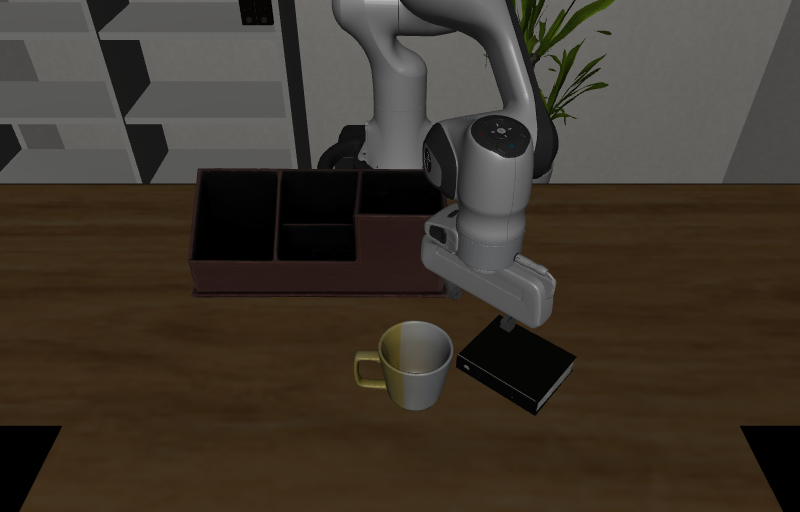}
\end{minipage}\hfill
\begin{minipage}[t]{0.49\textwidth}\centering
\small\textbf{4\,cm challenge / after: successful pickup}\\[2pt]
\includegraphics[width=\linewidth]{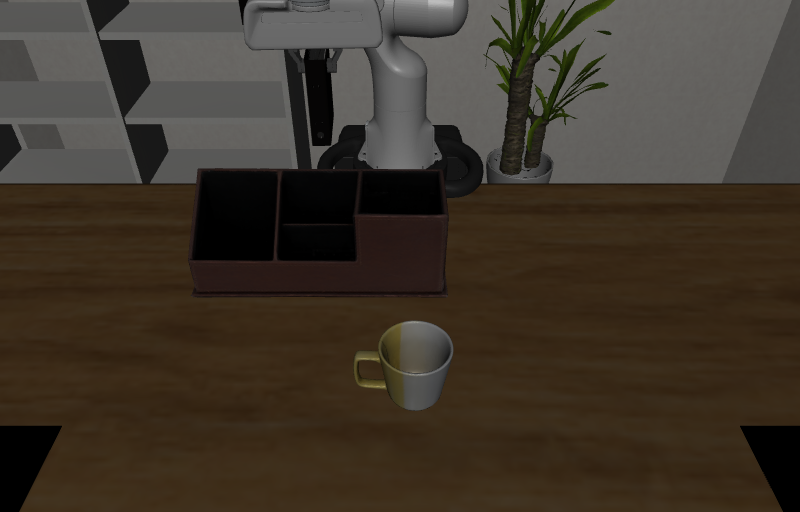}
\end{minipage}
\par\medskip
\begin{minipage}[t]{0.49\textwidth}
\small\textbf{P5: observe after descending}
\footnotesize
\begin{verbatim}
new_pts = locate_book(("black book", "book"))
if new_pts is not None:
    nbx, nby, ntop, nh, nq = book_params(new_pts)
    if (math.hypot(nbx - bx, nby - by) > 0.025
        and math.hypot(nbx - bx, nby - by) < 0.3):
        bx, by, b_top, b_h, quat = (
            nbx, nby, ntop, nh, nq)
        # ... rise and retry before closure ...
\end{verbatim}
\end{minipage}\hfill
\begin{minipage}[t]{0.49\textwidth}
\small\textbf{P10: observe during staged descent}
\footnotesize
\begin{verbatim}
for dh in (0.12, 0.07, 0.04, 0.025, 0.012):
    move_to_pose(
        Pose((bx, by, b_top + dh),
             quat, base_frame),
        tolerance=0.02, max_steps=200)
    probe = locate_book(("black book", "book"))
    if probe is None:
        continue
    pbx, pby = book_params(probe)[:2]
    sh = math.hypot(pbx - bx, pby - by)
    if 0.03 < sh < 0.3:
        # ... retreat and update the estimate ...
\end{verbatim}
\end{minipage}
\caption{\textbf{Successive refinement of observation timing with ASPIRE.} Top: P4/P5 under the 10\,cm movement. Bottom: P9/P10 under the later 4\,cm movement, on a different scene condition. Each row compares executions on the same saved challenge. P5 observes after descending but before closing the gripper. P10 also observes between shorter descent segments, allowing correction before the final approach. Images are original recorded executions.}
\label{fig:qual:book}
\end{figure*}

\minisection{Task and starting point}
In this LIBERO-10 task, Bob must pick up the upright book and place it in the back compartment of the caddy. We show P4/P5 and P9/P10. Both pre-repair policies pass 5/5 original checks; P9 includes intervening edits after P5.

\minisection{First challenge: stale target coordinates}
P4 localizes the book before approaching, then uses the cached coordinates to descend and close the gripper.

\begin{samepage}
\minisection{Alice's rationale (verbatim excerpt)}
\begin{quote}\small
Book pose is localized once at the start and never re-perceived or checked after the pre-grasp.
\end{quote}
\end{samepage}

\minisection{Challenge and physical failure}
The challenge moves the book 10\,cm in $+x$ when the gripper comes within 18\,cm. It tests the assumption that the initially measured position remains valid until grasping. The move occurs after localization, so Bob closes at the old position and proceeds toward the caddy without the book (Figure~\ref{fig:qual:book}, top).

\minisection{First repair: check after movement and before closure}
Re-localizing at hover did not help because the movement had not happened yet. P5 observes after descending but before closing. For a measured shift between 2.5 and 30\,cm, it updates position, orientation, and height, rises, and retries. The correction uses the newly measured position. Transport and placement remain unchanged.

\minisection{Second challenge: a smaller movement during descent}
The next selected challenge moves the book 4\,cm in $-y$. Alice's rationale refers to a threshold in the later policy P9's movement-handling logic.

\begin{samepage}
\minisection{Alice's rationale (verbatim excerpt)}
\begin{quote}\small
Re-localization loop only accepts shifts \textgreater{}2.5cm then treats later shifts \textless{}=4cm as occlusion noise; \ldots{}
\end{quote}
\end{samepage}

\minisection{Physical failure and second repair}
Alice's excerpt focuses on the rule that filters small shifts. Bob's repair analysis identifies a further failure: the book moves during the final descent, contacts an open finger, and tilts or falls before the post-descent check. P10 therefore descends in short segments and re-localizes between them. On detecting a shift, it rises for a clearer view, updates the estimate, and resumes the approach. Candidates that stopped checking too high still failed because movement occurred during the remaining descent.

\minisection{Measured outcome and reusable decision}
Each selected repair changes its challenge from 0/5 to 5/5, while original checks remain 5/5 and retention checks remain 9/9. The two cycles show a progression from cached localization, to checking before gripper closure, to checking throughout descent. The reusable decision is to verify the target while there is still time to react before contact changes the situation.

\subsubsection{ASPIRE: checking whether a detection fits the target}
\label{app:qual:pudding}
\begin{figure*}[t]
\centering
\begin{minipage}[t]{0.49\textwidth}\centering
\small\textbf{Occlusion / before: the wrong blue can}\\[2pt]
\includegraphics[width=\linewidth]{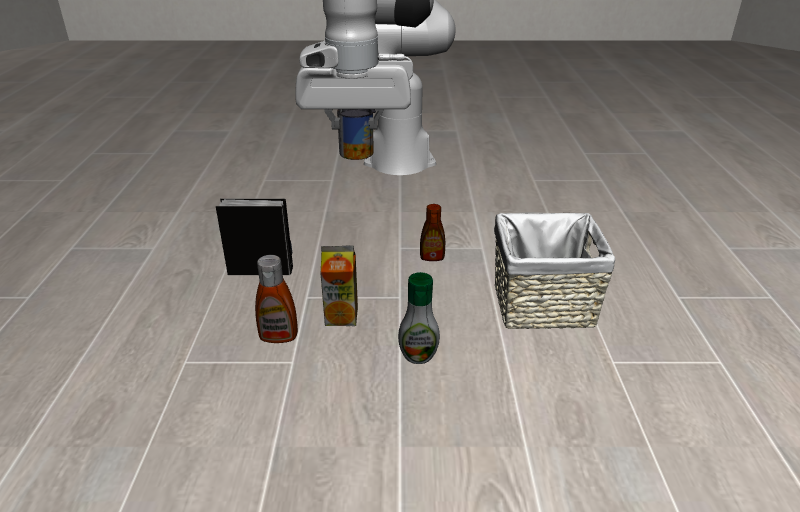}
\end{minipage}\hfill
\begin{minipage}[t]{0.49\textwidth}\centering
\small\textbf{Occlusion / after: pudding recovered}\\[2pt]
\includegraphics[width=\linewidth]{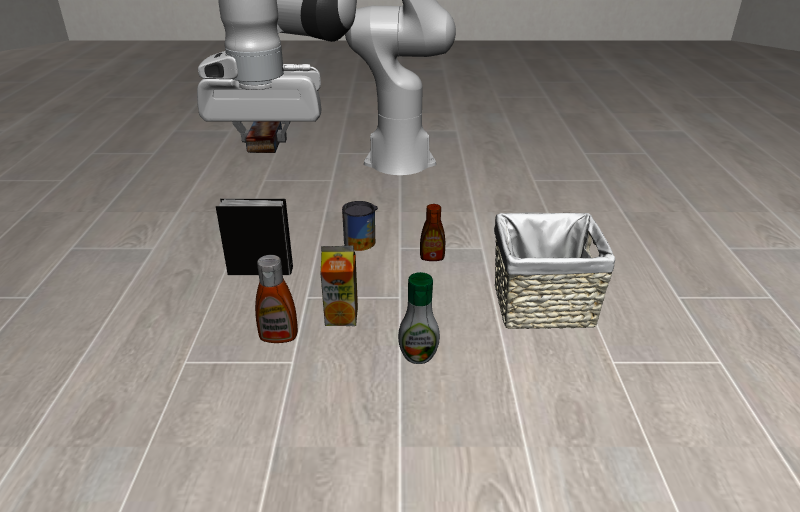}
\end{minipage}
\par\smallskip
\begin{minipage}[t]{0.49\textwidth}\centering
\small\textbf{Camera change / before: the wrong blue can}\\[2pt]
\includegraphics[width=\linewidth]{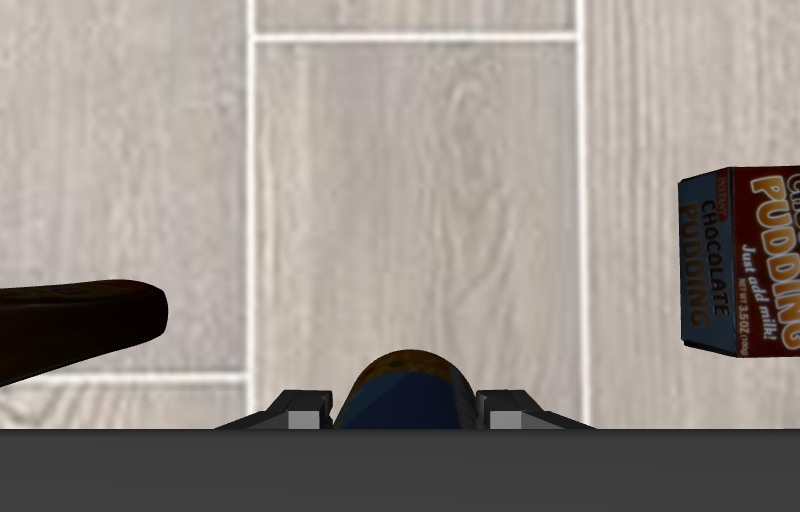}
\end{minipage}\hfill
\begin{minipage}[t]{0.49\textwidth}\centering
\small\textbf{Camera change / same repair: pudding}\\[2pt]
\includegraphics[width=\linewidth]{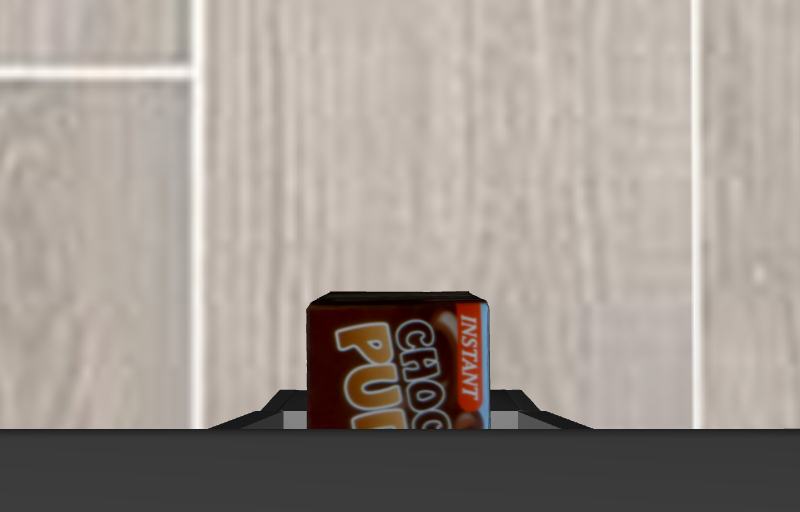}
\end{minipage}
\par\medskip
\begin{minipage}[t]{0.49\textwidth}
\small\textbf{P6: obtain another view when needed}
\footnotesize
\begin{verbatim}
if top_z > 0.06:
    flat = survey_flat_target()
    report["survey"] = flat
    if flat is not None:
        cx, cy, top_z = flat
\end{verbatim}
\end{minipage}\hfill
\begin{minipage}[t]{0.49\textwidth}
\small\textbf{Success with the same P5/P6 pair}\par\smallskip
\begin{tabular}{@{}lrr@{}}
\toprule
Condition & Before & After \\
\midrule
Occlusion challenge & 0/5 & 5/5 \\
Original checks & 5/5 & 5/5 \\
Camera-scale change & 0/3 & 3/3 \\
\bottomrule
\end{tabular}
\end{minipage}
\caption{\textbf{One observation rule addresses two sensing failures with ASPIRE.} Top: the occlusion challenge used for repair. Bottom: a separately screened camera-scale perturbation without the occluding book, shown through the wrist camera. Both rows use the identical P5/P6 pair; each illustrated before/after pair has matching reset records. Images are original recorded executions. \texttt{survey\_flat\_target} obtains an overhead wrist view and searches for a small, low candidate.}
\label{fig:qual:pudding}
\end{figure*}

\minisection{Task and starting point}
`In this LIBERO-Object task, Bob must pick up the chocolate pudding and place it in the basket. The incumbent P5 succeeds on 5/5 original checks. Its initial \texttt{find\_small} search relies on the front camera, although wrist-camera sensing is already available later in the program.

\begin{samepage}
\minisection{Alice's rationale (verbatim excerpt)}
\begin{quote}\small
find\_small relies on a single agentview mask of the pudding with no wrist-camera fallback; \ldots{}
\end{quote}
\end{samepage}

\minisection{Challenge and policy limitation}
The challenge places a book in front of the pudding, blocking the front-camera view while leaving the pudding visible from a wrist-camera view. It tests whether Bob obtains another observation when its initial detection is unreliable.

\minisection{Physical failure}
The initial search accepts a taller blue can as the target. Bob grasps and carries the can, so the program can finish normally while the task fails (Figure~\ref{fig:qual:pudding}, top).

\minisection{Bob's repair}
P6 checks \texttt{top\_z}, the estimated top-surface z-coordinate in the robot base frame. In these scenes, the low pudding box gives an estimate of about 3\,cm, whereas the wrong detection gives about 8\,cm. If the estimate exceeds 6\,cm, Bob moves to a high hover and searches with the wrist camera. It selects a small, low candidate consistent with the pudding, updates the target coordinates, and reuses the existing grasp-and-placement routine. This task-specific check detects a physical inconsistency and triggers another view.

\minisection{Outcome and reuse on another perturbation}
Occlusion success improves from 0/5 to 5/5, while original checks remain 5/5. The same P5/P6 pair is evaluated on a separately screened camera-scale change (\texttt{scale\_factor=1.9}), without the occluding book. Success changes from 0/3 to 3/3. The initial detection again has an implausible top height; the repaired program invokes the wrist survey and finds the pudding, replacing a 7.94\,cm estimate with 2.80\,cm in the illustrated run. The same observation rule addresses both perturbations: obtain another view when a detection conflicts with the target's physical properties.

\subsubsection{\method{}-Bob: reaching a moved plate}
\label{app:qual:reach}
\begin{figure*}[t]
\centering
\begin{minipage}[t]{0.49\textwidth}\centering
\small\textbf{Before: grasp succeeds; transport fails}\\[2pt]
\includegraphics[width=\linewidth]{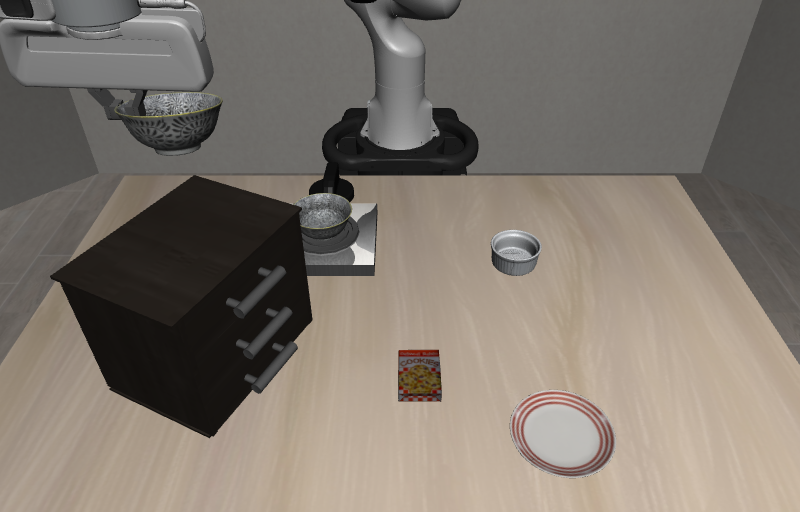}
\end{minipage}\hfill
\begin{minipage}[t]{0.49\textwidth}\centering
\small\textbf{After: bowl placed on the moved plate}\\[2pt]
\includegraphics[width=\linewidth]{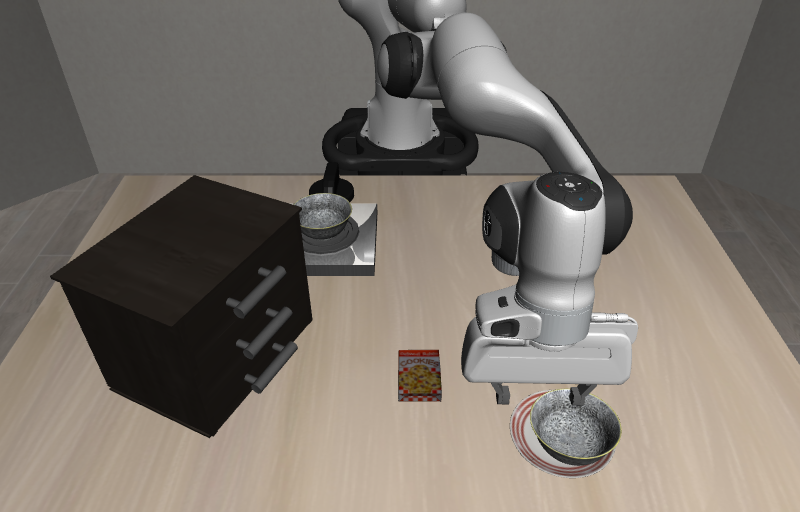}
\end{minipage}
\par\medskip
\begin{minipage}[t]{0.49\textwidth}
\small\textbf{Before (P9): retry only the height}
\footnotesize
\begin{verbatim}
for dz in (0.0, -0.04, -0.08):
    z = ctx["lift_z"] + dz
    if mv(ctx, tx, ty, z, 0.02, 260):
        # ...
\end{verbatim}
\end{minipage}\hfill
\begin{minipage}[t]{0.49\textwidth}
\small\textbf{After (P11): search pitch and height}
\footnotesize
\begin{verbatim}
for deg in (0.0, 15.0, 25.0, 35.0, 45.0):
    signs = ((1.0,) if deg == 0.0
             else (-1.0, 1.0))
    for sg in signs:
        q = (base_q if deg == 0.0
             else pitched_quat(base_q, sg * deg))
        for dz in (0.0, -0.04, -0.08):
            # ...
            if mv(ctx, tx, ty, z0 + dz,
                  0.02, 260, q):
                # ...
\end{verbatim}
\end{minipage}
\caption{\textbf{Reaching a relocated destination with \method{}-Bob.} The before policy correctly finds the moved plate, but the planner rejects transport poses at all three carrying heights with the fixed wrist orientation. The repair searches alternative orientations. Images are archived-action replays of the compared P9/P11 executions; the right image shows completed placement. In the code, \texttt{q} is the selected wrist orientation.}
\label{fig:qual:reach}
\end{figure*}

\minisection{Task and starting point}
In this LIBERO-Spatial task, Bob must pick up the black bowl on the wooden cabinet and place it on the plate. The incumbent P9 succeeds on 5/5 original checks. It already re-observes the plate during transport, but its fallback changes carrying height while keeping the wrist orientation fixed.

\begin{samepage}
\minisection{Alice's rationale (verbatim excerpt)}
\begin{quote}\small
Transport re-observes the plate, but a 10 cm +x move puts the target at the edge of reach; \ldots{}
\end{quote}
\end{samepage}

\minisection{Challenge and policy limitation}
The selected challenge moves the plate 10\,cm in $+x$ when the gripper approaches the bowl. It tests whether Bob can adapt its carrying pose when the destination becomes difficult to reach. Bob's existing fallback varies height while keeping wrist orientation fixed. Figure~\ref{fig:rallybob-adversarial-example} gives the corresponding code and execution records.

\minisection{Physical failure}
Bob grasps the bowl and correctly locates the moved plate. At each of three carrying heights, the planner finds no valid arm configuration with the current wrist orientation and reports inverse-kinematics (IK) failure. These commands execute no motion; Bob stops while holding the bowl (Figure~\ref{fig:qual:reach}). Matched checks succeed 3/3 without movement and 0/3 with it. Target localization is correct, but the carrying pose fails.

\minisection{Bob's repair}
P11 searches wrist pitch as well as carrying height and uses the first feasible pose. In the illustrated execution, a $-15^\circ$ tilt permits transport and placement. The grasping routine is unchanged.

\minisection{Measured outcome}
The challenge improves from 0/5 to 5/5, while original checks remain 5/5 and retention checks remain 9/9. The reusable decision is to change the carrying pose when the planner cannot reach a correctly localized destination with the current orientation.

\minisection{Reuse under other evaluation conditions}
The frozen final program P12 is evaluated on two additional variants of the same LIBERO-Spatial task: pick up the black bowl on the wooden cabinet and place it on the plate. P12 retains P11's carrying-search routine unchanged and also includes a later lifting repair. Neither evaluation case is used in any cycle:
\begin{itemize}
\item \textbf{Changed robot initial state (LIBERO-Plus).} The robot starts from a different configuration, while the plate remains stationary. The planner rejects three carrying heights with the original wrist orientation. Tilting the wrist by $-15^\circ$ produces a feasible transport pose, and Bob completes placement.
\item \textbf{Changed camera setting and diagonal plate movement (LIBERO-MAX).} The camera-scale parameter is 1.97, and the plate moves 10\,cm diagonally in the positive x and y directions. The planner again rejects three carrying heights with the original orientation. A $-15^\circ$ wrist tilt enables transport, and Bob completes placement.
\end{itemize}
Each case is one scored episode. These executions demonstrate reuse of the same transport routine under different conditions within this task.

\subsubsection{\method{}-Bob: accepting an updated basket observation}
\label{app:qual:basket}
\begin{figure*}[t]
\centering
\begin{minipage}[t]{0.44\textwidth}\centering
\small\textbf{Before: initial wrist view is clipped}\\[2pt]
\includegraphics[width=\linewidth]{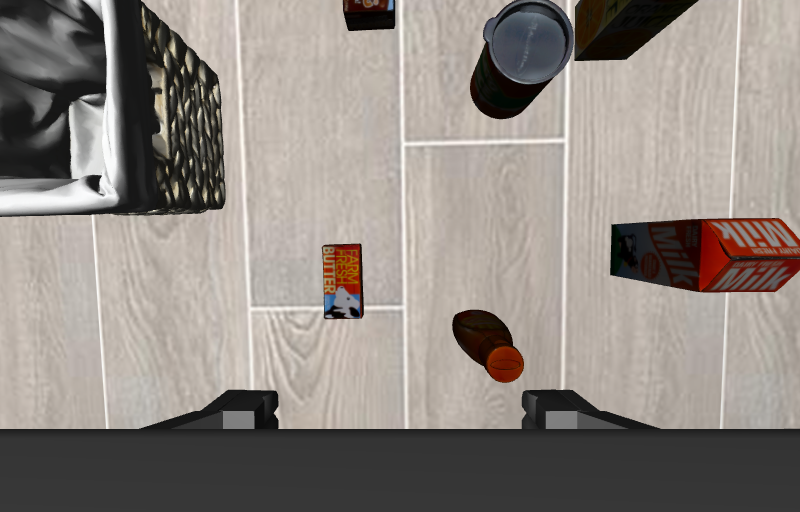}
\end{minipage}\hfill
\begin{minipage}[t]{0.44\textwidth}\centering
\small\textbf{After: fuller view of the moved basket}\\[2pt]
\includegraphics[width=\linewidth]{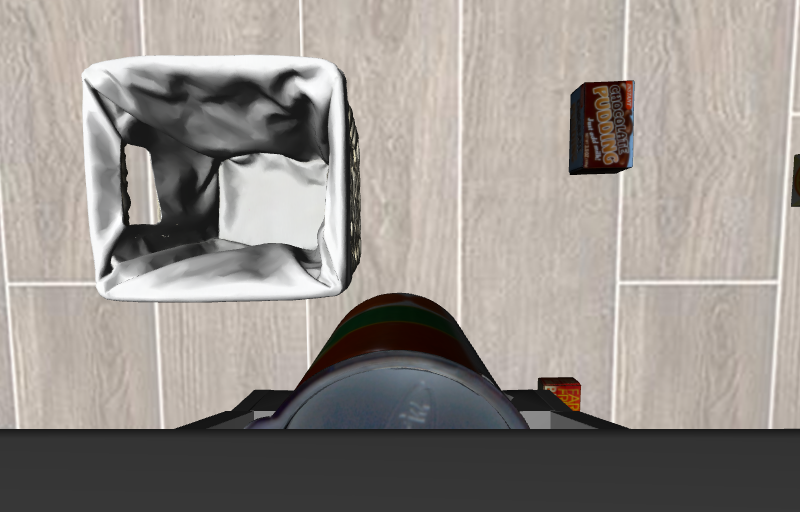}
\end{minipage}
\par\smallskip
\begin{minipage}[t]{0.44\textwidth}\centering
\small\textbf{Before: stale coordinates miss the basket}\\[2pt]
\includegraphics[width=\linewidth]{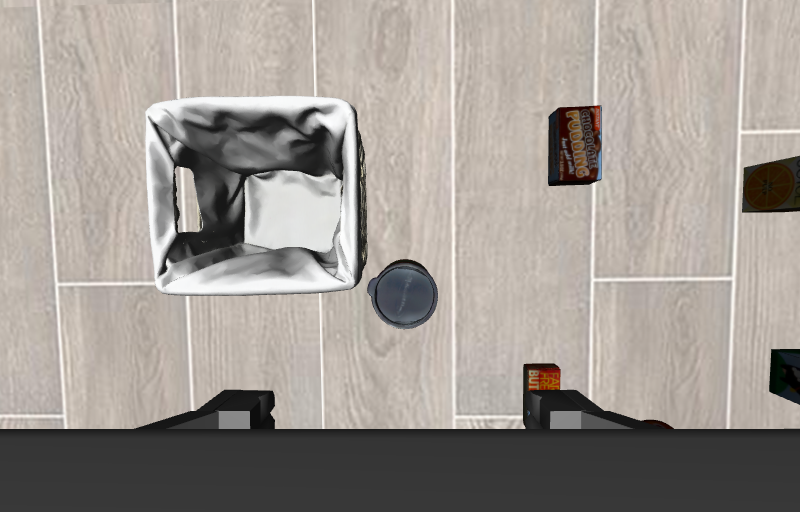}
\end{minipage}\hfill
\begin{minipage}[t]{0.44\textwidth}\centering
\small\textbf{After: corrected placement inside the basket}\\[2pt]
\includegraphics[width=\linewidth]{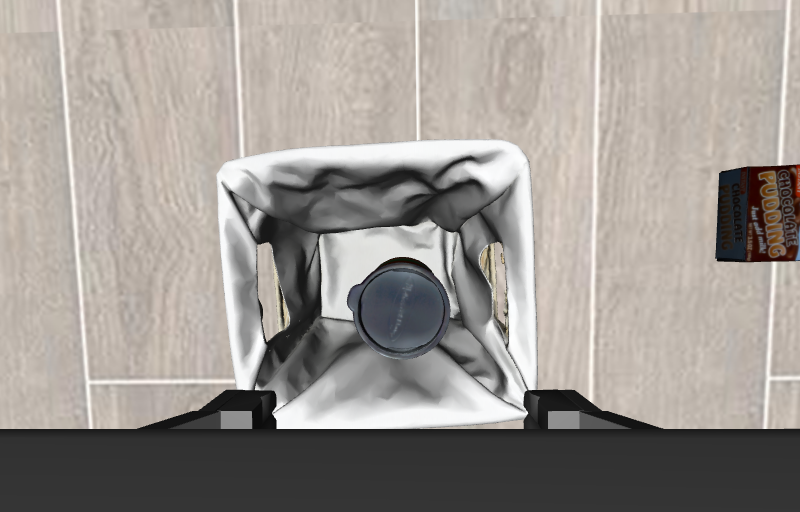}
\end{minipage}
\par\medskip
\begin{minipage}[t]{0.44\textwidth}
\small\textbf{Before (P8): reject increases above 25\%}
\footnotesize
\begin{verbatim}
if (w_new < 0.8 * w_old
        or w_new > 1.25 * w_old):
    # ... logging ...
    return False
\end{verbatim}
\end{minipage}\hfill
\begin{minipage}[t]{0.44\textwidth}
\small\textbf{After (P10): allow a fuller view}
\footnotesize
\begin{verbatim}
if w_new < 0.8 * w_old:
    # ... logging ...
    return False
if w_new > 1.8 * w_old:
    # ... logging ...
    return False
\end{verbatim}
\end{minipage}
\caption{\textbf{Updating the basket observation with \method{}-Bob.} Top-left: the clipped initial wrist view in the pre-repair P8 execution. Top-right: a fuller view of the moved basket in the repaired P10 execution. Bottom: P8 releases outside the basket; P10 accepts the fuller observation and uses its updated position to recenter before release. The code shows the revised size check; \texttt{w\_old} and \texttt{w\_new} are the stored and refreshed estimates of basket extent. Images are archived-action replays.}
\label{fig:qual:basket}
\end{figure*}

\minisection{Task and starting point}
In this LIBERO-Object task, Bob must pick up the tomato-sauce can and place it in the basket. The incumbent P8 succeeds on 5/5 original checks. It already has a basket-refresh routine and can correct its release position, but it rejects observations whose estimated basket size differs too much from the initial estimate.

\begin{samepage}
\minisection{Alice's rationale (verbatim excerpt)}
\begin{quote}\small
Combining two individually solved perturbations removes the fallbacks that each repair relies on: the blur forces the wrist camera for basket localization, and the basket move makes the stored pose stale.
\end{quote}
\end{samepage}

\minisection{Challenge and policy limitation}
The selected challenge combines strong front-camera blur with a 10\,cm diagonal basket move. Blur limits front-camera localization, while movement requires a fresh destination estimate. The combination tests whether Bob can accept and act on an updated wrist-camera observation. With blur held constant, matched checks succeed 3/3 without movement and 0/3 with it.

\minisection{Physical failure}
Bob checks a new basket observation against the size estimated from its first observation. The first wrist view shows only part of the basket, underestimating its extent as 0.234\,m. After the basket moves, a fuller view gives an extent of 0.314\,m and an updated position. The basket has not grown: more of it is visible. Because the apparent size increase is about 34\%, exceeding the program's 25\% limit, Bob rejects the update and releases the can at the old position (Figure~\ref{fig:qual:basket}). The failure comes from treating an incomplete initial observation as a reliable size reference.

\minisection{Bob's repair}
P10 raises the allowed extent increase from 25\% to 80\%, while retaining lower-size and rim-height checks. It also allows larger position updates from confident observations and increases the refresh-and-recenter limit from one attempt to two. In the illustrated execution, Bob accepts the updated observation, corrects its placement position by 15.2\,cm, and releases inside the basket.

\minisection{Measured outcome and reusable decision}
Challenge success improves from 0/5 to 5/5. Original checks remain 5/5, retention checks remain 9/9, and earlier-challenge checks remain 5/5. The reusable decision is to account for incomplete earlier views when checking a new observation, then update the action from the accepted estimate.

\endgroup

\subsection{Cooperative Alice: Learning Progress}
\label{sec:appendix:cooperative_learning_progress}

\minisection{Setup}
We evaluate whether different curricula produce useful improvements to Bob.
All comparisons use ASPIRE as Bob across LIBERO-Spatial, LIBERO-Object, LIBERO-Goal, and LIBERO-10.
Each method receives eight curriculum cycles in Spatial and Object and sixteen in Goal and LIBERO-10.
The curricula are described in Appendix~\ref{app:baseline_proposers}.

\minisection{Measurement}
For each cycle with an admitted practice task, we evaluate Bob before and after the cycle on the same five fresh layouts, separate from those supplied for repair.
The cycle's learning gain is the difference between these two success rates.
For example, improving from 1/5 to 4/5 gives a gain of 60 percentage points.
The after score uses the program retained at the end of the cycle.
\emph{Learning gain per cycle} averages these signed gains over all scheduled cycles, retaining negative changes.
\emph{Cycles with fresh-layout improvement} reports the fraction of scheduled cycles whose gain is strictly positive.
A cycle with no admitted task contributes zero realized learning gain and remains in both denominators.
Neither statistic filters cycles by its acceptance label.
We compute both statistics within each suite and average the four suite-level values equally.

\minisection{Results}
Full Alice achieves an average learning gain of 85.0\% per cycle and produces positive fresh-layout improvement in 95.3\% of cycles (Table~\ref{tab:cooperative_learning_progress}).
It exceeds Random, TSCL, OMNI-EPIC and Trajectory-only Alice on both measures: its curriculum produces larger and more frequent improvements on fresh layouts of the selected practice tasks.

\minisection{Interpretation}
These results are consistent with using Bob's policy to guide cooperative task selection.
Inspecting the program exposes which behaviors Bob can already reuse and which conditions or responses are missing, helping Alice identify useful extensions of its current capabilities.
The cooperative policy example in Figure~\ref{fig:rallybob-example} shows how accepted refinements add explicit branches for selecting the target bowl while reusing the actions for transporting and placing it.

\begin{table}[ht]
    \centering
    \caption{\textbf{Learning progress during cooperative training with ASPIRE as Bob.} Learning gain measures the before/after increase in success rate on fresh layouts of the selected practice task. The improvement rate is the fraction of scheduled cycles with a strictly positive gain. Both metrics are computed within each suite and averaged equally across the four suites. Higher is better.}
    \label{tab:cooperative_learning_progress}
    \begingroup
    \footnotesize
    \setlength{\tabcolsep}{3pt}
    \renewcommand{\arraystretch}{1.08}
    \begin{tabularx}{\columnwidth}{@{}>{\raggedright\arraybackslash}p{0.40\columnwidth}*{2}{>{\centering\arraybackslash}X}@{}}
        \toprule
        \textbf{Curriculum}
        & \shortstack{Learning gain\\per cycle\\(\%) $\uparrow$}
        & \shortstack{Cycles with\\fresh-layout\\improvement\\(\%) $\uparrow$} \\
        \midrule
        Random & 54.1 & 65.6 \\
        TSCL~\cite{matiisen2019teacher} & 56.3 & 75.0 \\
        OMNI-EPIC~\cite{faldor2025omniepic} & 42.2 & 51.6 \\
        Trajectory-only Alice & 60.2 & 75.6 \\
        \textbf{Full Alice} & \textbf{85.0} & \textbf{95.3} \\
        \bottomrule
    \end{tabularx}
    \endgroup
\end{table}

\begingroup
\raggedbottom
\clubpenalty=10000
\widowpenalty=10000
\newcommand{\coopframe}[2]{\begin{minipage}[t]{0.245\textwidth}\centering\includegraphics[width=\linewidth]{appendix/cooperative_qualitative/images/#1.png}\\[-1pt]{\scriptsize #2}\end{minipage}}
\newcommand{\coopframethree}[2]{\begin{minipage}[t]{0.325\textwidth}\centering\includegraphics[width=\linewidth]{appendix/cooperative_qualitative/images/#1.png}\\[-1pt]{\scriptsize #2}\end{minipage}}

\subsection{Cooperative Alice: Qualitative Case Studies}
\label{sec:appendix:cooperative_qualitative}
We examine how Alice uses Bob's current program to choose useful practice tasks. Alice identifies a missing capability and selects a task that builds on behaviors Bob already handles. Bob then refines its program on that task. Each case shows the relevant code, Alice's rationale, the practice task, and the resulting changes in robot behavior. We report whether the revision improves the practiced task, preserves earlier capabilities, and supports additional tasks. The Object and Spatial examples use ASPIRE as Bob's harness; the Goal and LIBERO-10 examples use \method{}-Bob.

A score of $x/y$ means $x$ successful executions out of $y$ attempts. \emph{Practice} results use the five layouts provided for Bob's repair. \emph{Fresh-layout} results use five additional layouts of the same task. \emph{Unpracticed} tasks are different tasks that were never selected for Bob's repair.

Quotations are verbatim excerpts from Alice's recorded proposals. Code excerpts show the relevant parts of Bob's program, with unrelated lines omitted and long statements wrapped for readability. Images show frames from recorded robot executions.

\subsubsection{Object / ASPIRE: a bowl rim grasp carries to other containers}
\label{app:coopqual:rim}

\begin{figure*}[t]
\centering
\small\textbf{Practice: black bowl to basket / before, P5}\\[2pt]
\coopframe{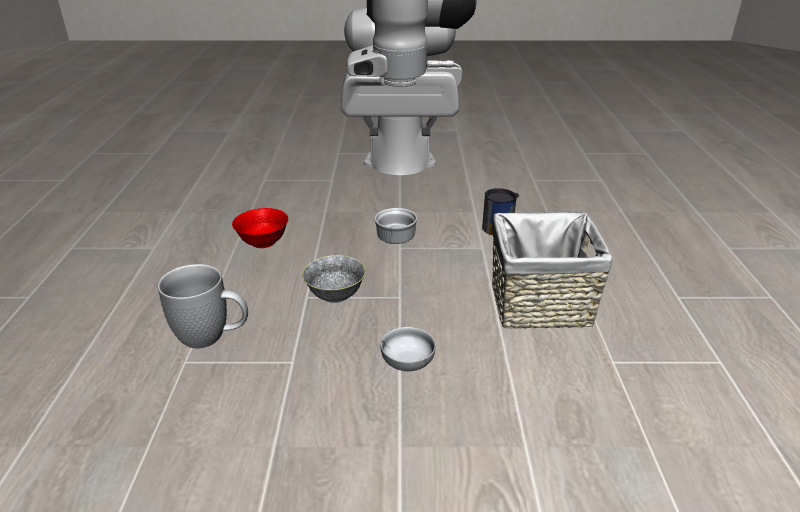}{Start}\hfill
\coopframe{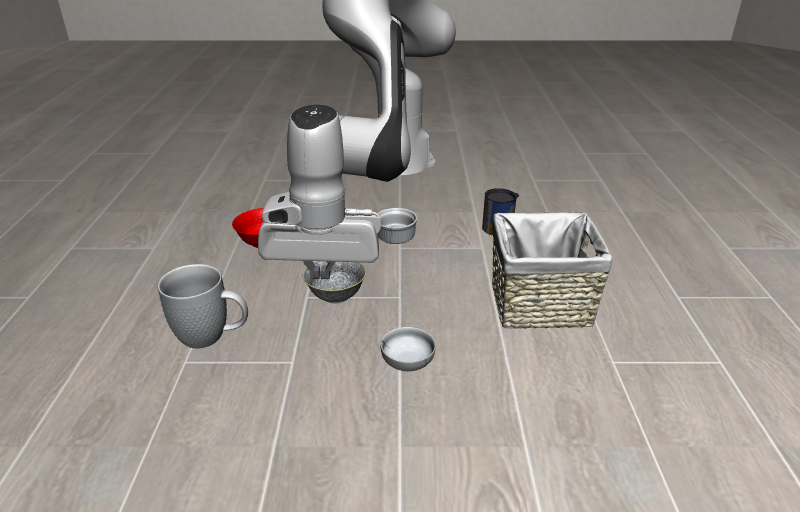}{Gripper closes}\hfill
\coopframe{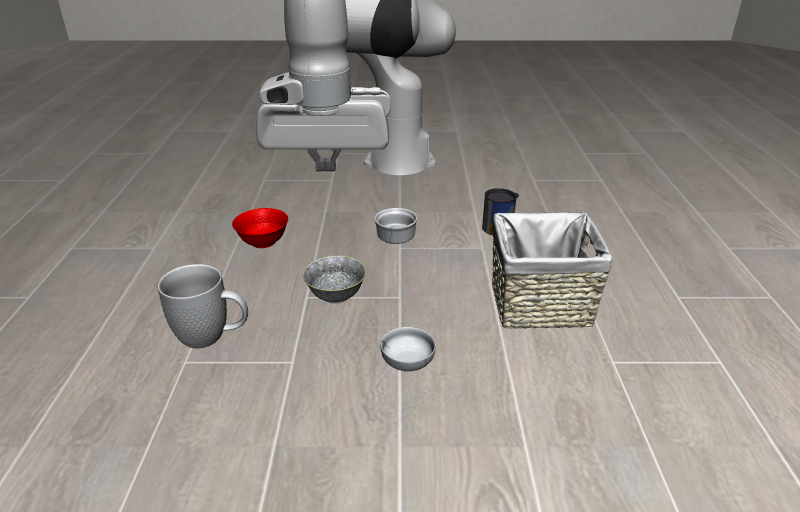}{Empty lift}\hfill
\coopframe{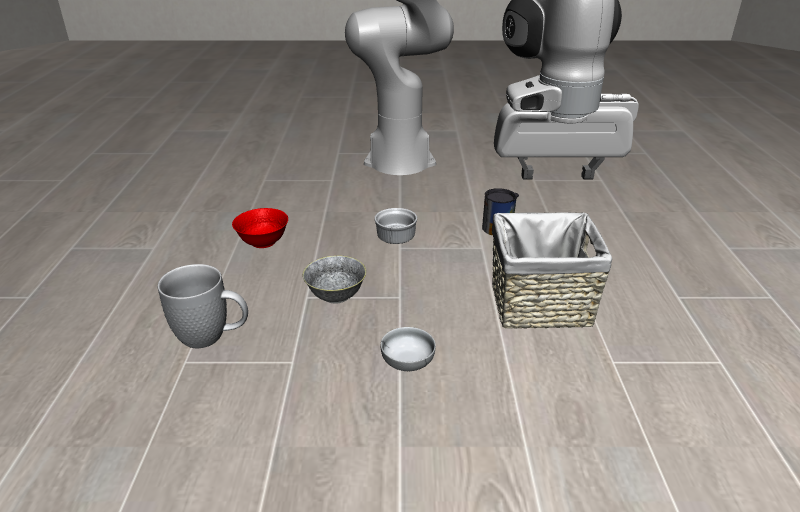}{Bowl left behind}
\par\smallskip
\textbf{Practice: black bowl to basket / after, P6}\\[2pt]
\coopframe{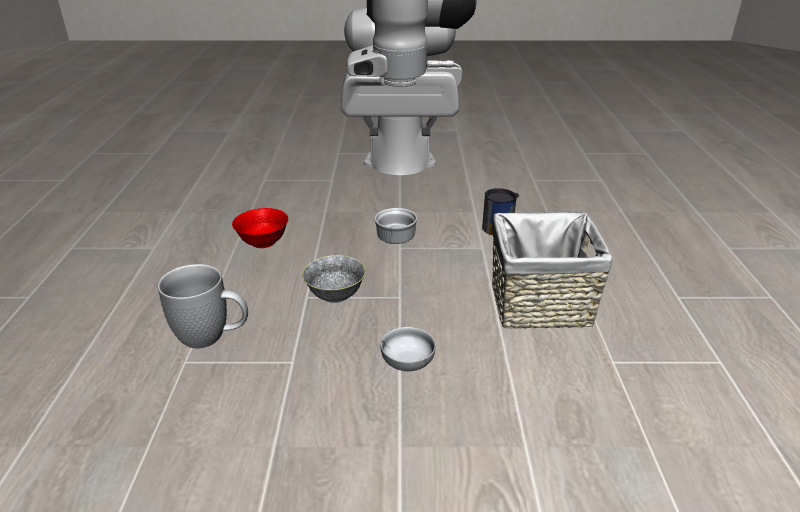}{Start}\hfill
\coopframe{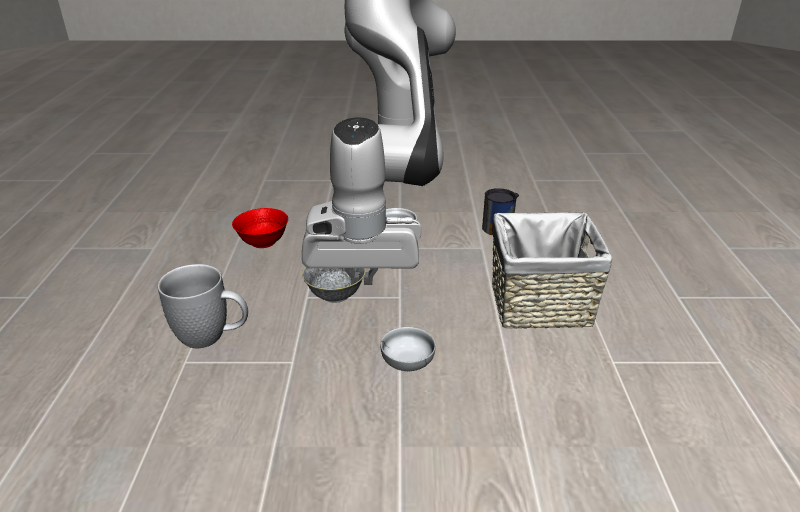}{Rim grasp}\hfill
\coopframe{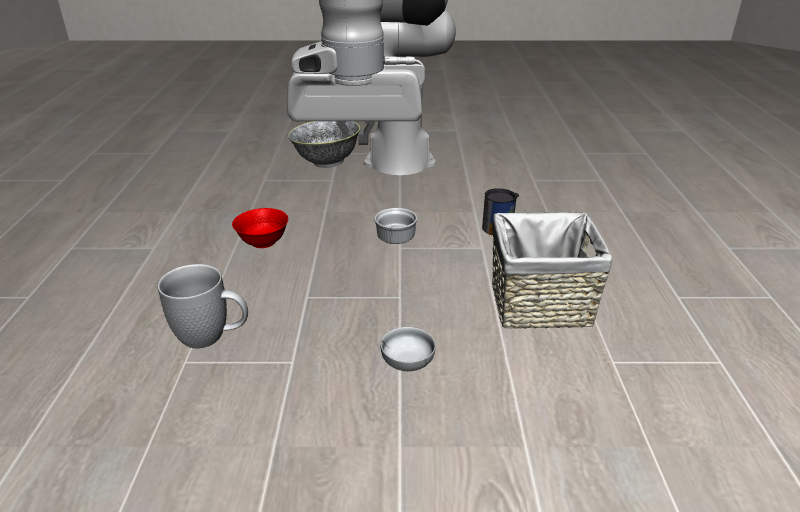}{Lift}\hfill
\coopframe{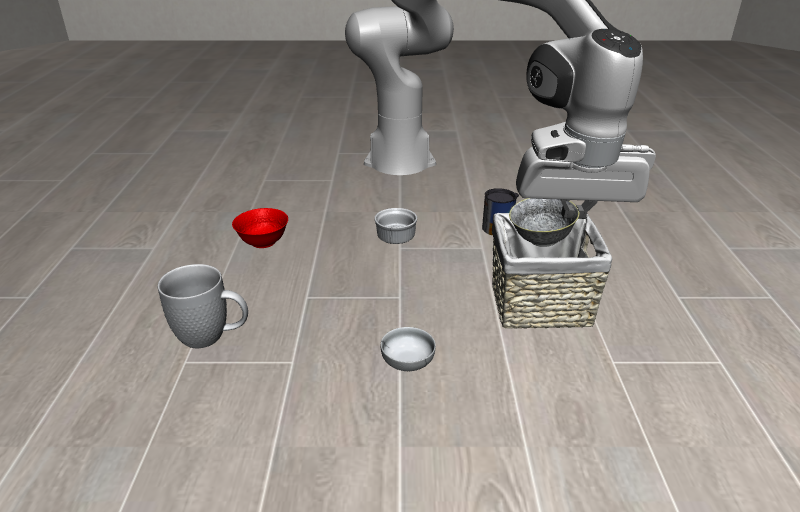}{Bowl in basket}
\par\smallskip
\textbf{Unpracticed: red coffee mug to basket / before, P5}\\[2pt]
\coopframe{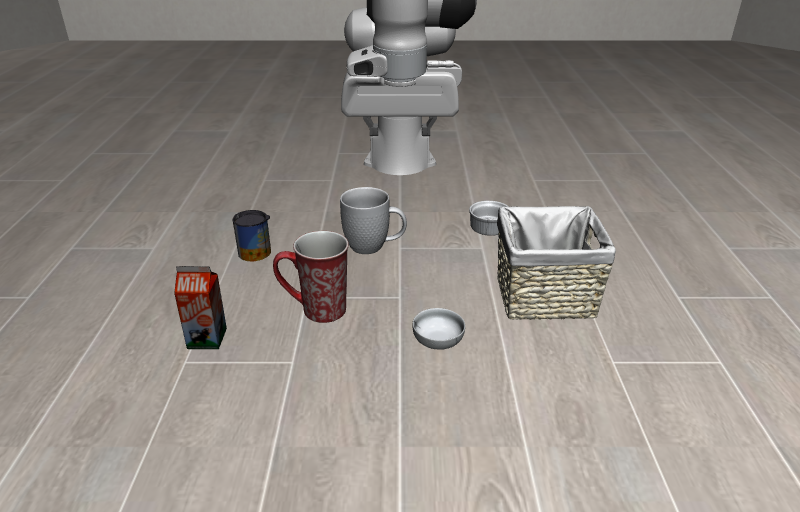}{Start}\hfill
\coopframe{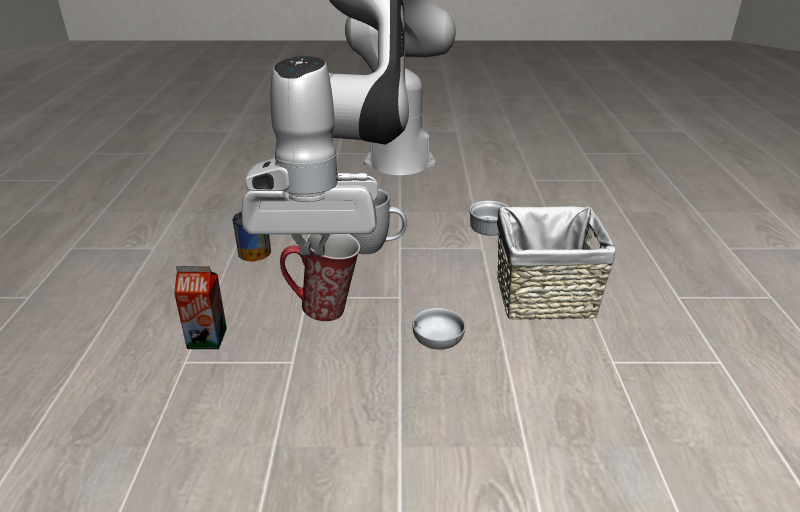}{Closes on nothing}\hfill
\coopframe{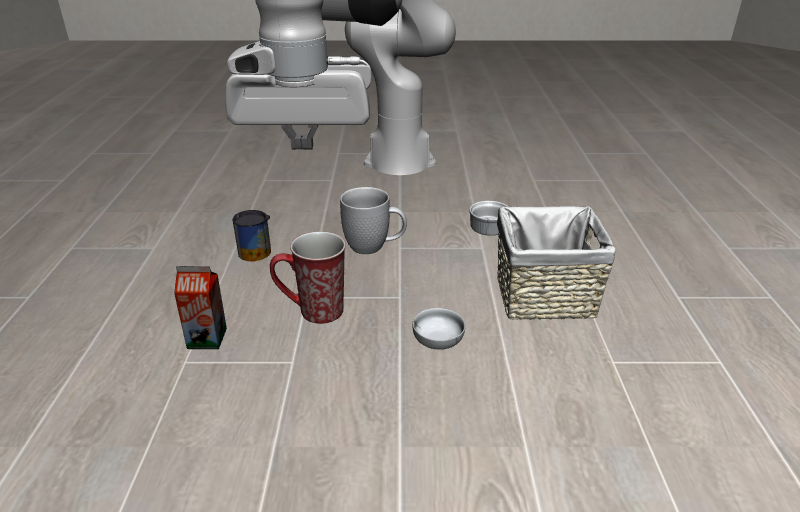}{Empty lift}\hfill
\coopframe{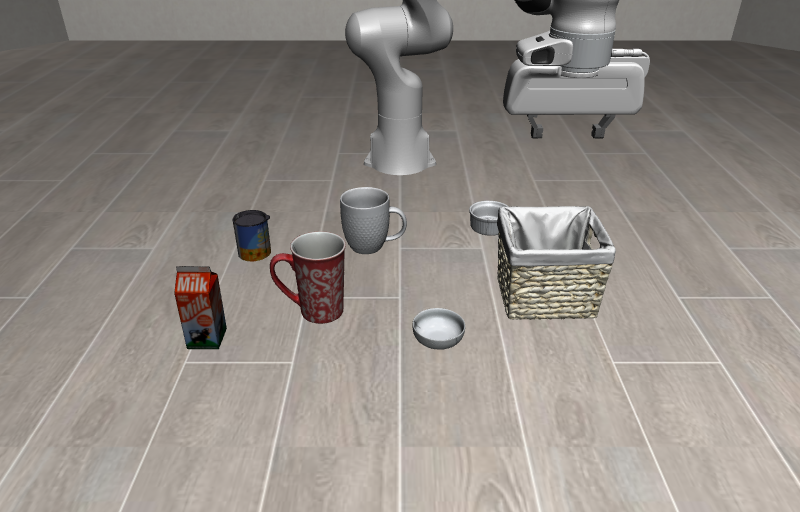}{Mug left behind}
\par\smallskip
\textbf{Unpracticed: red coffee mug to basket / same repair, P6}\\[2pt]
\coopframe{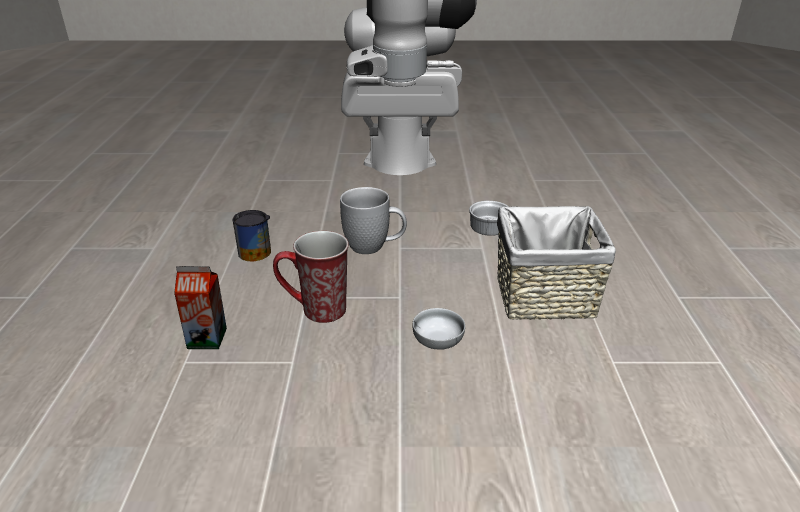}{Start}\hfill
\coopframe{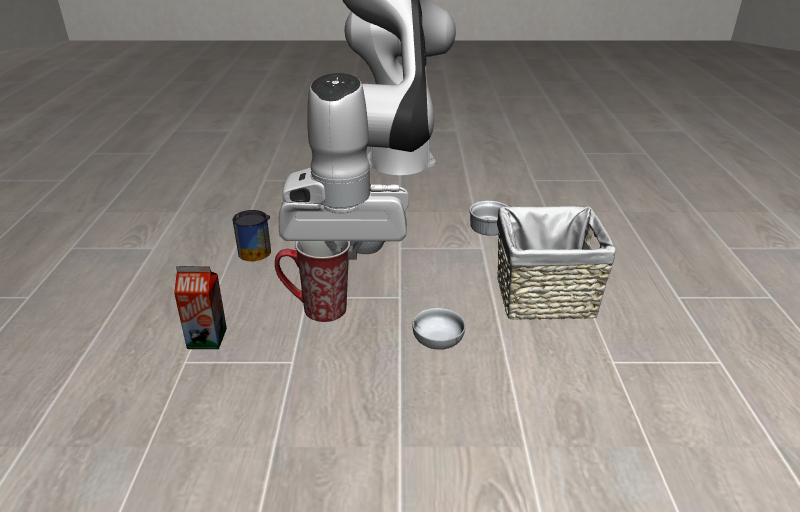}{Rim grasp}\hfill
\coopframe{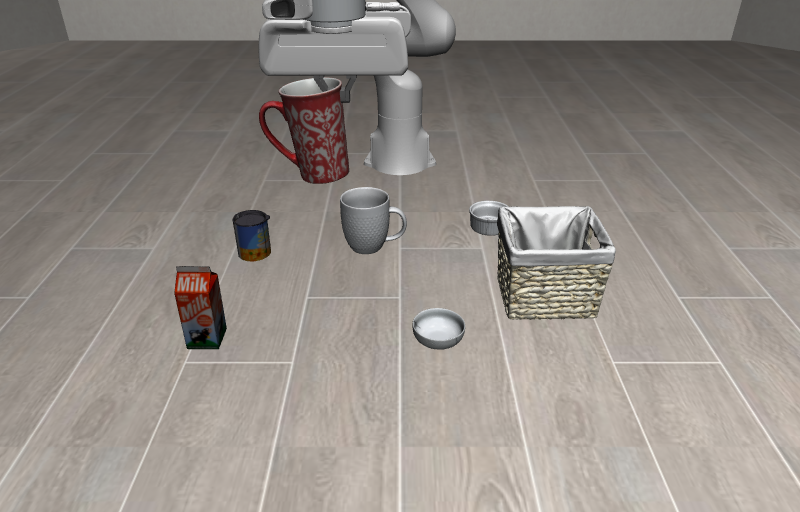}{Lift}\hfill
\coopframe{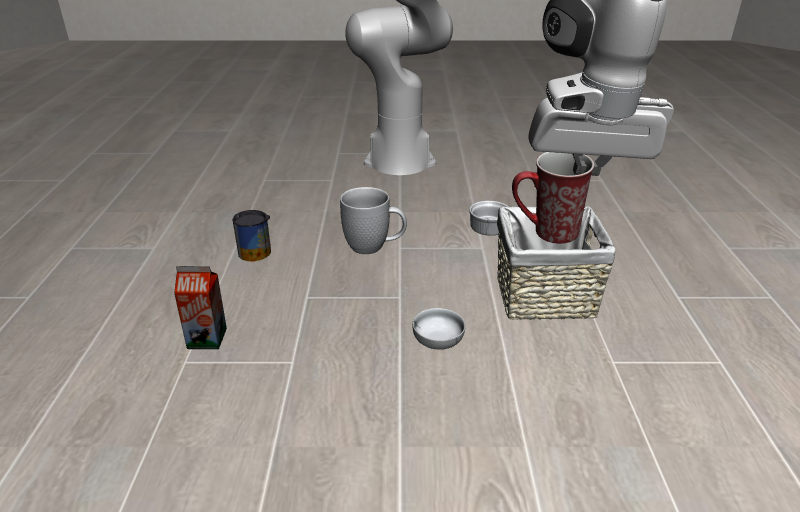}{Mug in basket}
\caption{\textbf{A rim grasp learned on a bowl also handles a mug with ASPIRE as the harness.} Each before/after pair shows the same task layout. The mug task changes the object class and distractors; Alice screened it but never selected it for repair. P6 uses its new \texttt{rim\_hold} branch in all three mug episodes.}
\label{fig:coopqual:rim}
\end{figure*}

\minisection{Current capability}
In cycle 6 of LIBERO-Object, it solves each of five earlier practice tasks in 5/5 layouts. An unknown item name falls through to a generic detector, then receives the default grasp at the centre of its top face:

\begingroup\footnotesize
\begin{verbatim}
# P5: unknown-name fallback (lines 111--123)
return item, {"prompts": (item,),
              "size": ANY_SIZE, "colour": None}
# Default grasp (lines 417--427)
grasp_z = max(target["top_z"] - 0.035,
              target["floor_z"] + 0.02)
...
go("pregrasp", target["cx"], target["cy"],
   target["top_z"] + 0.09)
go("grasp", target["cx"], target["cy"], grasp_z)
close_gripper()
\end{verbatim}
\endgroup
This can grasp a solid item across its body. A wide bowl instead puts the gripper over a hollow centre, with its rim wider than the finger opening.

\minisection{Alice's reasoning}
Alice screens several open containers, inspects failures, probes perception, and reads the fallback and grasp code. Alice's submitted hypothesis identifies the common physical assumption:
\begin{quote}\small
The program has no hold for an open, wide item. An unknown name goes through the name-query fallback (lines 111-123, ANY\_SIZE, best score) and then the default can-style hold (line 417, used at lines 426-427): the hand goes down on the centre of the top face. For a bowl that centre is the hollow; the bowl is 0.09-0.15 m wide, too wide to hold across. The descent times out and the gripper closes on nothing. Release (lines 418-420) also assumes the hand is over the item's centre.
\end{quote}
Alice's rationale distinguishes a grasp failure from the additional selection failures of other containers:
\begin{quote}\small
Perception already works here: 'black bowl' scores 0.70 on the target against 0.30 and 0.22 on the next two masks, and the basket is found. All 3 layouts reach the grasp step and fail only there (move\_to\_pose timeout, no item moved). Other hollow items are farther: white bowl, red ramekin and white porcelain mug (S13-S15) are not found by name and the fallback carried the milk or tomato sauce. So black bowl isolates the one missing part, the rim hold. That hold should carry to red and yellow bowl, ramekins and mugs once their names are found; black book already works 3/3 (S11).
\end{quote}
Alice therefore selects a task whose perception already works, while explicitly proposing a grasp that could be reused beyond that task.

\minisection{Practice task and reusable parts}
The selected task is ``Pick the black bowl and place it in the basket,'' among other bowls, a ramekin, and a mug. The detector, basket localization, and pick-and-place sequence are already available. The new requirement is to hold a wide, hollow object and account for the offset between the grasp and its centre. This task and the two transfer tasks below differ from the official Object tests, whose targets are grocery items.

\minisection{Physical revision}
P6 adds an open-container family keyed by \texttt{bowl}, \texttt{ramekin}, \texttt{mug}, or \texttt{cup}, together with selection using the full item name and colour. It estimates both rim crossings, pinches the wall on the side nearer the robot's centreline, and carries higher. It offsets the release position so that the container's centre, rather than the hand, is over the basket:

\begingroup\footnotesize
\begin{verbatim}
# P6: rim grasp and offset placement (excerpts)
OPEN_NOUNS = ("bowl", "ramekin", "mug", "cup")
...
if container:
    target["top_z"] = target["z_max"]
    y_lo, y_hi = target["rim_y"]
    if abs(y_hi) <= abs(y_lo):
        hold_y = y_hi - 0.004
    else:
        hold_y = y_lo + 0.004
    side_y = hold_y - target["cy"]
    ...
    target["cy"] = hold_y
    grasp_z = max(target["top_z"] - 0.022,
                  target["floor_z"] + 0.012)
...
if container:
    basket["cy"] = basket["cy"] + side_y
    transit_z = (basket["rim_z"]
                 + bottom_offset + 0.13)
    release_z = basket["rim_z"] + 0.03
\end{verbatim}
\endgroup
Figure~\ref{fig:coopqual:rim} shows the change from an empty grasp to holding the rim, on both the practiced bowl and an unpracticed mug task.

\minisection{Outcomes, retention, and generalization}
Practice and fresh-layout success each rise from 0/5 to 5/5. Original checks remain 5/5, an additional ten-episode retention check remains 10/10, and all five earlier tasks remain 5/5 each. Without being selected for repair, the red-coffee-mug and white-bowl tasks each improve from 0/3 to 3/3. The mug episodes execute the new rim-grasp branch; the white-bowl episodes also correct the earlier selection of the milk. This supports reuse of both selection and grasping beyond the practiced black bowl.

\subsubsection{Spatial / ASPIRE: a grasp-side rule carries across instructions}
\label{app:coopqual:side}

\begin{figure*}[t]
\centering
\small\textbf{Practice: bowl between cookies box and stove / before, P7}\\[2pt]
\coopframe{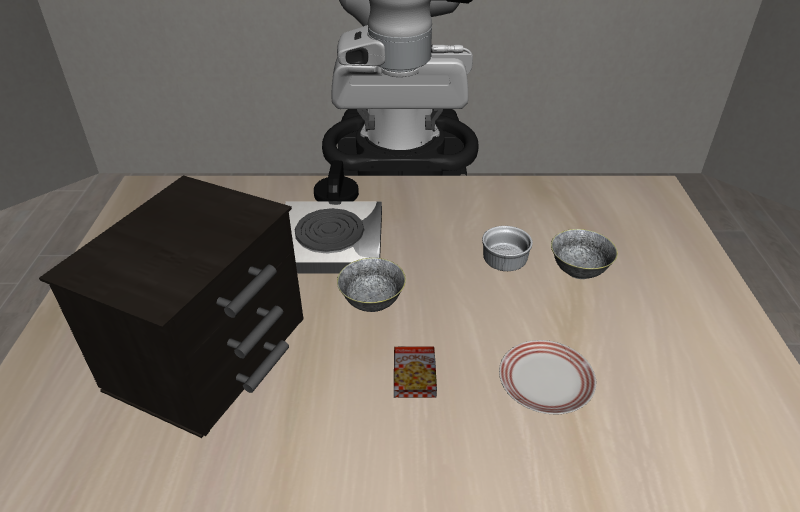}{Start}\hfill
\coopframe{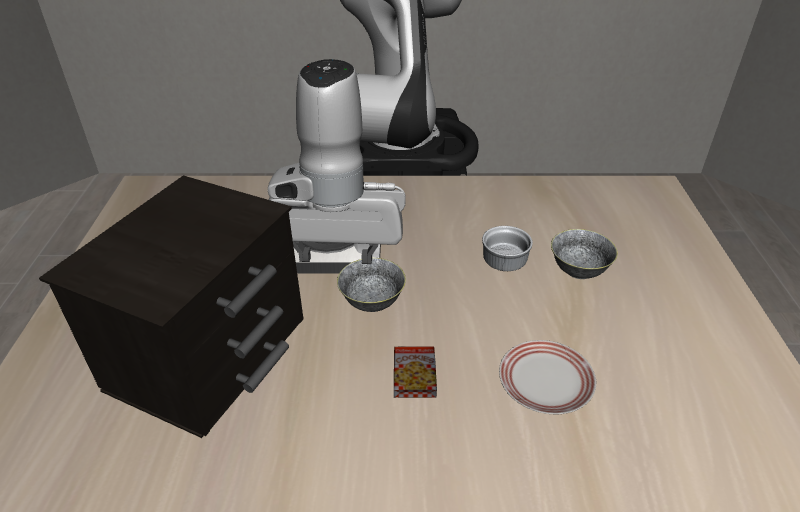}{Descent stalls}\hfill
\coopframe{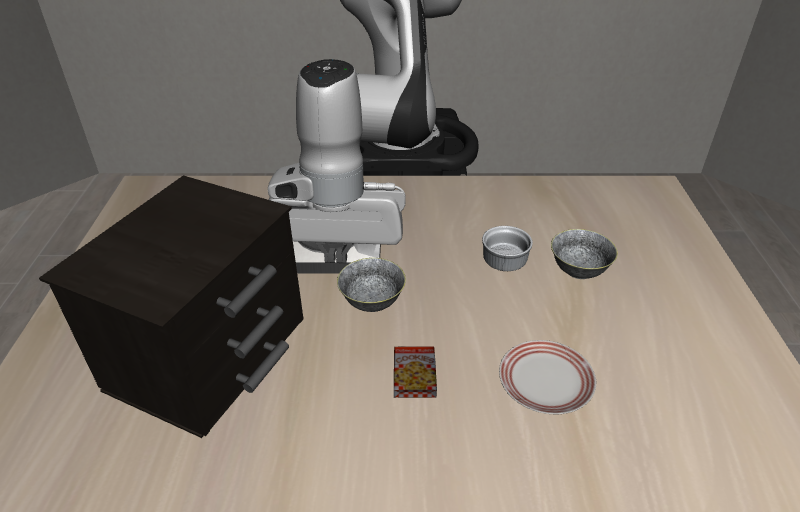}{Empty grasp}\hfill
\coopframe{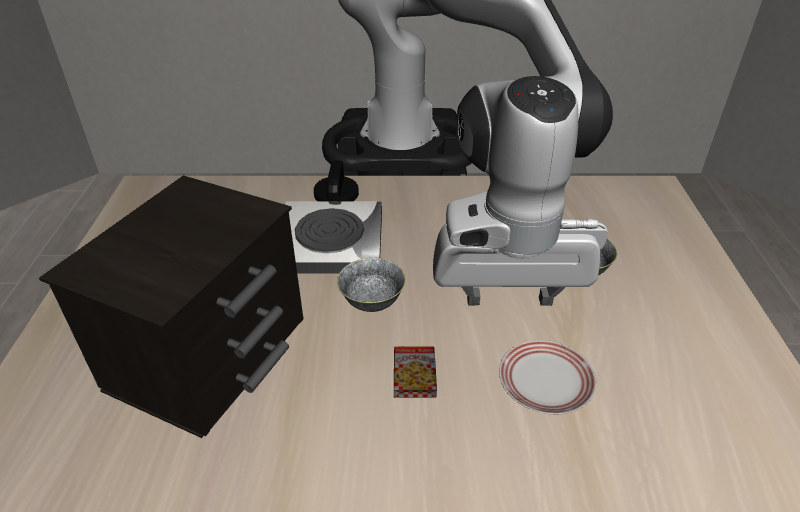}{Bowl left behind}
\par\smallskip
\textbf{Practice: same layout / after, P8}\\[2pt]
\coopframe{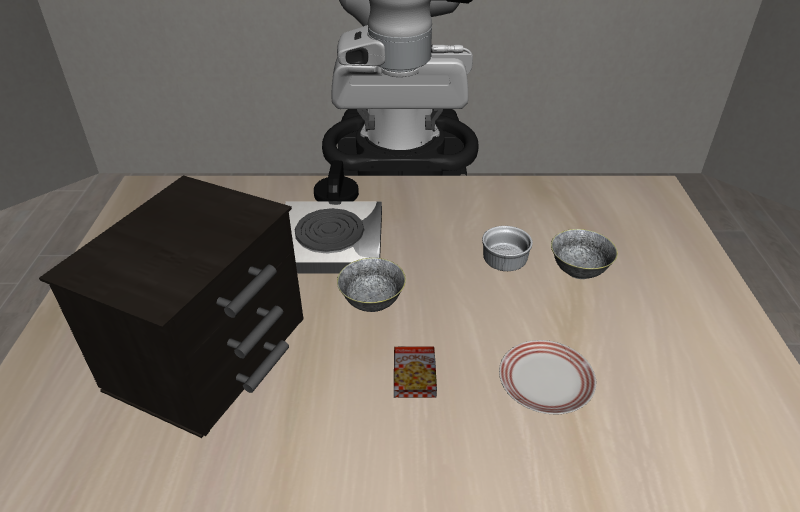}{Start}\hfill
\coopframe{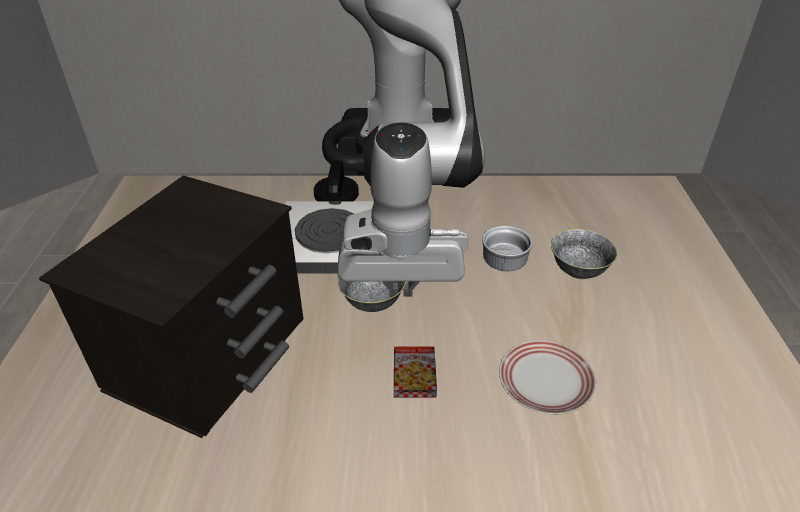}{Opposite rim grasped}\hfill
\coopframe{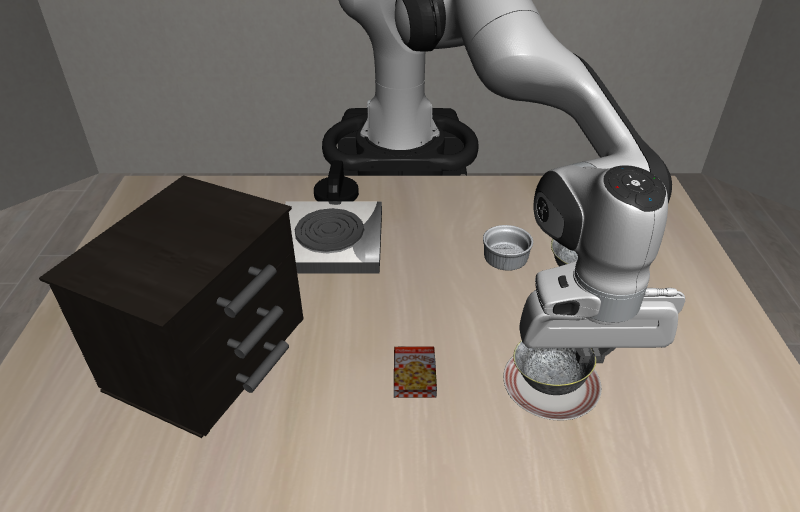}{Over plate}\hfill
\coopframe{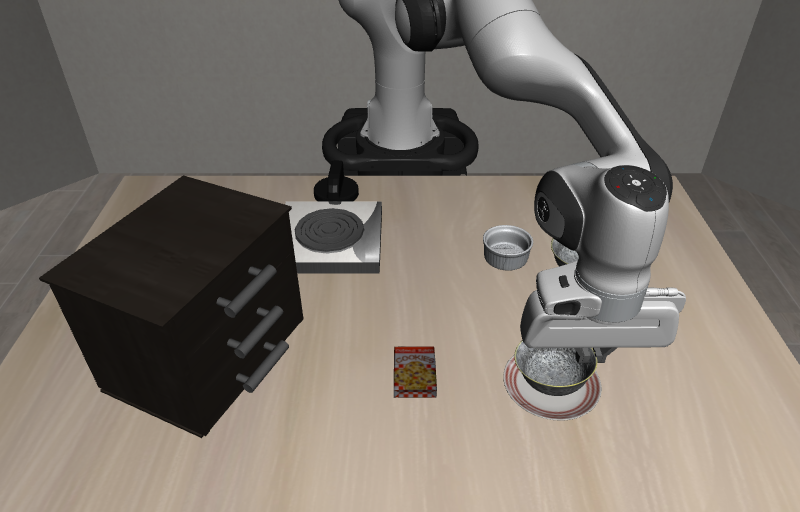}{Bowl on plate}
\par\smallskip
\textbf{Unpracticed: bowl next to wooden cabinet / before, P7}\\[2pt]
\coopframe{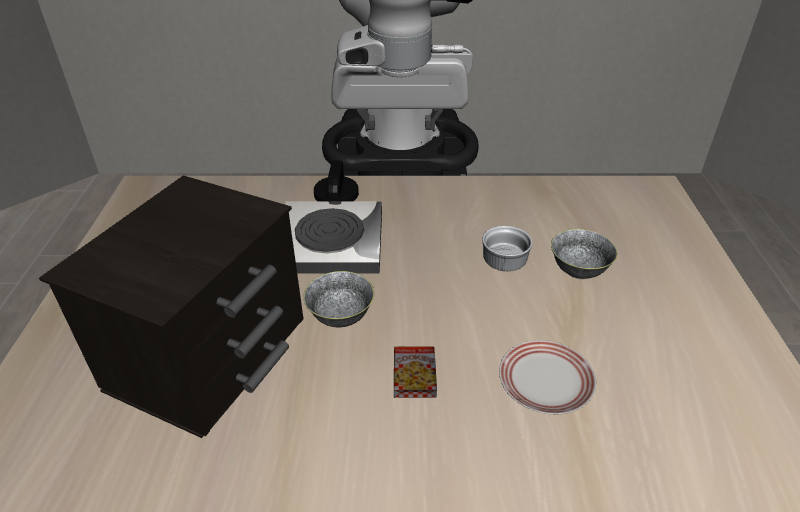}{Start}\hfill
\coopframe{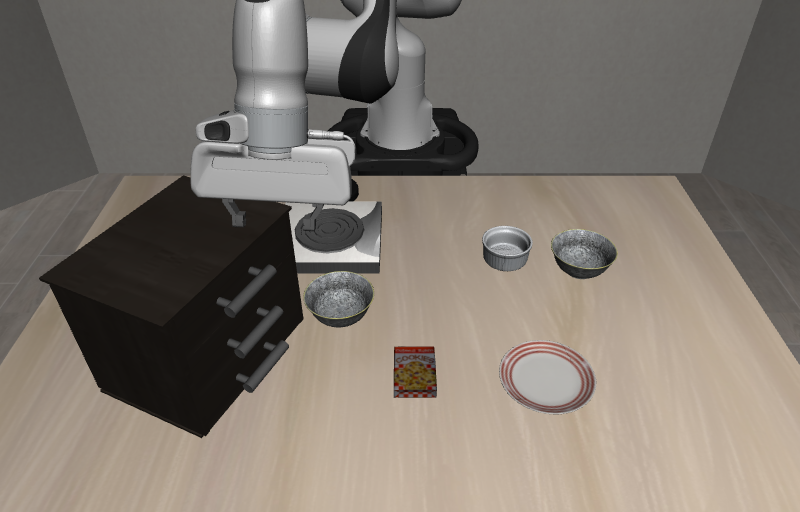}{Descent stalls}\hfill
\coopframe{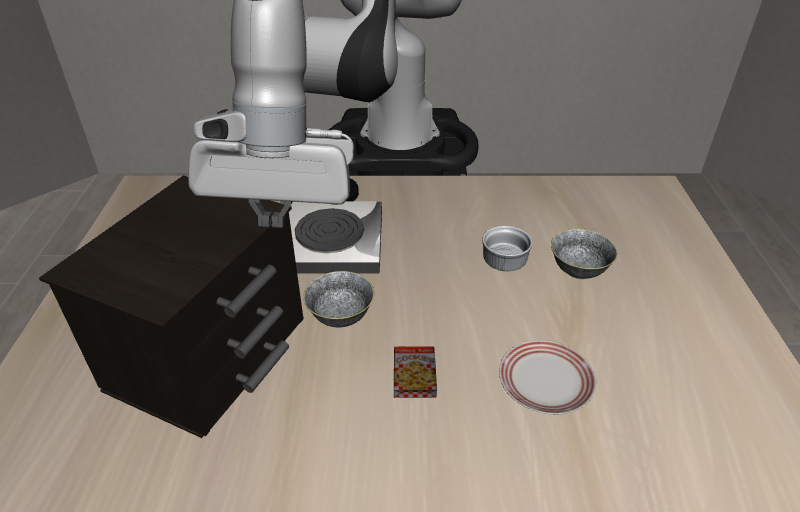}{Empty grasp}\hfill
\coopframe{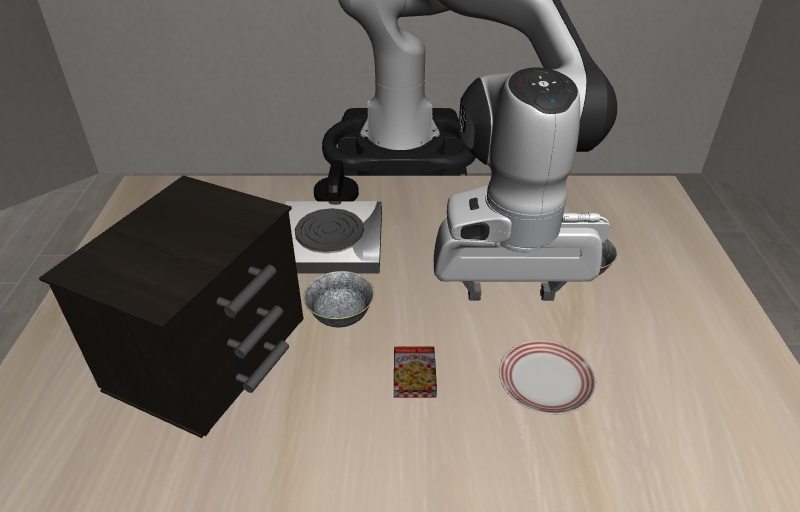}{Bowl left behind}
\par\smallskip
\textbf{Unpracticed: same layout / same repair, P8}\\[2pt]
\coopframe{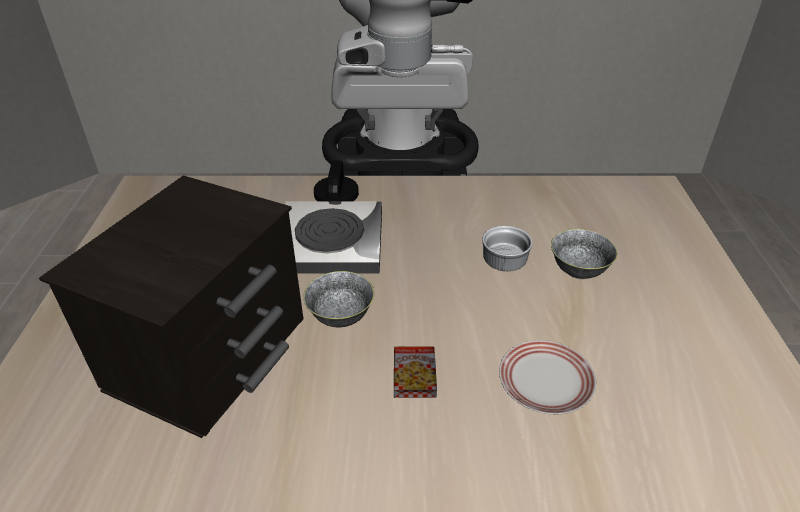}{Start}\hfill
\coopframe{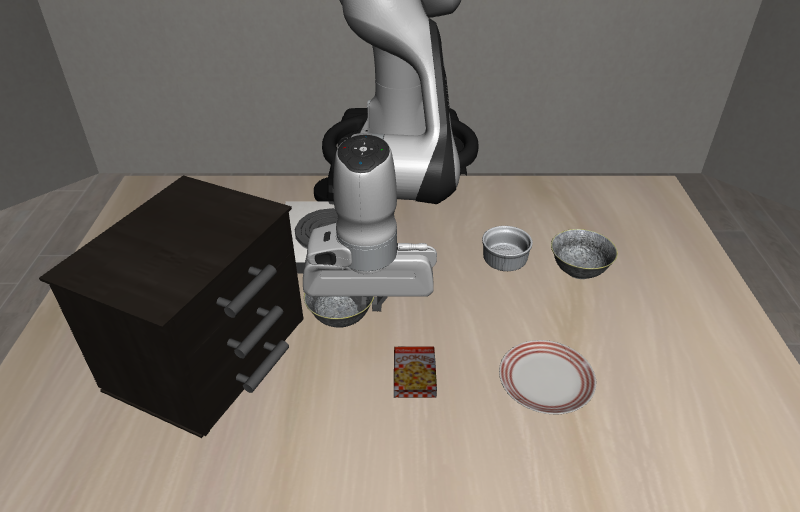}{Opposite rim grasped}\hfill
\coopframe{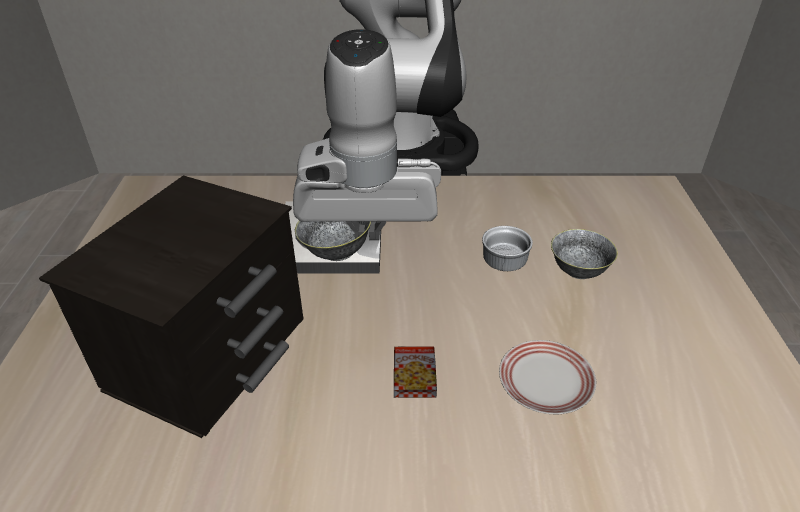}{Lift}\hfill
\coopframe{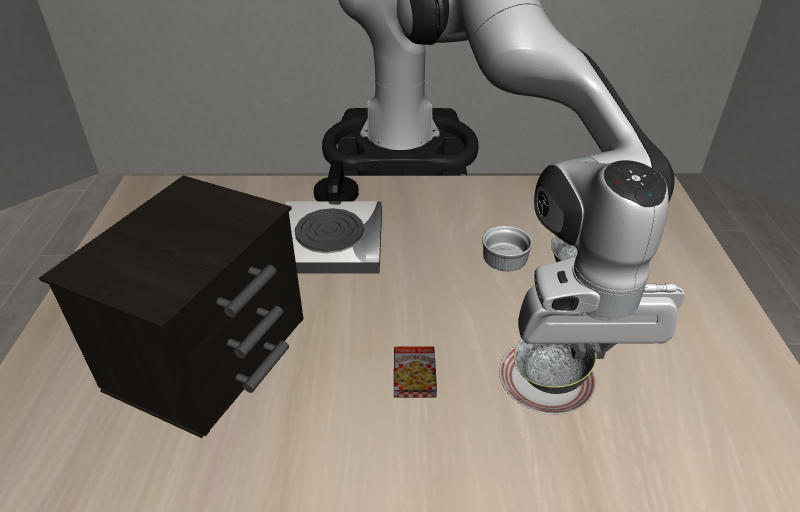}{Bowl on plate}
\caption{\textbf{Reusing a safer rim choice with ASPIRE as the harness.} P7 chooses the cabinet-side rim and fails to acquire the bowl. P8 chooses the opposite rim. The transfer task changes both the instruction's selection rule and the target position; its three successful P8 episodes report \texttt{side\_flipped: true}.}
\label{fig:coopqual:side}
\end{figure*}

\minisection{Current capability}
In cycle 8 of LIBERO-Spatial, P7 already selects the bowl ``between the cookies box and the stove'' and solves its earlier layout in 5/5 episodes. Its rim-grasp rule chooses the side whose outer finger is farthest from nearby object centres. The cabinet is absent from this list. A safer preference for the rim nearer the robot's centreline exists only for \texttt{table\_center} instructions:

\begingroup\footnotesize
\begin{verbatim}
# P7: rim-side choice (lines 387--408)
radius = 0.5 * bowl["wy"]
neigh = ((other, plate) if on_ramekin
         else (ramekin, other, plate))
best_side = None
best_clear = None
for side in (1.0, -1.0):
    fx = bowl["cx"]
    fy = bowl["cy"] + side * (radius + 0.03)
    clear = min(math.hypot(fx - o["cx"],
                          fy - o["cy"])
                for o in neigh)
    if best_clear is None or clear > best_clear:
        best_clear = clear
        best_side = side
side = best_side
if rule == "table_center":
    # ... existing centreline-rim preference ...
\end{verbatim}
\endgroup

\minisection{Alice's reasoning}
Alice compares failures at the same target position with different positions of the other bowl, probes perception, and reads the selection and grasp code. Alice's hypothesis separates the motion failure from parsing and selection:
\begin{quote}\small
Motion, not wording or selection. The rim-side choice (lines 387-408) scores the two Y rims only against point centres of ramekin, other bowl and plate; the wooden cabinet is not a neighbour. For a table bowl in front of the cabinet (x inside its x-range, y about -0.05) all neighbours lie at +Y, so it takes the -Y rim (hand at y about -0.115, facing the cabinet front and drawer handles at y -0.15) and the descent stalls three times in the plain branch (lines 472-489); the grip closes empty.
\end{quote}
Alice's rationale uses the screening controls to isolate the side choice:
\begin{quote}\small
The wording is already trained (5/5 at target (-0.13,-0.05)). Moving the bowl 4.5 cm to (-0.085,-0.05), in front of the cabinet, gives 0/3 (S7). The same start with the other bowl on the cabinet top, where the code happens to pick the +Y rim, gives 3/3 (S9); table-center at (-0.075,-0.05), which uses the centreline-rim rule, gives 3/3 (S10). So one change in the side choice should fix it.
\end{quote}
Changing the other bowl affects the side chosen by the existing clearance calculation. This comparison, together with the successful \texttt{table\_center} branch, points to a reusable motion rule.

\minisection{Practice task and reusable parts}
The instruction remains ``Pick the akita black bowl between the cookies box and the stove and place it on the plate.'' The target moves 4.5\,cm from $(-0.13,-0.05)$ to $(-0.085,-0.05)$, in front of the cabinet. Parsing, landmark localization, bowl selection, transport, and placement are reused. This selected instruction and the transfer instruction below differ from the official Spatial test instructions.

\minisection{Physical revision}
P8 extends the safer side choice to other selection rules for bowls on the table. If the chosen rim has $y<-0.06$\,m, it uses the opposite rim when that side is nearer the robot's centreline and its outer finger has at least 0.12\,m clearance. The selection and downstream motion code remain unchanged:

\begingroup\footnotesize
\begin{verbatim}
# P8: added after the existing side choice
if (rule != "table_center" and (not raised)
    and (not on_ramekin)
    and bowl["top"] - BOWL_H - table_z < 0.02
    and bowl["cy"] + side * radius < -0.06):
    fy = bowl["cy"] - side * (radius + 0.03)
    clear = min(math.hypot(bowl["cx"] - o["cx"],
                          fy - o["cy"])
                for o in neigh)
    report["flip_clear"] = clear
    if (clear >= 0.12
        and abs(bowl["cy"] - side * radius)
        < abs(bowl["cy"] + side * radius) - 0.005):
        side = -side
        report["side_flipped"] = True
\end{verbatim}
\endgroup
The revision extends an existing geometric preference to other selection rules. Figure~\ref{fig:coopqual:side} shows the resulting change in approach.

\minisection{Outcomes, retention, and generalization}
Practice and fresh-layout success each rise from 0/5 to 5/5; original checks remain 5/5, and all seven earlier practice tasks retain their scores. The unpracticed task ``Pick the akita black bowl next to the wooden cabinet and place it on the plate'' improves from 0/3 to 3/3. It places the bowl at $(-0.05,-0.10)$ and selects it through \texttt{cabinet\_level}, rather than \texttt{between}. All three P8 episodes execute the new side-flipping branch, with measured clearance of 0.23--0.26\,m. Thus the physical rule carries to a different instruction and position.

\subsubsection{Goal / \method{}-Bob: a handle-dependent grasp carries to a different mug}
\label{app:coopqual:mug}

\begin{figure*}[t]
\centering
\small\textbf{Practice: white porcelain mug to stove / P8, handle $+y$, grasp $-y$}\\[2pt]
\coopframethree{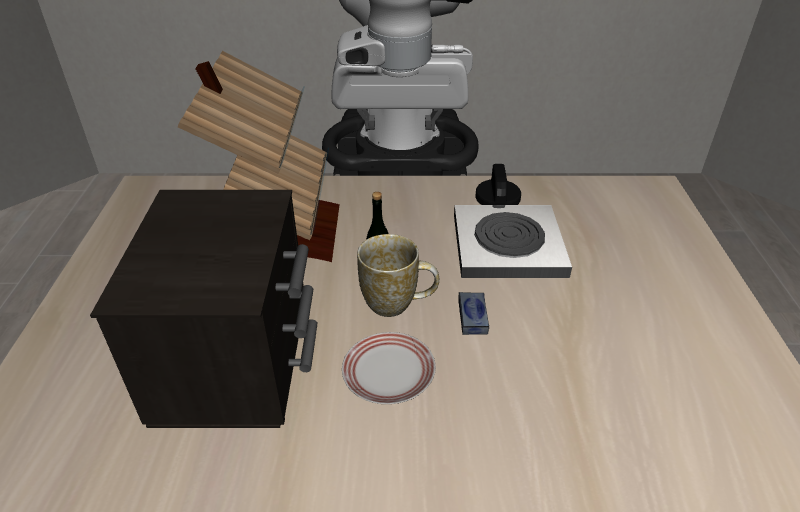}{Initial scene}\hfill
\coopframethree{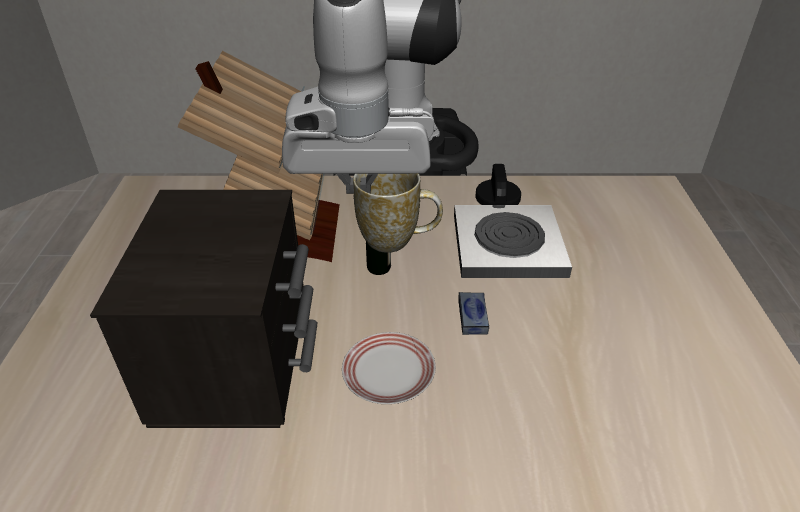}{Lift}\hfill
\coopframethree{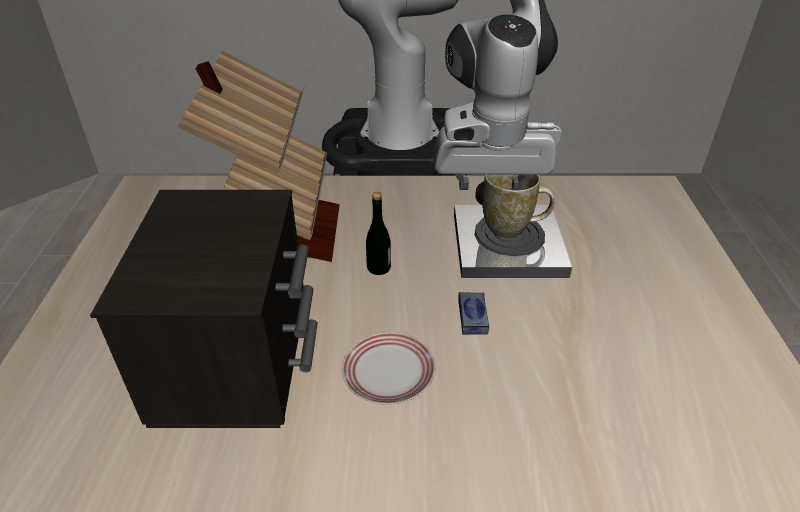}{Placement on stove}
\par\smallskip
\textbf{Unpracticed: red coffee mug to plate / same P8, handle $-y$, grasp $+y$}\\[2pt]
\coopframethree{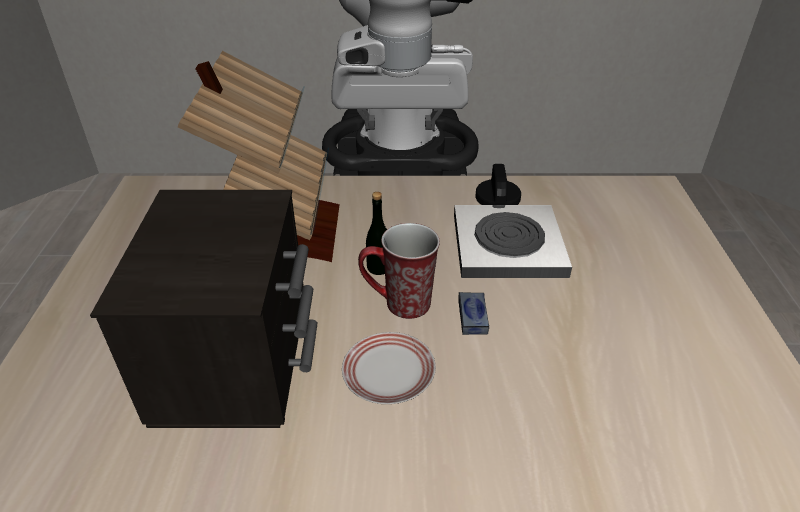}{Initial scene}\hfill
\coopframethree{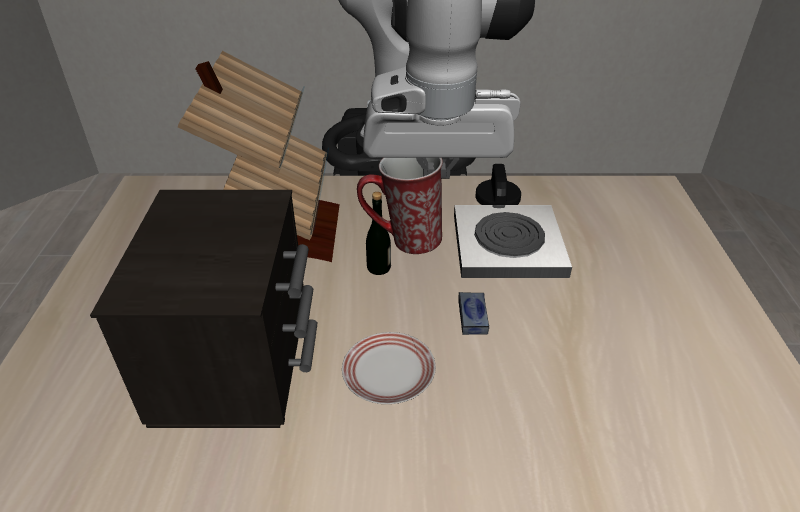}{Lift}\hfill
\coopframethree{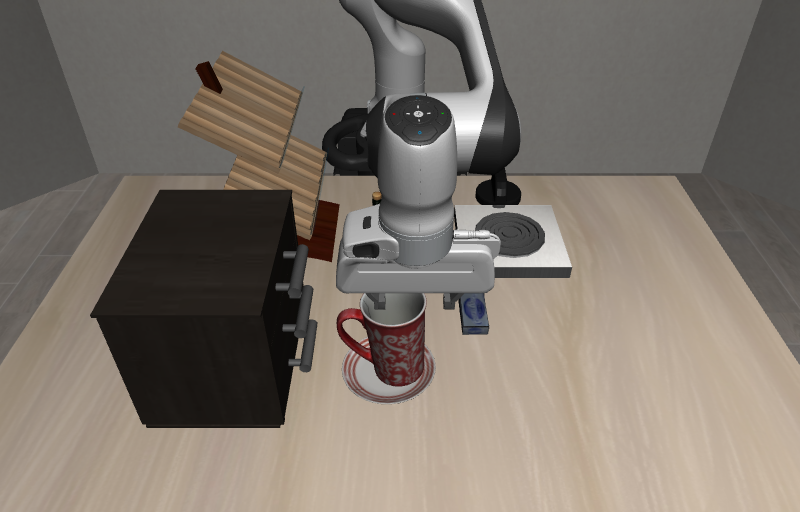}{Placement on plate}
\caption{\textbf{Adapting the grasp to the handle with \method{}-Bob.} Both rows use the same frozen P8. The second task changes the mug, handle direction, and destination; Alice screens it in the next cycle but never selects it for repair. P8 succeeds in 3/3 screens of the second task.}
\label{fig:coopqual:mug}
\end{figure*}

\minisection{Current capability}
In cycle 11 of LIBERO-Goal, P7 already grasps bowl-like objects and uses the stove and plate destinations. Its rim routine chooses a side by retry index, without estimating a handle. A separate item guard rejects mugs before any motion:

\begingroup\footnotesize
\begin{verbatim}
# P7: grasp side (lines 589--593)
side = 1.0 if attempt == 0 else -1.0
rim_y = bowl["y99"] if side > 0 else bowl["y01"]
gx = bowl["cx"]
gy = rim_y - side * 0.006
gz = bowl["ztop"] - (0.013 + 0.003 * attempt)
...
# P7: item guard (lines 769--771)
def run_pick_and_place(item, dest, level=None):
    if not is_bowl_like(item):
        raise ValueError(
            "pick_place supports only bowl-like "
            "items (bowl, <colour> bowl, ramekin); "
            "got " + str(item) + " -> " + str(dest))
\end{verbatim}
\endgroup

\minisection{Alice's reasoning}
Alice reads the program, screens candidate tasks, and probes mug perception. Alice's submitted hypothesis explicitly identifies both limitations:
\begin{quote}\small
run\_pick\_and\_place rejects every item that is not bowl-like (line 770-771: is\_bowl\_like checks name endswith bowl/ramekin) and item\_queries/BOWL\_QUERIES (line \textasciitilde{}420) and the rim grasp in subtask\_grasp\_bowl\_rim assume a bowl. 'Put the white porcelain mug on the stove' raises ValueError before any motion (0/3, no item moved).
\end{quote}
Alice's rationale selects a nearby extension that reuses the current chain:
\begin{quote}\small
Bowl/ramekin on plate works 3/3 (red ramekin), so the pick-and-place chain, stove destination and rim grasp already work. A mug is the nearest non-bowl item: a cylinder grasped top-down by its rim. Perception probe: query 'mug' scores 0.92 on the mug in the exterior and 0.94 on the wrist camera ('white porcelain mug' only 0.19). Smallest extension: let the guard accept mugs, using 'mug' as the query. Carries to mugs/moka pots on plate, cabinet top, shelf top, microwave top and other non-bowl items.
\end{quote}
The proposal thus connects the unsupported item type to a physical adaptation of the existing grasp.

\minisection{Practice task and reusable parts}
The task is ``Put the white porcelain mug on the stove.'' It preserves the known stove destination and pick-and-place workflow while introducing a taller vessel with a protruding handle. Both this task and ``Put the red coffee mug on the plate,'' used below, differ from the official Goal test instructions.

\minisection{Physical revision}
P8 admits mugs, changes the perception queries, and estimates the mug body from the upper 1.2\,cm of its point cloud. Comparing the full cloud with this rim estimates the handle's protrusion direction. The grasp chooses the opposite rim, with mug-specific depth and lifted-object checks:

\begingroup\footnotesize
\begin{verbatim}
# P8: isolate the upper rim (lines 437--438)
top = st["ztop"]
rim = [p for p in pts if p[2] >= top - 0.012]
...
# P8: mug grasp (lines 663--669)
hd = bowl.get("handle_dir")
if hd == "+y":
    side = -1.0
elif hd == "-y":
    side = 1.0
depth = 0.02 - 0.008 * attempt
rim_y = bowl["y99"] if side > 0 else bowl["y01"]
\end{verbatim}
\endgroup
In the white-mug execution, the estimated handle points in $+y$ and Bob grasps the $-y$ rim. In the red-mug execution, the handle points in $-y$ and the same frozen P8 chooses the $+y$ rim (Figure~\ref{fig:coopqual:mug}). The contact changes with the observed geometry while destination handling is reused.

\minisection{Outcomes, retention, and generalization}
White-mug practice improves from 0/5 to 4/5. Original drawer checks remain 5/5. In cycle 12, the same P8 succeeds in 3/3 screens of the red-coffee-mug-to-plate task, which is never selected for repair. The handle-dependent grasp is reused across a different mug, handle direction, and destination.
\subsubsection{LIBERO-10 / \method{}-Bob: learned placement routines compose across tasks}
\label{app:coopqual:long}

\begin{figure*}[t]
\centering
\small\textbf{Practice, cycle 8: white mug to cabinet / repaired P6}\\[2pt]
\coopframethree{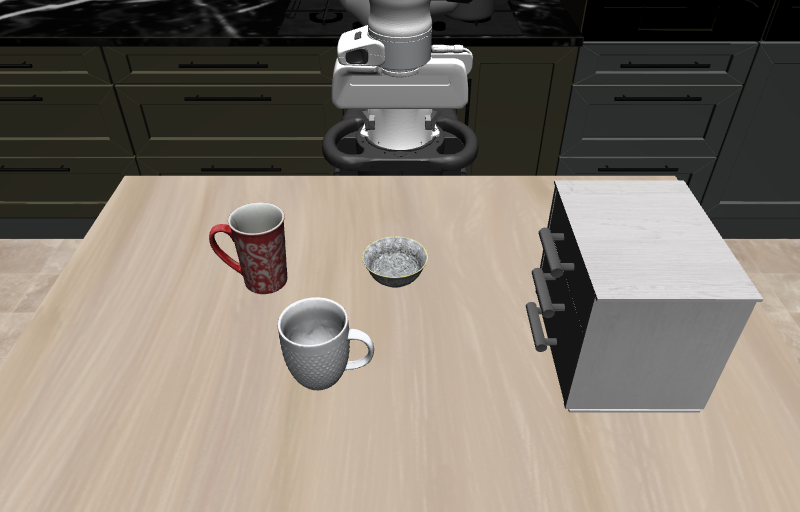}{Initial scene}\hfill
\coopframethree{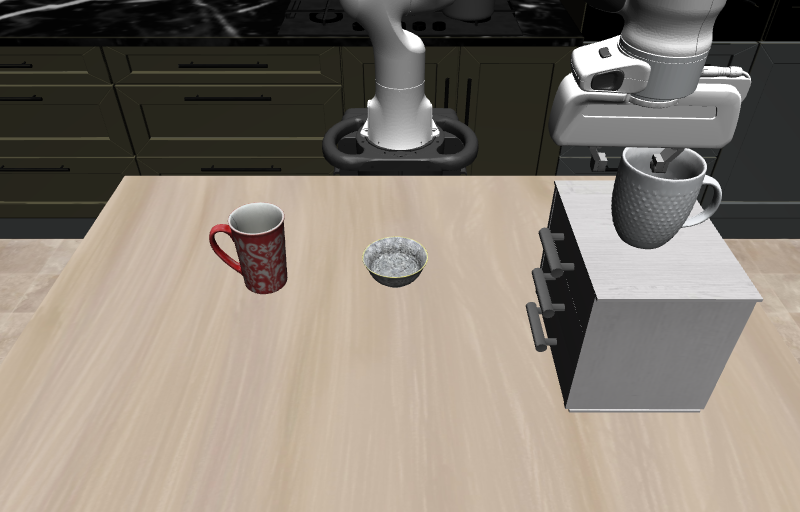}{Mug above the cabinet}\hfill
\coopframethree{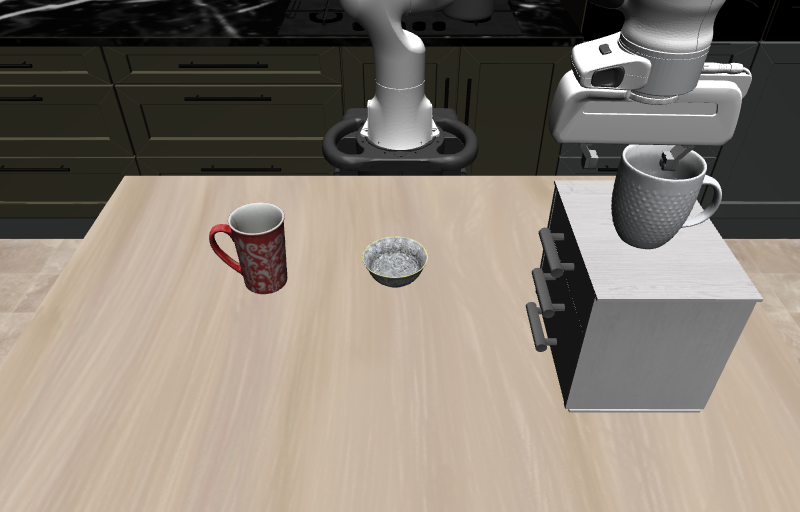}{Mug placed on top}
\par\smallskip
\textbf{Unpracticed composition: white mug to stove, red mug to cabinet / P11}\\[2pt]
\coopframethree{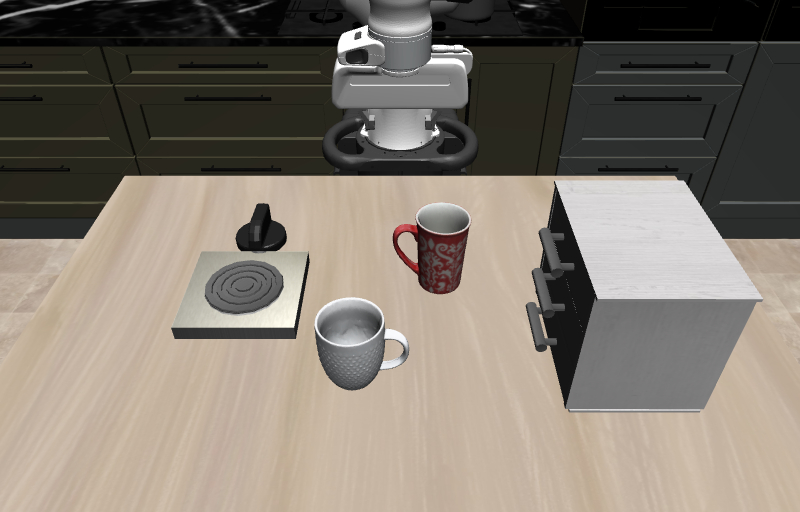}{Initial scene}\hfill
\coopframethree{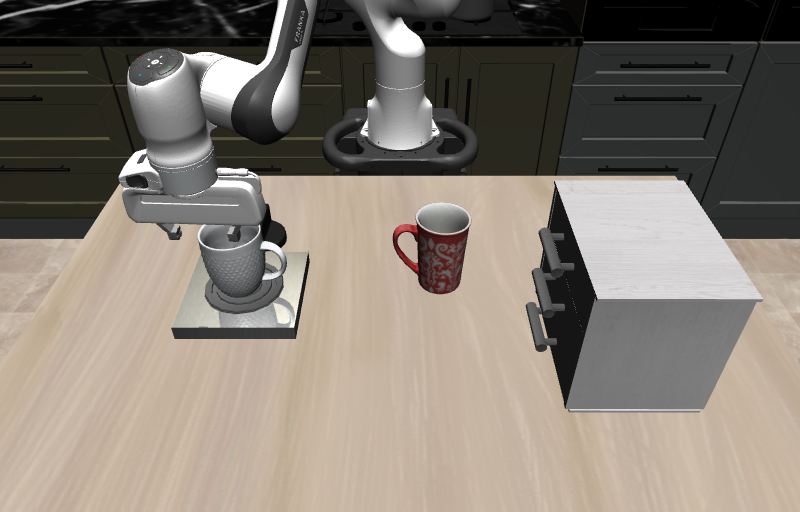}{White mug placed on stove}\hfill
\coopframethree{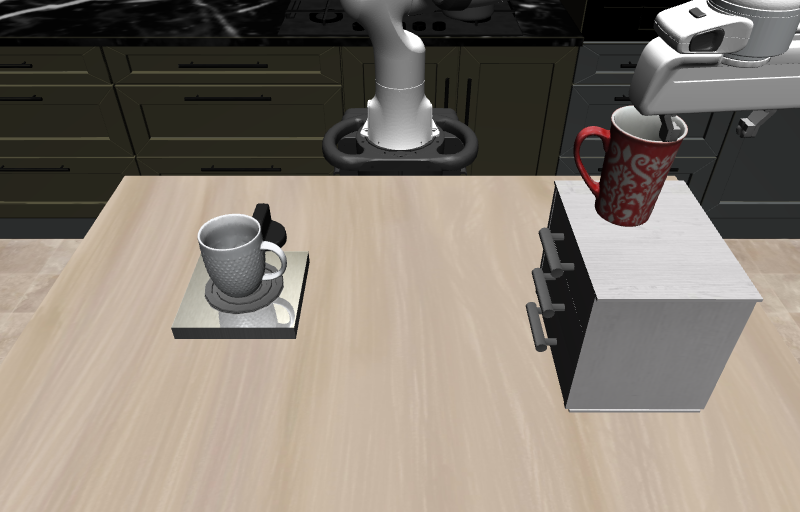}{Red mug also placed on cabinet}
\par\smallskip
\textbf{Unpracticed composition: red mug to stove, white mug to cabinet / same P11}\\[2pt]
\coopframethree{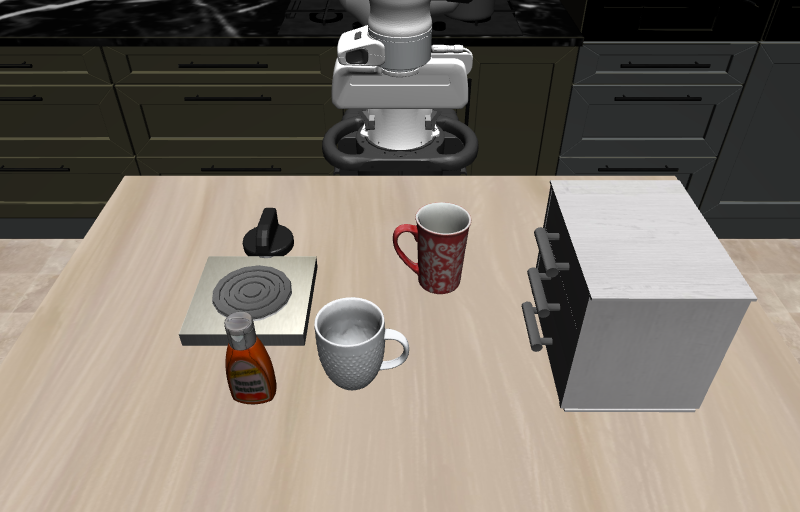}{Initial scene}\hfill
\coopframethree{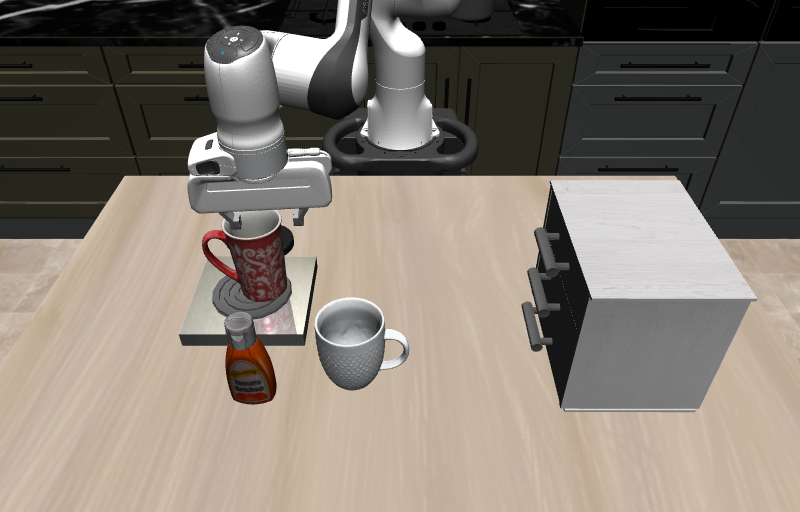}{Red mug placed on stove}\hfill
\coopframethree{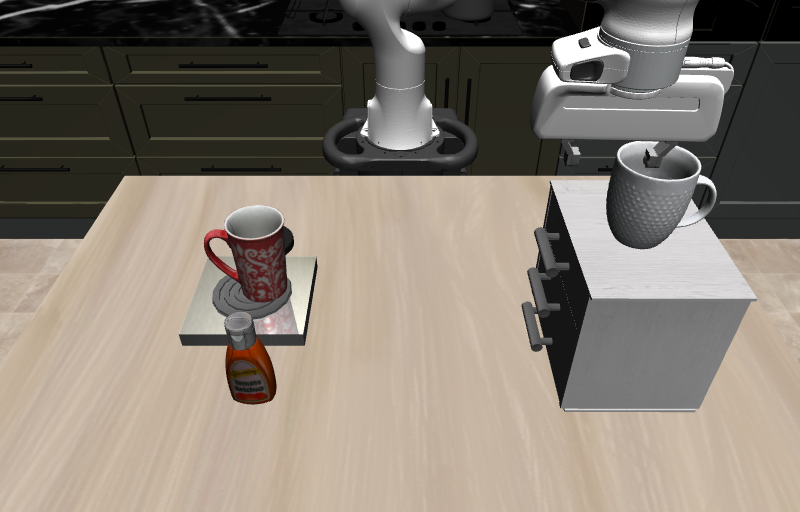}{White mug also placed on cabinet}
\caption{\textbf{Reusing a physical placement routine in new task compositions with \method{}-Bob.} The first row shows cabinet-placement practice. The remaining rows show two compositions that Alice screens but never selects for repair, executed by the same frozen P11. The cabinet perception, transport, release, and verification routines remain unchanged from P6. Both placement goals hold at the recorded end of each composition.}
\label{fig:coopqual:long}
\end{figure*}

\minisection{Current capability}
In cycle 7 of LIBERO-10, P5 places the white mug on the stove in 5/5 practice layouts. Bob can already find the mug, grasp its rim, carry it, and release it on a low support. Its parser accepts plates and the stove, and its placement workflows use destination-specific heights. It cannot yet place the mug on top of a cabinet:

\begingroup\footnotesize
\begin{verbatim}
# P5: parse_put_on, destination handling (excerpt)
elif dest in STOVE_DEST_NAMES:
    target = "stove"
else:
    raise Exception(
        "parser: unsupported destination '%s' "
        "in '%s'" % (dest, language))
\end{verbatim}
\endgroup
The cabinet request therefore stops before motion. Supporting it requires both admitting the new destination and defining how to carry the held mug above a raised surface.

\minisection{Alice's reasoning}
In cycle 8, Alice receives an excerpt of the current program and its subtask list, then screens cabinet and microwave-top placement. Alice's recorded proposal identifies the missing destination support and connects it to the existing workflows:
\begin{quote}\small
parse\_put\_on (lines \textasciitilde{}190-215) only accepts destinations plate/stove and raises 'unsupported destination' for cabinet (and microwave top) before any motion; transport/release exist only for plate and stove (subtask\_transport\_to\_plate/stove, release\_on\_*), with fixed heights.
\end{quote}
Alice's rationale states the intended incremental extension:
\begin{quote}\small
Same kind as the already-working mug-on-stove (kitchen, mug, put\_on): only the destination changes.
\end{quote}
Alice thus reuses the successful mug grasp while targeting destination geometry that the current policy does not handle.

\minisection{Practice task and reusable parts}
The task is ``Put the white mug on top of the cabinet.'' The object and grasp are reused; the changing axis is the destination and its height. This task and the two composition probes below differ from the official LIBERO-10 test instructions.

\minisection{Physical revision}
P6 recognizes the destination and estimates the cabinet's top surface from camera and depth observations. It reuses mug acquisition, but computes transport height from the observed surface and the distance from the gripper to the mug's bottom. If the current hand is too low, it first lifts vertically, then moves above the destination:

\begingroup\footnotesize
\begin{verbatim}
# P6: subtask_transport_to_top (excerpts)
z = max(ctx["transit_z"],
        top["surface_z"] + ctx["held_drop"] + 0.06)
...
cur = (get_robot_state()
       .end_effector_poses["primary"].position)
if cur[2] < z - 0.02:
    if not go_to(ctx, cur[0], cur[1], z, 0.02, 250):
        ...
        return False
ok = go_to(ctx, tx, ty, z, 0.02, 300)
\end{verbatim}
\endgroup
Here \texttt{held\_drop} accounts for the held object's extent below the hand. In the displayed episode, the observed cabinet top is 0.215\,m high and the mug extends 0.080\,m below the hand, giving a transport height of 0.355\,m. Bob then lowers the mug onto the surface and releases it. The simulator confirms successful placement.

\minisection{Outcomes, retention, and composition}
The cabinet task improves from 0/5 to 5/5 on practice layouts and from 0/5 to 5/5 on fresh layouts. The original task (placing both the alphabet soup and tomato sauce in the basket) remains 5/5, and an additional ten-episode retention check remains 10/10.

Later in the curriculum, Alice screens two compositions, neither of which is selected for repair. They exchange which mug goes to the stove and which goes to the cabinet. The same final P11 succeeds on both:

\begin{center}
\footnotesize
\begin{tabularx}{\linewidth}{@{}>{\raggedright\arraybackslash}Xr@{}}
\toprule
Task & Success \\
\midrule
Practice: white mug on cabinet (cycle 8) & P6: 5/5 \\
Unpracticed: white mug on stove;\newline red mug on cabinet & P11: 3/3 \\
Unpracticed: red mug on stove;\newline white mug on cabinet & P11: 3/3 \\
\bottomrule
\end{tabularx}
\end{center}

Both placement goals hold at the recorded end of all six P11 episodes. Five cabinet routines---surface detection, transport, release, placement verification, and their coordinating workflow---have identical function bodies in P6 and P11. The later multitask program therefore reuses the learned physical behavior unchanged when the instruction combines two placements and assigns a different mug to the cabinet.

\endgroup

\section{Benchmarks}
\label{sec:appendix:benchmarks}

\subsection{LIBERO Suites}
\label{app:libero_suites}
Each of the four suites of LIBERO~\cite{liu2023libero} has ten tasks, and each task has saved start states.
Bob refines on start states 10 to 24 of a task, the development scenes, and the held-out results use start states 25 to 39 (Table~\ref{tab:settings-evaluation}).
The tasks of a suite share most of their sub-goals and differ as follows.

\minisection{LIBERO-Spatial}
Every task picks up one of two identical black bowls and places it on the plate.
The instruction names the bowl by where it is, for example between the plate and the ramekin, next to the cookie box, on the stove, or in the top layer of the wooden cabinet, and the layout of the scene differs between tasks (Table~\ref{tab:coop-pertask-spatial}).

\minisection{LIBERO-Object}
Every task picks up one grocery item and places it in the basket, and the tasks differ in the item: alphabet soup, cream cheese, salad dressing, BBQ sauce, ketchup, tomato sauce, butter, milk, chocolate pudding, or orange juice.
The other items of the scene are distractors (Table~\ref{tab:coop-pertask-object}).

\minisection{LIBERO-Goal}
All tasks share one scene with a cabinet with drawers, a stove, a wine rack, a bowl, a plate, a wine bottle, and a cream cheese box, and differ in the goal: opening a drawer, putting an object on or in a fixture, pushing the plate, or turning on the stove (Table~\ref{tab:coop-pertask-goal}).
Most of these tasks still share the sub-goals of finding, grasping, and placing an object, and the others differ in their final sub-goals, such as pushing the plate or turning on the stove.

\minisection{LIBERO-10}
The ten long-horizon tasks take place in different scenes, such as a kitchen, a living room, and a study, and most of them chain two sub-tasks, such as putting two objects in the basket, or turning on the stove and putting the moka pot on it (Table~\ref{tab:coop-pertask-long}).
A run of LIBERO-10 has up to 4000 simulator steps, against 1000 in the other suites.

\subsection{Robustness Benchmarks}
\label{app:robustness_benchmarks}
We evaluate the frozen policies on LIBERO-Pro, LIBERO-Plus, and LIBERO-MAX across the 40 LIBERO tasks: ten perturbation types in LIBERO-Pro, perturbation categories and difficulty levels in LIBERO-Plus, and eight dynamic events in LIBERO-MAX.
All methods use the same test cases for each task.

\minisection{LIBERO-Pro}
LIBERO-Pro~\cite{liberopro2025} has four original perturbations: \emph{object}, which replaces the task objects with models of another colour or size; \emph{position}, which swaps the positions of the objects; \emph{semantic}, which paraphrases the instruction; and \emph{task}, which asks for a different goal in the same scene.
Its official repository also provides six extensions added by the community, which we use as well: \emph{camera view angle}, a fixed oblique camera pose; \emph{initial pose}, which starts the target object 18\,cm away and rotated by 45°; \emph{object shape}, which shrinks the target object to half its size; \emph{object texture}, which recolours the destination to match the table or the floor; \emph{view occlusion}, a large book between the camera and the target; and \emph{visual noise and glare}, dim lighting with image noise.

\minisection{LIBERO-Plus}
LIBERO-Plus~\cite{liberoplus} has seven perturbation categories: background textures, camera viewpoints, language instructions, light conditions, object layout, robot initial states, and sensor noise, each labelled with a difficulty level from 1 to 5.

\minisection{LIBERO-MAX}
LIBERO-MAX~\cite{liberomax} has eight dynamic events, each at a mild or a strong level: \emph{camera shift}, where the camera moves and turns as the gripper nears the target; \emph{distractor burst}, where five objects appear around the target; \emph{illumination switch}, where the light drops or rises during the run; \emph{obstacle insertion}, where an object appears on the approach path; \emph{receptacle relocation} and \emph{target relocation}, where the destination or the target object moves by 6 or 12\,cm; \emph{sensor noise onset}, where image noise and a black patch appear; and \emph{visual theme switch}, where the colours of the scene change.
Every case is run twice from the same start state, setup, and seed: without its event (Base) and with it (Dynamic).
The Base scene is itself a perturbed static scene from LIBERO-Pro or LIBERO-Plus.

\subsection{RoboTwin 2.0}
\label{app:robotwin}
RoboTwin 2.0~\cite{robotwin2} is a simulated benchmark of 50 manipulation tasks for a dual-arm robot.
Each task has a clean setting and a domain-randomized setting.
The randomized setting adds distractor objects on the table, random table and background textures, random lighting, and a random table height, and rewords the instruction; its textures and instruction templates are split into a seen set and an unseen set reserved for testing.
We use 14 tasks: Adjust Bottle, Beat Block Hammer, Click Bell, Handover Block, Move Can Pot, Move Pillbottle Pad, Open Laptop, Place Container Plate, Place Mouse Pad, Place Object Scale, Place Phone Stand, Place Shoe, Press Stapler, and Stack Blocks Two (Table~\ref{tab:robotwin}).
Bob refines on clean scenes only.
The test set of a task has 100 scenes from the randomized setting, with the unseen textures and instruction templates and with start states disjoint from those seen during refinement; we keep only start states that pass the official validity check of RoboTwin 2.0.
An execution may take up to 10,000 actions, and we also report success within the step limits of RoboTwin 2.0.

\section{Baselines}
\label{sec:appendix:baselines}

\subsection{Robotic Harnesses (Bob)}
\label{app:baseline_harnesses}

\minisection{Shared interface}
Alice works with every Bob through the same interface.
Bob provides its policy as source code, a way to run the policy on a list of scenes that returns the success and the recorded execution of each run, and a refinement step.
The refinement step receives the current policy, the accepted proposal with its instances and the recorded runs on them, and the scenes that the policy must keep; it returns a candidate policy.
Bob refines with its own prompts, tools, and optimizer; only the scenes and tasks that it refines on come from Alice.
ASPIRE, RHO, and \method{}-Bob run in the same LIBERO evaluator with the same perception and control tools, and GaP keeps its own runtime.
Both players use Sonnet 5.5 unless stated otherwise.

\minisection{ASPIRE}
The policy of ASPIRE~\cite{lu2026aspire} is one Python program.
A refinement is one Claude Code session with ASPIRE's own repair prompt: the session reads the recorded runs, may run the policy on the training scenes, and writes a new program with its diagnosis.
A session lasts at most 20 minutes (40 minutes in Goal and LIBERO-10 in the cooperative game).
The starting policy is the one that ASPIRE generates and refines on the development scenes of the task.

\minisection{RHO}
The policy of RHO~\cite{rho2026} is a small repository of files, with reusable skills and an entry point.
We run RHO's own search unchanged: in each cycle, one search of two generations of mutations takes the place of Bob's two refinement attempts, with at most 40 model turns per mutation and minibatches of 15 scenes.
The starting policy is the best policy of RHO's search on the development scenes of the task.

\minisection{GaP}
The policy of GaP~\cite{chen2026gap} is a graph of perception and action nodes with their scripts.
A refinement is one session of GaP's official self-learning agent, with its own skills and at most 80 turns and 20 minutes.
The session receives only public feedback: the outcome of every run, and the node trace and up to 16 camera frames of up to six runs.
At run time, the graph calls the vision--language model Gemini 3.1 Flash Lite at temperature 0.
The starting graph is the best of GaP's first graph and up to eight repairs on the development scenes.
Our adapter changes only the interface: the step limit, the objective given to the session, the conversion of positions to the world frame for Alice's tools, and a reminder of earlier rejected candidates.

\minisection{Frontier-model baseline (Fable 5.2)}
Fable 5.2 runs ASPIRE as Bob alone on the 30 tasks of Spatial, Object, and Goal, with the guideline of Appendix~\ref{sec:appendix:frontier-prompt} in its prompt.
Like the Bob-only baseline with Sonnet 5.5, it refines on the development scenes of the original task.
Table~\ref{tab:api-costs} gives the API cost of every task.
For Sonnet 5.5 with Alice, the cost covers generating and refining the policy on the original scenes and the eight cycles with Alice, for both players.
All costs are estimated from the recorded token counts at list prices (per million tokens, Sonnet 5.5: \$2 input, \$10 output, \$0.10 cache read, \$2.50 cache write; Fable 5.2: \$5, \$25, \$0.50, and \$6.25).
\begin{table}[t]
    \centering
    \caption{\textbf{API cost per task (USD)} of Table~\ref{tab:frontier-model}, with ASPIRE as Bob, for the 30 tasks of Spatial, Object, and Goal. For Sonnet 5.5 with Alice, the cost covers generating and refining the policy on the original scenes and the eight cycles of the adversarial game, for Alice and for Bob. Costs are estimated from the recorded tokens at list prices.}
    \label{tab:api-costs}
    \begingroup
    \footnotesize
    \setlength{\tabcolsep}{6pt}
    \renewcommand{\arraystretch}{1.0}
    \begin{tabular}{@{}lrr@{}}
        \toprule
        & Sonnet 5.5 + Alice & Fable 5.2 \\
        \midrule
        \multicolumn{3}{@{}l}{\textit{Spatial}} \\
        \quad Task 0 & 7.67 & 22.41 \\
        \quad Task 1 & 8.93 & 23.98 \\
        \quad Task 2 & 6.49 & 21.47 \\
        \quad Task 3 & 5.48 & 23.90 \\
        \quad Task 4 & 11.24 & 27.57 \\
        \quad Task 5 & 5.28 & 23.07 \\
        \quad Task 6 & 7.59 & 19.81 \\
        \quad Task 7 & 5.58 & 22.75 \\
        \quad Task 8 & 8.42 & 19.45 \\
        \quad Task 9 & 8.56 & 22.59 \\
        \addlinespace[2pt]
        \multicolumn{3}{@{}l}{\textit{Object}} \\
        \quad Task 0 & 5.70 & 22.59 \\
        \quad Task 1 & 6.04 & 20.57 \\
        \quad Task 2 & 5.43 & 22.80 \\
        \quad Task 3 & 7.49 & 22.96 \\
        \quad Task 4 & 14.60 & 24.47 \\
        \quad Task 5 & 5.14 & 22.08 \\
        \quad Task 6 & 5.69 & 22.92 \\
        \quad Task 7 & 8.04 & 22.63 \\
        \quad Task 8 & 7.78 & 20.62 \\
        \quad Task 9 & 6.47 & 21.76 \\
        \addlinespace[2pt]
        \multicolumn{3}{@{}l}{\textit{Goal}} \\
        \quad Task 0 & 6.52 & 23.12 \\
        \quad Task 1 & 7.78 & 21.12 \\
        \quad Task 2 & 7.03 & 23.92 \\
        \quad Task 3 & 13.49 & 27.01 \\
        \quad Task 4 & 7.32 & 22.59 \\
        \quad Task 5 & 12.52 & 26.54 \\
        \quad Task 6 & 7.96 & 21.00 \\
        \quad Task 7 & 5.10 & 23.74 \\
        \quad Task 8 & 6.71 & 21.60 \\
        \quad Task 9 & 6.45 & 23.67 \\
        \addlinespace[2pt]
        \midrule
        Mean, Spatial & 7.52 & 22.70 \\
        Mean, Object & 7.24 & 22.34 \\
        Mean, Goal & 8.09 & 23.43 \\
        \textbf{Mean, 30 tasks} & \textbf{7.62} & \textbf{22.82} \\
        Total, 30 tasks & 228.52 & 684.71 \\
        \bottomrule
    \end{tabular}
    \endgroup
\end{table}

\subsection{Robustness Guideline for the Model}
\label{sec:appendix:frontier-prompt}
When Fable 5.2 runs as Bob alone (Sec.~\ref{subsec:eval:adversarial}, Table~\ref{tab:frontier-model}), its Bob prompt is augmented with a guideline that asks Bob to make the policy robust to changes in the camera images and during the execution.
The guideline also gives examples of these changes, as in Table~\ref{tab:scene_changes}, together with changes to the configuration and control of the robot.

\subsection{Proposer and Curriculum Baselines (Alice Ablations)}
\label{app:baseline_proposers}
Every proposer starts from the same policy and the same Bob, here ASPIRE with Sonnet 5.5, and replaces only Alice's choice of the next scene or task.
In the adversarial ablation (Table~\ref{tab:baselines-ablations}), every proposer has at most eight cycles, two refinement attempts per cycle, and 50 test runs per cycle to screen candidates (Alice: 48 runs and 4 perception probes).
A candidate scene applies one scene function with its parameters to five training layouts.
For random proposals, SFL, and Eurekaverse, Bob's refined policy is kept under the same two conditions as with Alice (Appendix~\ref{app:refinement_cycle}): it succeeds on at least four of the five instances and on more of them than the current policy, and it loses none of the scenes of the original task and of earlier proposals.
In the cooperative ablation (Table~\ref{tab:cooperative-curricula}), every curriculum runs in the same cycle as cooperative Alice, with the same acceptance and number of cycles (Table~\ref{tab:settings-curricula}).

\minisection{Uniformly random proposals}
In the adversarial game, the proposer draws a scene function uniformly, draws its parameters uniformly within their ranges, and makes it static or dynamic with equal probability.
It screens up to ten such draws and takes the first on which the current policy fails at least once in five runs.
In the cooperative game, it draws a task uniformly from the scene functions of the suite.
Neither reads more than the success of each run, and neither calls a model.

\minisection{Sampling for Learnability (SFL)}
SFL~\cite{rutherford2024noregrets} screens nine random scenes on five layouts each and scores each by its learnability $p(1-p)$, where $p$ is the success rate of the current policy.
It keeps the four best scenes in a buffer and, with probability 0.5, takes one of them at random; otherwise it screens and takes a new random scene.
The buffer is rebuilt in every cycle, and SFL reads only the success of each run.

\minisection{Eurekaverse}
Eurekaverse~\cite{liang2024eurekaverse} evolves scenes with a language model, here Sonnet 5.5, in four generations of two branches from the same parent policy.
The model sees the instruction, the objects, the scene functions with their ranges, the parent scene, and earlier scenes with the success of the policy before and after each refinement, but not the policy, its runs, or images.
It is asked for a harder scene if the parent policy succeeds on more than 80\% of the runs, an easier one below 20\%, and a variation otherwise.
Each branch refines the policy at most twice; the parent and the two branches are scored on all valid scenes generated so far, and the best continues.

\minisection{Teacher--Student Curriculum Learning (TSCL)}
TSCL~\cite{matiisen2019teacher} groups the tasks that the scene functions of a suite can build into categories and keeps the last 10 rewards of each category, where a reward is the change in Bob's success on the task of a cycle.
In each cycle it samples one reward per category and draws a random task of the category with the largest absolute sampled reward (Table~\ref{tab:settings-curricula}).
It calls no model.

\minisection{OMNI-EPIC}
OMNI-EPIC~\cite{faldor2025omniepic} uses the same model as Alice, without Alice's tools.
The model sees up to five learned and five failed tasks of the archive and writes a next task that is new and learnable; a rejected task is returned with the error, up to five times, and a second call judges whether the task is interesting next to ten tasks of the archive. A task that Bob fails is modified once (Table~\ref{tab:settings-curricula}).

\minisection{Trajectory-only Alice}
Trajectory-only Alice follows the full procedure of Alice, with the same tools, budgets, and checks, but without the policy: it cannot read the policy's source code, and the recorded runs that it sees keep the tool calls, camera images, robot states, and outcomes but not the names of the policy's functions, its return values, or its exceptions.

\section{Hardware Setup}
\label{sec:appendix:hardware_setup}
The real-world experiments (Sec.~\ref{subsec:eval:real}) run the frozen programs from simulation on a Franka arm of the DROID platform~\cite{khazatsky2024droid}.
The robot implements the same API as the simulator, so each program runs without any change.
No language model is called during a trial, and neither Alice nor Bob takes part in it.

\subsection{Embodiment}

\subsubsection{Franka Arm}
The arm is a 7-DoF Franka Emika Panda, controlled through the software stack of DROID.
The robot environment of DROID, a bridge to it, the program, and the SAM~3 and PyRoki services all run on one workstation with an NVIDIA RTX A6000 GPU, and the bridge accepts commands only from this machine.
The API plans and retimes every joint path on the side of the program, and the bridge only plays it out on the arm at 15\,Hz.
The bridge stops the arm and refuses further commands when the tracking error exceeds 0.15\,rad, a command misses its deadline of 0.1\,s, the robot state or an image is stale, or the arm does not settle within 3\,s after a motion.

\subsubsection{Robotiq Gripper}
The gripper is a Robotiq 2F-85, whose stroke of 85\,mm is close to the 80\,mm of the Panda gripper of LIBERO.
It only opens or closes: a gripper command above 0.5 closes it, and one below opens it.
The tool centre point lies 0.16\,m from the hand frame, so a position computed by a program refers to the point between the jaws, as in simulation.

\subsection{Camera Setup}
\label{app:hardware_setup:cameras}
The scene camera is a ZED~2i and the wrist camera is a ZED Mini.
Both give 1280$\times$720 RGB-D images at 60 frames per second, with metric depth from the neural stereo of the ZED SDK (version 5.0.7) from 0.1\,m.
The control loop of 15\,Hz reads one pair of frames with each robot state, and a program sees a new image only when it calls a perception tool, so it re-observes the scene at the points in its code that call such a tool, not at every frame.
From each camera, a program receives the left RGB image, the depth from the same frame, the intrinsics, and the pose of the camera in the robot base frame.
The cameras keep the names that the programs use in simulation, so the programs read them without any change.

\minisection{Scene camera}
The scene camera stands about 0.56\,m to the right of the robot, roughly level with its base and 0.39\,m above the table, and looks across the workspace.
It plays the role of the front camera of LIBERO, and its pose is calibrated with the ChArUco procedure of DROID and does not change during a trial.
All perception of the policy of Bob alone uses this camera.

\minisection{Wrist camera}
The wrist camera is mounted on the hand, and its pose in the base frame is its calibration composed with the current state of the arm.

\subsection{Perception and Control Interface}
\label{app:hardware_setup:control}
A program calls the tools of Table~\ref{tab:real-tools}, the same tools as in simulation.

\begin{table}[t]
    \centering
    \caption{\textbf{Tools of a program on the robot.} The names and arguments are those of simulation; the second column gives how the robot implements each tool.}
    \label{tab:real-tools}
    \begingroup
    \footnotesize
    \setlength{\tabcolsep}{5pt}
    \begin{tabularx}{\columnwidth}{@{}>{\raggedright\arraybackslash}p{0.46\columnwidth}>{\raggedright\arraybackslash}X@{}}
        \toprule
        \textbf{Tool} & \textbf{On the robot} \\
        \midrule
        \texttt{segment\_text(\allowbreak camera,\allowbreak  text)} & Masks, boxes, and scores of the objects named by the text, from SAM~3 \\
        \texttt{mask\_to\_point\_cloud(\allowbreak mask,\allowbreak  camera)} & Points of a mask in the base frame, from the depth and camera pose of the frame that produced the mask \\
        \texttt{localize\_object(\allowbreak text,\allowbreak  camera)} & Points of a named object in the base frame \\
        \texttt{get\_robot\_state()} & Joint positions, pose of the hand, and gripper width \\
        \texttt{move\_to\_pose(\allowbreak pose)} & Inverse kinematics and a joint path from PyRoki, retimed by the API and played out by the bridge \\
        \texttt{go\_home()} & A move to the home pose of the station \\
        \texttt{open\_gripper()}, \newline \texttt{close\_gripper()} & A gripper command \\
        \texttt{step(\allowbreak action)} & A joint target: a small move is streamed to the bridge, a larger one runs as a retimed path \\
        \bottomrule
    \end{tabularx}
    \endgroup
\end{table}

\subsubsection{Perception Tools}
A program finds an object by asking for it in words: SAM~3~\cite{carion2025sam3} segments the named object in an image of one camera, and the mask is turned into points in the base frame with the depth and the camera pose of the same frame.
The programs compute every grasp and drop position from these points, such as the centre and top of the can or the rim of the box, so no table height or object position is fixed in a program.

\subsubsection{Control Tools}
A program moves the arm with three tools: \texttt{move\_to\_pose} solves the inverse kinematics and plans a joint path with PyRoki~\cite{kim2025pyroki}, \texttt{go\_home} returns the arm to its home pose, and \texttt{step} sends a joint target.

\end{document}